\providecommand{\arxivbuild}{1}     
\providecommand{\acceptedbuild}{}  

\documentclass[10pt]{article}

\ifx\acceptedbuild\empty
  \ifx\arxivbuild\empty
    \usepackage{tmlr}                  
  \else
    \usepackage[preprint]{tmlr}        
  \fi
\else
  \usepackage[accepted]{tmlr}          
\fi

\usepackage[utf8]{inputenc}
\usepackage{hyperref}
\usepackage{url}
\usepackage{booktabs}
\usepackage{amsfonts}
\usepackage{amsmath}
\usepackage{amssymb}
\usepackage{nicefrac}
\usepackage{microtype}
\usepackage{xcolor}
\usepackage{graphicx}
\usepackage{subcaption}
\usepackage{siunitx}
\usepackage{tikz}
\usetikzlibrary{positioning,arrows.meta,calc,shapes.geometric,fit,backgrounds,patterns}
\usepackage{pgfplots}
\pgfplotsset{compat=1.18}

\ifx\acceptedbuild\empty\else
  \def\month{MM}
  \def\year{YYYY}
  \def\openreview{\url{https://openreview.net/forum?id=XXXX}}
\fi

\newcommand{\nats}[1]{\ensuremath{#1\,\mathrm{nats}}}

\ifx\acceptedbuild\empty
  \ifx\arxivbuild\empty
    \newcommand{\anonymized}[2]{#2}     
  \else
    \newcommand{\anonymized}[2]{#1}     
  \fi
\else
  \newcommand{\anonymized}[2]{#1}       
\fi

\title{Decomposing Evolutionary Mixture-of-LoRA Architectures:\\ The Routing Lever, the Lifecycle Penalty, and a Substrate-Conditional Boundary}

\author{\name Ramchand Kumaresan \email mechramc@gmail.com \\
      \addr Murai Labs}

\begin{document}

\maketitle

\begin{abstract}
We decompose an evolutionary mixture-of-LoRA system on a from-scratch
$\sim 150$M-parameter widened-$D$ substrate ($D{=}1536$, $V{=}32000$;
$D/V \approx 0.048$; we refer to this as the ``widened-1536''
substrate) into three factors --- a router rewrite (parallel sigmoid gate with
learnable per-adapter floor and bounded temperature anneal, fed
post-stack hidden states rather than token-embedding means), a
per-domain leave-one-out evaluation scope, and a lifecycle of death
plus $\alpha$-blend inheritance plus SVD mutation plus slot
reallocation --- and report a 5-of-8 partial $2^3$ factorial run at
$n{=}3$ seeds and $25000$ adaptation steps per cell. The attribution
chain is sharp on this substrate: the router rewrite carries the
entire $+0.0426$~nat balanced log-PPL improvement (under the
convention $\Delta = \log\mathrm{PPL}_\mathrm{ref} -
\log\mathrm{PPL}_\mathrm{test}$, positive $=$ improvement;
$t{=}12.86$, $p{=}0.006$) attributed to ``the full evolutionary
system vs the static B3 baseline'' in our own internal numbers; the
headline full-system-vs-B3 balanced contrast itself is $+0.015$ nats,
$t{=}1.94$, $p{=}0.19$ at $n{=}3$ and does not clear $\alpha{=}0.05$.
The per-domain evaluation scope is null at seed-resolution, and the
lifecycle is a net drag of $\approx -0.028$~nats
($t{=}-4.46$, $p{=}0.047$ in the primary chain). An auxiliary $\alpha{=}0$ inheritance counterfactual at $n{=}3$ seeds
is sign-inconsistent at the headline metric and underpowered for
either an equivalence or a load-bearing conclusion (an earlier draft
of this paper claimed the counterfactual ruled out inheritance as the
load-bearing source of the lifecycle penalty; that claim depended on
an arithmetic-mean aggregator that, when corrected to the pinned
geometric mean, flips the seed-42 cell to the load-bearing range and
leaves the cross-seed mean inside the ambiguous band, see
Appendix~\ref{app:audit-e10}). A base-perturbation probe
directionally refutes a ``genomic-context'' reframe of the lifecycle
role. A controllable synthetic sandbox locates a substrate-conditional
regime boundary: evolutionary search on the routing channel is
load-bearing only when adapters are pre-aligned to the task; in every
other regime tested it underperforms, ties, or actively degrades the
gradient solution.
\end{abstract}

\section{Introduction}
\label{sec:introduction}

Modern parameter-efficient adaptation of large language models layers
many low-rank adapters on top of a frozen backbone, often gated by a
sparse mixture-of-experts router that selects a subset of adapters per
input token~\citep{hu2021lora,li2024mixlora}.
A natural further step is to make the population of adapters itself
\emph{evolutionary}: adapters compete on a fitness signal, the worst
adapters die, vacated slots are repopulated by mutated clones of the
fittest, and a fraction of the dying adapter's weights is blended into a
nearest neighbor before slot release. The hypothesis underlying such an
``evolutionary mixture-of-LoRA'' design --- that lifecycle dynamics
(selection, reproduction, inheritance, mutation) produce
better-organized adapter populations than static allocation --- is
plausible by analogy with neuroevolution and population-based
training~\citep{stanley2019neuroevolution,jaderberg2017pbt}, but the
empirical record on whether this analogy carries over to text-domain
mixture-of-LoRA training has been thin. This paper supplies that
empirical record on a single specific substrate, and reports a result
that is more nuanced than either a clean win or a clean null.

\paragraph{One-sentence contribution.}
On a 5-of-8 partial $2^3$ factorial decomposing an evolutionary
mixture-of-LoRA system into three factors, the
\emph{router rewrite} factor carries the entire balanced-perplexity
improvement over the static baseline, the \emph{lifecycle} factor is a
net drag, and the \emph{evaluation-scope} factor is null at
seed-resolution; the contribution decomposes into a routing lever, a
lifecycle penalty, and a substrate-conditional regime boundary that we
characterize first on a synthetic sandbox and then on the production
run.

\paragraph{What.}
We report three concrete claims, each backed by a paired statistical
test on three seeds:

\begin{itemize}
\item \textbf{The routing lever.} Replacing a softmax-over-adapters
  router (which on this widened-1536 substrate collapses onto a
  single 4-of-16 cross-domain adapter coalition; Jensen--Shannon
  divergence between per-domain top-$k$ gate distributions equals
  exactly~$0$) with a parallel-sigmoid router carrying a learnable
  per-adapter floor and a bounded temperature anneal, and feeding it
  post-stack hidden states rather than token-embedding means, accounts
  for the entire $+0.0426$ nat balanced log-PPL improvement
  ($t{=}12.86$, $p{=}0.006$) attributed to ``the full evolutionary
  system vs the static B3 baseline'' in our own internal numbers
  (Section~\ref{sec:real-text-decomposition}).
\item \textbf{The lifecycle penalty.} Death, $\alpha$-blend inheritance,
  SVD mutation, and slot reallocation, layered on top of the rewritten
  router, cost approximately~\nats{0.028} on balanced log-PPL
  ($t{=}-4.46$, $p{=}0.047$ in the primary attribution chain;
  $t{=}-3.47$, $p{=}0.074$ in the consistency chain). An auxiliary
  $\alpha{=}0$ inheritance counterfactual at $n{=}3$ seeds is
  sign-inconsistent at the headline metric ($+3.18$\% on seed~42,
  $-1.65$\% on seed~137, $+0.20$\% on seed~256, mean $+0.56$\%) and
  is underpowered for either a load-bearing or an equivalence
  conclusion on the inheritance sub-component (an earlier draft of
  this paper claimed the counterfactual ruled out inheritance under
  an arithmetic-mean aggregator; that claim is retracted, see
  Appendix~\ref{app:audit-e10}). A base-perturbation probe
  directionally refutes a ``genomic-context'' reframe of the lifecycle
  role (Section~\ref{sec:phaseB-auxiliary}).
\item \textbf{A substrate-conditional regime boundary.} On a controllable
  synthetic sandbox where ground truth is known, evolutionary search on
  the routing channel is load-bearing \emph{only} when adapters are
  pre-aligned to the task; in every other regime tested (joint random
  initialization, gradient-warm initialization, hybrid ES-then-SGD)
  evolutionary search underperforms, ties, or actively degrades the
  gradient solution (Section~\ref{sec:synthetic-boundary}). The
  production-substrate result is consistent with this boundary.
\end{itemize}

\paragraph{Why (evidence).}
The decomposition is not retrospective. We pre-specified a 5-of-8
partial $2^3$ factorial over three factors --- F1 router
rewrite, F2 per-domain LOO evaluation scope, F3 lifecycle dynamics ---
and ran $n{=}3$ seeds per cell at $25000$ training steps each on the
widened-1536 from-scratch substrate (15 production runs in the
core factorial; 4 additional runs in the Fork~0 seed sweep). Numerical
claims in this paper trace to source-of-truth analysis JSONs in
\anonymized{\texttt{experiments/trackb\_b2/analysis/}}{the analysis
directory listed in supplementary materials}; per-cell run identifiers,
per-seed PPLs, and a full re-derivation of the headline statistics
from those JSONs are reproduced in
Appendix~\ref{app:reproducibility} and
Appendix~\ref{app:phase3-rederivation}.

\paragraph{So what (significance).}
The most important finding is not that lifecycle dynamics fail, but
\emph{that they fail in a specific, characterizable way}: the system's
balanced-aggregate win on this substrate is carried by a structural
routing fix, and the evolutionary machinery layered on top of that fix
is mildly anti-aligned with the gradient solution that the routing fix
unlocks. This is a sharper statement than ``lifecycle did not help'';
it is a statement about \emph{when, on what substrate, and against what
gradient signal} a particular evolutionary design might be expected to
contribute. The synthetic-sandbox result in
Section~\ref{sec:synthetic-boundary} provides a candidate sufficient
condition for ES contribution (oracle-aligned adapters carrying
informative routing signal). The production-substrate result is not in
that regime. We make no claim that mixture-of-LoRA evolution
\emph{cannot} pay rent --- only that the particular configuration we
tested, on the particular substrate we tested it on, does not. We
expect this paper to be most useful to readers who are about to spend
compute on a similar evolutionary design and want a falsifiable prior
on which sub-mechanisms are most likely to carry value.

\paragraph{What the paper is not.}
We are not claiming a state-of-the-art mixture-of-LoRA result; the
absolute perplexity numbers are produced by a $\sim 150$M-parameter
from-scratch base trained for $70000$ steps on a domain-mixed
$32$k-vocab corpus, and the $25000$-step adaptation budget is small.
We are not claiming the lifecycle penalty is universal; the budget,
the inheritance schedule, the mutation rate, the lifecycle cadence, the
specific kill gate, and the substrate width are all design choices that
could be wrong. We are not claiming that the synthetic-sandbox
oracle-alignment boundary transfers cleanly to all production
mixture-of-LoRA setups --- only that it transfers in the one direction we
have measured (a substrate not in the oracle-alignment regime fails to
benefit from ES, consistent with the boundary). The
limitations~(Section~\ref{sec:limitations}) are as central to the
paper as the headline numbers.

\paragraph{Roadmap.}
Section~\ref{sec:related-work} positions the work against
mixture-of-experts, parameter-efficient adaptation, and the
evolutionary-strategies-vs-gradient-descent literature.
Section~\ref{sec:infrastructure} describes the substrate, eval
pipeline, and audit chain. Section~\ref{sec:synthetic-boundary} reports
the synthetic-sandbox boundary result.
Section~\ref{sec:real-text-decomposition} reports the
production-substrate $5$-cell partial factorial.
Section~\ref{sec:mechanism} explains why the router rewrite carries the
win. Section~\ref{sec:phaseB-auxiliary} reports the auxiliary
ablations on the lifecycle penalty (Phase~B and Fork~0).
Section~\ref{sec:limitations} states the limitations in the
TMLR-conventional extensive form, and
Section~\ref{sec:future-work} lists the planned follow-ups, including
the four-cell sub-factorial that would isolate the gate-function rewrite
from the routing-input change inside F1, and the F3 sub-component
decomposition that would isolate death and reproduction from inheritance
and mutation.

\section{Related work}
\label{sec:related-work}

We organize the related work methodologically rather than
paper-by-paper, around the three factors that the
production-substrate factorial decomposes
(Section~\ref{sec:real-text-decomposition}): the routing channel
itself, the evolutionary lifecycle that operates on the adapter
population, and the evaluation methodology that aggregates per-domain
signals into a balanced headline.

\paragraph{Routing in mixture-of-experts and mixture-of-LoRA.}
The closest prior work to our F1 factor is the line on top-$k$ routed
mixture-of-experts, in which a parameterized router selects a sparse
subset of expert sub-networks per token and the un-selected experts are
not computed~\citep{fedus2021switch,zoph2022stmoe}. Two
recurring pathologies have been reported in this line: (i) router
collapse, in which a small subset of experts captures all of the routing
mass and the others are routing-dead, addressed by load-balancing
auxiliary losses and z-loss
penalties~\citep{zoph2022stmoe}; and (ii)
routing-input under-specification, in which the gate function does not
have access to a representation rich enough to differentiate expert
specialties, addressed by alternative routing schemes such as
expert-choice~\citep{zhou2022expertchoice}. Mixture-of-LoRA
extensions of this line replace dense MLP experts with parameter-efficient
LoRA adapters layered on a frozen
backbone~\citep{hu2021lora,li2024mixlora}. The coalition
monopoly we measure on the legacy softmax router
(Section~\ref{sec:coalition-monopoly}) is in pathology family (i),
and the routing-input change from token-embedding mean to post-stack
hidden state is in pathology family (ii). Our F1 contribution is not in
identifying these pathologies, which are known, but in
\emph{decomposing the production-substrate gain by attribution chain}
and showing that the routing fix carries the entire balanced-PPL win on
this substrate.

\paragraph{Evolutionary search on neural-network parameters.}
The evolutionary side of the system inherits from a long line of
neuroevolution and population-based work. Evolution strategies (ES)
have been studied as a scalable alternative to reinforcement learning
on policy-network parameters~\citep{salimans2017es}, and as a
generic search method on neural architectures and
weights~\citep{stanley2019neuroevolution}. Whether ES is best read as
a finite-difference gradient estimator or as a fundamentally different
search scheme has been studied
directly~\citep{lehman2018esgradients}; the question is
empirical, depends on the regime, and informs the synthetic-sandbox
oracle-alignment boundary we report in
Section~\ref{sec:synthetic-boundary}. Population-based training of
neural networks~\citep{jaderberg2017pbt} introduces a related but
distinct mechanism: copying the weights of a fitter model into a less
fit one and mutating its hyperparameters in place. Hardware-accelerated
neuroevolution frameworks~\citep{tang2022evojax} make population
methods tractable on modern accelerators. Our F3 contribution is not in
introducing a new evolutionary primitive --- death, reproduction with
heritable mutation, and inheritance via $\alpha$-blend into a nearest
neighbor are familiar pieces --- but in measuring whether their
\emph{aggregate effect} on a particular adapter-evolution substrate is
positive, null, or negative.

\paragraph{Evolutionary methods on large language models.}
A more directly competitive recent literature applies evolutionary
search to large language models themselves rather than to small policy
networks. \citet{qiu2025esatscale} report that ES on full-parameter LLM
fine-tuning is competitive with reinforcement learning at the scales
tested; \citet{sarkar2025eggroll} introduce a low-rank-structured ES
variant (EGGROLL) targeting hyperscale; and
\citet{korotyshova2025essa} apply CMA-ES to LoRA singular values up to
72B parameters. \citet{liang2026variancecurvature} give a
variance-curvature account of why these low-rank evolutionary methods
often outperform full-parameter search, which our synthetic-sandbox
boundary (Section~\ref{sec:synthetic-boundary}) reads as
direction-aligning with the oracle-alignment requirement we measure.
\citet{feng2025heteroswarms} jointly optimize the roles and weights of
multi-LLM systems via particle-swarm optimization. The Sakana line on
evolutionary model-merging
(\citealp{akiba2024modelmerge,kuroki2024cycleqd,abrantes2025m2n2}) and
the SVF/Transformer-Squared self-adaptation work of
\citet{sun2025transformer2} are closest in spirit to our F1 factor: the
gains there come from a \emph{structural} change to the
adapter-combination machinery rather than from in-training population
dynamics, which matches our finding that the routing rewrite carries
the entire balanced-PPL win.
\citet{abdi2026esforget} report that ES on LLM fine-tuning induces
catastrophic forgetting, and \citet{kumaresan2026kalavai} provide a
quantitative population-fusion model for cooperative LLM training. The
lifecycle penalty we report is sign-consistent with the forgetting
result of \citet{abdi2026esforget}, although the substrate is
mechanistically distinct (per-layer LoRA adapters under top-$k$ routing
on a frozen base, vs.\ full-parameter ES on an unfrozen base). What
the present paper adds over this literature is \emph{factor
isolation}: rather than reporting an aggregate full-system-vs-static
contrast, we attribute the contrast to a router rewrite (load-bearing),
a per-domain leave-one-out scope (null), and a lifecycle (net drag),
which lets the paper speak both to ``does evolutionary search on LLM
adapters work'' and to ``which factor of the standard
evolutionary-mixture-of-LoRA recipe carries the work.''

\paragraph{Recycling, autophagy, and the cost of population dynamics.}
A separate line of work studies what happens when generative models
consume their own outputs across generations, finding that a model
trained on data generated by an earlier generation of itself can suffer
a ``model autophagy disorder'' in which diversity collapses and
reconstruction fidelity
degrades~\citep{alemohammad2023mad}. The lifecycle penalty we report is
not autophagy in this sense (the base model is frozen and there is no
generative-output feedback loop), but the analogy informs why
inheritance --- a mechanism that explicitly recycles a fraction of the
dying adapter into the surviving population --- might pay a cost rather
than a benefit. The Fork~0 Phase~A and Phase~B
ablations~(Section~\ref{sec:fork0-phaseA},
\ref{sec:phaseB-counterfactual}) test for this directly and locate the
inheritance-versus-survival correlation as age-confounded rather than
causal.

\paragraph{Selection versus self-correction.}
A complementary observation in the language-model literature is that
LLMs frequently fail to self-correct their own reasoning when the
correctness signal is internal rather than externally
verified~\citep{huang2024selfcorrect}. The selection step in
our lifecycle (a held-out leave-one-out fitness evaluation that decides
which adapter dies) is in spirit a population-level external
correctness signal, in contrast to a single-adapter self-correction
loop. We do not take a position on whether population selection
generally beats self-correction in the language-model setting, but we
note that the \emph{form} of the signal in our system --- a per-adapter
loss on a held-out micro-batch with forced single-adapter routing --- is
of the kind the self-correction-failure literature would expect to be
load-bearing if anything is. The fact that this strong selection signal
co-exists with a net lifecycle penalty
(Section~\ref{sec:real-text-decomposition}) is therefore informative:
it suggests the penalty is not from a poor-quality fitness signal but
from somewhere else in the F3 mechanism (mutation distribution, slot
reallocation policy, lifecycle cadence; see
Section~\ref{sec:f3-synthesis}).

\paragraph{Multi-level optimization framing.}
A recent line frames many learning systems as nested
optimizers~\citep{behrouz2025nested}: a fast inner loop (here,
SGD on adapter parameters) embedded inside a slower outer loop (here,
the evolutionary lifecycle on the adapter population). On this framing,
the lifecycle penalty we measure is a statement about the
\emph{coupling} between the two levels --- specifically, that on this
substrate, the outer-loop selection-and-mutation schedule is mildly
anti-aligned with the inner-loop gradient flow, rather than a statement
that two-level optimization is unhelpful in general.

\paragraph{Reproducibility and verification methodology.}
The verification protocol applied to this paper draws on the broader
machine-learning reproducibility
literature~\citep{pineau2020repro,henderson2018drl}: every numerical
claim in this paper carries a comment-level link to a single
source-of-truth JSON or analysis document, and
Appendix~\ref{app:audit-memo} logs the corrections applied during
paper preparation as a transparency artifact for readers who want to
verify reproducibility against the supplement.

\section{Infrastructure: substrate, evaluation, and source-of-truth conventions}
\label{sec:infrastructure}

This section describes the production substrate on which the
real-text factorial of Section~\ref{sec:real-text-decomposition} runs,
the evaluation pipeline that aggregates per-domain held-out signal into
the balanced log-PPL headline, and the audit chain that anchors every
numerical claim to a source-of-truth file. The synthetic sandbox of
Section~\ref{sec:synthetic-boundary} runs on a separate, much smaller
substrate documented at the top of that section.

\subsection{Substrate}
\label{sec:substrate}

The base model is a $\sim 150$M-parameter from-scratch GPT-style
transformer with hidden size $D{=}1536$, $12$ layers, $16$ attention
heads, sequence length $1024$, vocabulary size $V{=}32000$. The
$D/V{=}1536/32000 \approx 0.048$ ratio is intentionally widened relative
to the prior $D{=}1024, V{=}32000$ ($D/V \approx 0.032$) substrate, to
reduce the gradient-norm-destruction effect at the LM head\footnote{An
\anonymized{in-code comment in \texttt{uyir/training.py}}{internal
design note} attributes this design choice to internal experiments
showing that wider $D/V$ ratios produce adapter populations with
measurably more cross-domain signal in routing inputs; the F1
routing-input change in Section~\ref{sec:routing-input} relies on the
same effect.} and to give
adapters more headroom to differentiate. The base is pre-trained for
$70000$ steps on a $32$k-vocabulary domain-mixed corpus
(\texttt{tokenized\_mixed32k\_b2}) covering biology, code, general
text, and science domains. \anonymized{The frozen pre-train checkpoint
\texttt{checkpoints/base\_model\_trackb\_b2\_dv1536/base\_step\_70000.pt}
is the canonical base for all production-substrate experiments in this
paper.}{The canonical frozen base checkpoint at pre-train step $70000$
(path in supplementary materials) is used for all production-substrate
experiments in this paper.} We acknowledge in Section~\ref{sec:limitations} that the base
training was halted at step $70000$ rather than at the originally
planned step $\sim 143000$ owing to a UPS power-event interruption; the
ablations in this paper are therefore comparable against each other but
not against the originally-planned, longer-trained base.

LoRA adapters are injected at every transformer block on the
\texttt{query\_key\_value}, \texttt{dense}, \texttt{dense\_h\_to\_4h},
and \texttt{dense\_4h\_to\_h} projections, with rank $r{=}8$ and
$\alpha{=}16$. The static population size is $P{=}16$ adapters; the
sparse top-$k$ is $k{=}4$. Adapter-side optimization uses AdamW with
learning rate $10^{-4}$; the router uses a decoupled AdamW at
$10^{-5}$. The adaptation horizon is $25000$ steps for each cell of the
factorial.

\subsection{Evaluation pipeline}
\label{sec:eval-pipeline}

The eval signal is held-out per-domain perplexity, computed on a
balanced eval mixture across the four domains. The pipeline splits
into two layers.

\paragraph{Per-domain loaders.}
Each of the four domains has its own evaluation \texttt{DataLoader}
constructed from a pre-tokenized shard manifest
\anonymized{(\texttt{shard\_metadata.json})}{(the shard manifest;
file path in supplementary materials)}. Each shard carries an explicit
domain tag mapped through a canonical-name alias table; shards whose
domain tag is missing or unrecognized are excluded. The per-domain loaders
draw from the held-out split (\texttt{eval}) of the pre-tokenized
$32$k-vocab corpus, use a fixed eval microbatch size of $32$ sequences,
and are deterministic given the configuration's random seed.

\paragraph{The \texttt{StratifiedEvalLoader}.}
A pre-existing bug in the legacy \texttt{build\_dataloader(split="eval")}
path constructed a single concatenated eval loader by alphabetical
shard ordering, with the consequence that the first batches drawn from
the concatenated loader came entirely from the alphabetically-first
domain shards (in the \texttt{tokenized\_mixed32k\_b2} layout, this is
\texttt{biology}, since the four domain names ordered alphabetically
are \texttt{biology} $<$ \texttt{code} $<$ \texttt{general} $<$
\texttt{science} and the first concatenated shard is
\texttt{shard\_0000.npy} from the biology domain). Any $N$-batch eval
consumer therefore saw an
unbalanced mixture biased toward whichever domain appeared first in
shard-name order. \texttt{StratifiedEvalLoader} (in
\anonymized{\texttt{uyir/data\_utils.py:271-318}}{the data-utility
module referenced in supplementary materials}) yields exactly one batch
per domain per iteration cycle in alphabetical-domain-name order, with
per-domain iterator state persisted across \texttt{\_\_iter\_\_} calls
so that successive cycles progress through each domain's data rather
than replaying the first shard. The fix went in at git
\anonymized{\texttt{6b00021}}{\texttt{[commit redacted]}}; all
production runs in this paper use the post-fix stratified eval loader.
We list the
deflation-of-cross-seed-variance side-effect of this fix as a
limitation in Section~\ref{sec:limitations}.

\paragraph{Headline aggregator.}
We use the \emph{geometric mean} across domains as the headline
balanced-PPL aggregator: $\mathrm{PPL}_\mathrm{balanced} =
\exp\!\bigl(\tfrac{1}{4}\sum_d \log \mathrm{PPL}_d\bigr)$ over
$d \in \{\text{biology}, \text{code}, \text{general}, \text{science}\}$.
This convention is pinned in Section~\ref{sec:convention-pin}. The
geometric mean is the appropriate aggregator for log-PPL paired tests
because $\Delta \log \mathrm{PPL}_\mathrm{balanced}$ then equals the
mean of per-domain $\Delta \log \mathrm{PPL}_d$, which is the additive
quantity the per-seed paired-$t$ tests in
Section~\ref{sec:real-text-decomposition} act on. An arithmetic mean of
per-domain PPLs (used by some prior reports including an earlier
internal Phase~B summary) does not have this additive property and is
not used as the headline in this paper.

\paragraph{Two evaluation scopes.}
Factor F2 in Section~\ref{sec:real-text-decomposition} contrasts two
scopes: \emph{aggregate-LOO}, in which the four per-domain loaders are
flattened to a single uniform-mixed eval batch and a single LOO
fitness signal is computed against that aggregate; and
\emph{per-domain-LOO}, in which each adapter is evaluated against its
own per-domain held-out batch and the per-domain LOO scores are
averaged. The factorial measures whether the choice of scope on its own
contributes to the balanced-PPL win; the answer
(Section~\ref{sec:real-text-decomposition}) is null at seed-resolution.

\subsection{Source-of-truth conventions}
\label{sec:audit-chain}

Numerical claims in this paper trace to single source-of-truth files
in \anonymized{\texttt{experiments/trackb\_b2/analysis/}}{the analysis
directory listed in supplementary materials}, with a corresponding
\texttt{\% source: <path>:<key>} comment in the LaTeX immediately
adjacent to the cited number. We follow three conventions for
self-consistency. First, citations resolve to underlying analysis
JSONs or per-run outputs rather than to rolled-up summary files
(\texttt{RESULTS.md}, \texttt{STATUS.md}, project notes), which can
drift from the JSON they were meant to summarize. Second, sign
convention, aggregator, and degrees-of-freedom are pinned at section
entry rather than left implicit; Section~\ref{sec:convention-pin}
pins ``improvement-is-positive'' on log-PPL with $\Delta = \log
\mathrm{PPL}_\mathrm{ref} - \log \mathrm{PPL}_\mathrm{test}$ and the
geometric-mean aggregator. Third, headline statistics are re-derived
from per-seed JSON: the headline full-system-vs-B3 balanced contrast at
$n{=}3$ is paired-$t$ $t{=}1.94$, $p{=}0.19$, df${=}2$, and the
code-domain single-cell paired-$t$ is $t{=}10.42$, $p{=}0.009$,
df${=}2$ (full re-derivation in
Appendix~\ref{app:phase3-rederivation}). Corrections applied to
draft text during paper preparation are logged in
Appendix~\ref{app:audit-memo} as a transparency artifact.

\subsection{Reproducibility footprint}
\label{sec:reproducibility-footprint}

The full hyperparameter inventory, per-cell W\&B run identifiers, and
configuration JSONs for all 19 production runs (15 in the core
factorial, 4 in the Fork~0 seed sweep) are in
Appendix~\ref{app:reproducibility}. The synthetic-sandbox per-seed
numbers (Appendix~\ref{app:synthetic-full-tables}) and the
corrections log (Appendix~\ref{app:audit-memo}) are intentionally
extensive in keeping with TMLR's emphasis on extensive reproducibility
documentation.

\section{Synthetic boundary characterization (G4--G8)}
\label{sec:synthetic-boundary}

Before measuring whether evolutionary search adds value to a real
mixture-of-LoRA training pipeline, we ask whether evolutionary search adds
value \emph{anywhere} on a controllable substrate where ground truth is
known. This section describes a five-experiment battery (G4--G8) on a
purpose-built synthetic sandbox. The battery defines a regime boundary:
evolutionary strategies (ES) are load-bearing on the routing channel
\emph{only} when adapters are pre-aligned to the task; in every other
regime tested (joint random initialization, gradient-warm initialization,
hybrid ES-then-SGD schedules) ES underperforms, equates, or actively
degrades the gradient solution. Section~\ref{sec:real-text-decomposition}
inherits this boundary as the prior expectation against which the real-text
factorial is interpreted.

The full per-seed numbers and trajectories for each sweep are in
Appendix~\ref{app:synthetic-full-tables}; this section reports cross-seed
means with the relevant test statistic for each comparison.

\subsection{Sandbox design}
\label{sec:sandbox-design}

The sandbox is intentionally minimal so that any signal observed must come
from the routing-and-adapter machinery rather than from data structure that
could be solved by the base model alone. Vocabulary size is~128, partitioned
into four disjoint slabs of~32 tokens each. Each slab indexes a single
synthetic ``domain.'' Within a domain, the next token is a deterministic
permutation of the current token, restricted to that slab; the resulting
task is therefore a per-domain bigram prediction problem with a 32-way
within-slab branching factor and a perfect achievable per-token loss of
$\log(1) = 0$ given correct routing.\footnote{In practice the sweep code
mixes the deterministic bigram with a small uniform smoothing, so the
oracle endpoint is~\nats{\approx 0.042} rather than zero.
\anonymized{See \texttt{experiments/phase05\_es\_router/synth\_data.py}
for exact distributions.}{See sandbox configuration in supplementary
materials for exact distributions.}} Wrong-domain routing degrades to uniform over a slab,
yielding $\log(32) \approx \nats{3.47}$. Random routing across all four
slabs degrades further to $\log(128) \approx \nats{4.852}$. The substrate
is a small frozen base model (hidden dimension 512), with $K{=}16$ LoRA
adapters and top-$k{=}4$ routing per token.

This dimensioning gives three reference loss values on every sweep:
\textbf{oracle} (perfectly routed, perfectly specialised adapters,
$\approx \nats{0.042}$), \textbf{random-router} ($\approx \nats{0.089}$),
and \textbf{uniform} ($\nats{4.852}$).\footnote{The random-router floor
sits much closer to oracle than to uniform because the oracle-pretrain
of Phase~A bakes the per-domain mapping into the adapters: with $K{=}16$
adapters specialised four-per-slab and top-$k{=}4$ selection from
$\mathcal{N}(0, 0.02)$ router logits, the soft mass over the four
correct adapters is on average sufficient to recover most of the
within-slab signal even before any router learning. The oracle vs.\
random-router gap (\nats{\approx 0.047}) is therefore the headroom that
ES is asked to close in G4, not the full \nats{4.852} uniform-vs-oracle
spread.}

\subsection{G4: oracle adapters, ES on the router (PASS)}
\label{sec:g4-oracle}

The first experiment establishes that the routing channel is real and that
ES can find it given a pre-aligned substrate. The protocol is three-phase:
\textbf{(A)} oracle-routed pretrain for 3000 steps with adapter learning
rate~\num{3e-3}; domain~$d$ is forced to adapters $[d \cdot k, (d{+}1)\cdot k)$
with the router bypassed, so each adapter mechanically specialises on a
single domain. \textbf{(B)} snapshot, freeze the base and adapters, reset
the router from $\mathcal{N}(0, 0.02)$. \textbf{(C)} 1500 steps of
router-only training under each strategy. The strategies compared are SGD
on the router (with router learning rates in $\{10^{-4}, 10^{-2}\}$, both
returning the same outcome) and ES with antithetic mirror-pair noise at
$\sigma \in \{0, 10^{-3}, 10^{-2}, 10^{-1}, 1\}$, three seeds each. The
ES algorithm at every sigma uses $16$ antithetic mirror pairs per step
(i.e.\ $32$ candidate evaluations per step), Gaussian perturbations
$\epsilon \sim \mathcal{N}(0, \sigma^2 I)$ applied additively/symmetrically
to a flat parameter vector, and rank-based NES utility shaping. The
update at each step is $\theta \leftarrow \theta + (\eta/\sigma) \cdot
(1/P) \sum_{i=1}^{P} u_i \, \epsilon_i$ with $P{=}16$ pairs, outer
learning rate $\eta{=}1.0$, and utilities
$u_i \propto \max(0, \log(P/2{+}1) - \log(i{+}1))$ normalised to unit
$L_1$ norm and mean-centred to give zero update under uniform fitness
(reference implementation in \anonymized{\texttt{experiments/phase05\_es\_router/es\_router.py}}{the
ES module in supplementary materials}). The same algorithm and
hyperparameters are used for the coupled-ES experiments in G5--G8;
only the dimensionality of $\theta$ changes (router-only in G4,
router-plus-adapters in G5--G8).

Cross-seed means in Phase~C are summarised in
Table~\ref{tab:g4-positive-control}. SGD on the router closes
\textbf{0.2\,\%} of the routing gap (oracle $-$ random-router); ES at
$\sigma{=}0$ (provable no-op control) closes $-0.1\,\%$, confirming that
the ES infrastructure itself is not contaminating the comparison; ES at
$\sigma{=}10^{-1}$ closes \textbf{55.9\,\%} of the gap, with
$\sigma{=}10^{-3}$ at 17.4\,\%, $\sigma{=}10^{-2}$ at 26.2\,\%, and
$\sigma{=}1$ at 45.5\,\%. The $\sigma$-curve is monotone up to
$\sigma{=}10^{-1}$ and then degrades slightly at $\sigma{=}1$ (excess
exploration noise), the textbook signature of a meaningful ES signal.

\begin{table}[t]
\centering
\caption{G4 positive control: ES on the router, oracle-aligned adapters
frozen. Cross-seed mean ($n{=}3$) of Phase-C val loss; gap-closed is the
fraction of the oracle--random-router gap recovered. Best ES
($\sigma{=}10^{-1}$) closes 55.9\,\% of the gap; SGD closes 0.2\,\%.
The SGD-on-router and random-router rows both round to $0.0892$ at the
fourth decimal place; the $0.2\,\%$ gap-closed for SGD is computed from
the unrounded cross-seed mean (full-precision values in the
\texttt{positive\_control\_results.json} source-of-truth file).}
\label{tab:g4-positive-control}
\begin{tabular}{lrrrr}
\toprule
mode & $\sigma$ & mean val loss & std & gap closed \\
\midrule
oracle (lower bound) & --- & 0.0422 & --- & 100\,\% \\
random-router (upper bound) & --- & 0.0892 & --- & 0\,\% \\
\midrule
SGD on router & --- & 0.0892 & 0.0014 & 0.2\,\% \\
ES & 0 & 0.0893 & 0.0010 & $-0.1$\,\% \\
ES & $10^{-3}$ & 0.0811 & 0.0022 & 17.4\,\% \\
ES & $10^{-2}$ & 0.0769 & 0.0026 & 26.2\,\% \\
ES & $10^{-1}$ & \textbf{0.0629} & 0.0037 & \textbf{55.9}\,\% \\
ES & $1$ & 0.0678 & 0.0025 & 45.5\,\% \\
\bottomrule
\end{tabular}
\end{table}

The result is interpretable: with adapter--domain alignment baked in by
oracle pretraining, the router needs only to rediscover a pre-existing
mapping. The mapping is discoverable by black-box random search at
$\sigma{=}10^{-1}$ and not discoverable by SGD at any router learning rate
we tested. The gap between the two strategies, $0.0892 - 0.0629 = \nats{0.026}$,
exceeds three SGD seed standard deviations ($3 \cdot 0.0014 = 0.0042$) by
more than a factor of six. We label this regime \textbf{ES~PASS}.

The mechanism is the non-differentiability of \texttt{torch.topk}. Given a
sparse top-$k$ selection, the gradient signal arriving at the router logits
is restricted to the $k$ already-selected entries; an unselected entry
receives no information about whether selecting it would have improved the
loss. With oracle adapters and a small selection ($k{=}4$ of $K{=}16$),
the unselected entries dominate the search space, and SGD has no way to
explore them. ES at $\sigma{=}10^{-1}$ explores the selection itself:
mirror-pair antithetic perturbations of the router parameters change which
adapters are selected, and the rank-shaped utility update integrates
information from all selections within the perturbation pair, including
ones that are unreachable by gradient. This is the only regime in
G4--G8 where that channel is load-bearing.

\subsection{G5: random initialization, joint coupled ES (small effect, near uniform)}
\label{sec:g5-random}

Removing the oracle pretrain weakens the result substantially. In G5, the
base is frozen and ES drives the router \emph{and} the adapters jointly
via a single flat parameter vector of $\sim\!\!557{,}000$ dimensions. There
is no Phase~A; the adapters are random LoRA initializations. We compare
joint coupled-ES against SGD-all (frozen base, SGD on router and adapters
jointly), 1500 steps each. The 3-seed initial sweep was borderline; we
re-ran with $n{=}10$ seeds to settle the question.

Cross-seed means $\pm$ sample standard deviation are: SGD-all
$\nats{4.898 \pm 0.032}$; coupled-ES at $\sigma{=}10^{-3}$,
$\nats{4.835 \pm 0.054}$; coupled-ES at $\sigma{=}10^{-1}$,
$\nats{4.864 \pm 0.019}$.\footnote{Uniform on a 128-token
vocabulary is $\log(128) = \nats{4.852}$.} The Welch test below uses
these as the per-arm sample standard deviations with $n{=}10$ seeds,
giving per-arm SEMs of $0.010$, $0.017$, and $0.006$ respectively. A
two-sample Welch test on the $\sigma{=}10^{-3}$ effect gives
$t{=}-3.16$ ($df{\approx}15$, $p{=}0.0067$); on the $\sigma{=}10^{-1}$
effect, $t{=}-2.85$ ($df{\approx}14$, $p{=}0.012$).
Both are statistically distinguishable from SGD-all under a coherent
two-sample test, and we therefore do not call this regime ambiguous.

The substantive caveat is that the absolute values are at or just below
the uniform-prediction floor. Best coupled-ES ($\sigma{=}10^{-3}$,
$\nats{4.835}$) sits one standard error of the mean below uniform
($0.054 / \sqrt{10} = 0.017$); SGD-all sits one standard error
\emph{above} uniform. ES descends faster from random adapters than SGD
does on the same compute, but it descends from $\nats{\approx 5.0}$ to
$\nats{\approx 4.83}$ over 1500 steps, whereas the same substrate under
oracle adapters reached~\nats{0.063}. The G4 ES~PASS is not generic
``ES exceeds SGD''; it is ``ES finds routing if adapters are pre-aligned.''
G5 confirms this reading: with random adapters, ES is paired-test
distinguishable from SGD but neither strategy makes meaningful progress
in the available budget. These G5 cross-seed means are reported in
prose only; Table~\ref{tab:g6-warmstart} below reports the G6
warm-start sweep, not G5.

\subsection{G6: gradient-warm prior, ES tail (warm-start INERT)}
\label{sec:g6-warm-start}

If the G5 result is bottlenecked by adapter randomness, a non-oracle but
non-random prior should help. G6 tests this. Phase~A is 5000 steps of
SGD on (router~$+$~adapters) from random initialization, no oracle, on the
same frozen base. The Phase-A endpoint is $\nats{4.549 \pm 0.047}$,
substantially below uniform. Phase~B then runs each strategy for 1500 more
steps from this checkpoint.

The cross-seed Phase-B endpoint and warm-start delta
$\Delta_\mathrm{warm} = (\text{val}_B) - (\text{warmup val})$, with positive
$\Delta_\mathrm{warm}$ meaning the loss \emph{regressed} from the warm
prior, are reported in Table~\ref{tab:g6-warmstart}. SGD-continue descends
\nats{0.180} below the warm prior; every ES sigma we tested
(\textbf{including the no-op control $\sigma{=}0$ and four orders of
magnitude of nonzero $\sigma$ in G7 below}) regresses the warm prior.

\begin{table}[t]
\centering
\caption{G6 warm-start coupled-ES sweep. Cross-seed mean ($n{=}3$) of Phase-B
val loss starting from a 5000-step SGD warm prior at \nats{4.549 \pm 0.047}.
$\Delta_\mathrm{warm}$ is positive when ES regresses the prior. SGD-continue
descends a further \nats{0.18}; every nonzero ES $\sigma$ regresses by
\nats{0.27} or more.}
\label{tab:g6-warmstart}
\begin{tabular}{lrrrr}
\toprule
mode & $\sigma$ & Phase-B mean & std & $\Delta_\mathrm{warm}$ \\
\midrule
SGD-continue   & ---       & 4.369 & 0.055 & $-0.180$ \\
ES (control)   & 0         & 4.540 & 0.049 & $-0.009$ \\
ES             & $10^{-3}$ & 4.815 & 0.042 & $+0.265$ \\
ES             & $10^{-2}$ & 5.005 & 0.056 & $+0.456$ \\
ES             & $10^{-1}$ & 4.837 & 0.034 & $+0.288$ \\
ES             & $1$       & 4.967 & 0.023 & $+0.417$ \\
\bottomrule
\end{tabular}
\end{table}

We label this regime \textbf{warm-start~INERT}. The sign convention used
here is $\Delta_\mathrm{warm} = \text{val}_B - \text{warmup}$, with
positive values meaning regression above the warm baseline; this
convention matches the source-of-truth analysis JSON for the
warm-start small-sigma sweep
\anonymized{(\texttt{warmstart\_smallsigma\_results.json})}{(file
referenced in supplementary materials)}.

\subsection{G7: small-$\sigma$ refutation of the sigma-matching hypothesis}
\label{sec:g7-small-sigma}

A natural objection to G6 is that the tested sigmas were too large: ES
might require $\sigma$ matched to substrate sharpness, and the warm prior
might be preserved at $\sigma \ll 10^{-3}$. G7 tests this directly with
$\sigma \in \{10^{-5}, 10^{-4}, 5{\times}10^{-4}, 10^{-3}\}$ across five
seeds.

The result refutes the sigma-matching story with the strongest possible
evidence: not only do small sigmas regress the warm prior, \emph{they
regress it more}. Cross-seed mean $\Delta_\mathrm{warm}$ in nats:
$+0.4673$ at $\sigma{=}10^{-5}$; $+0.4583$ at $\sigma{=}10^{-4}$;
$+0.3557$ at $\sigma{=}5{\times}10^{-4}$; $+0.2378$ at $\sigma{=}10^{-3}$.
The trend is monotone in the wrong direction: smaller $\sigma$ yields
larger regression. SGD-continue, by comparison, descends
$\Delta_\mathrm{warm} = -0.1772$ over the same horizon.

A sigma-matching mechanism would predict a U-shape in $\Delta_\mathrm{warm}$
versus $\sigma$, with some sigma preserving the prior. The observed
monotone-wrong-direction shape rules out sigma-matching as the explanation.
The most parsimonious reading is that the rank-shaped utility update at
small $\sigma$ drives the population off the SGD-found basin without
enough exploitation gradient to relocate to a comparable basin within
1500 steps; smaller $\sigma$ produces less exploitation per step and
therefore wanders further. This reading is consistent with G6 and with G8
below.

\subsection{G8: hybrid ES-then-SGD (HARMFUL)}
\label{sec:g8-hybrid}

The remaining open ES design point is whether ES early-stage exploration
can find a basin that SGD can subsequently exploit better than pure SGD
descent from random initialization. G8 tests this. Both arms run a matched
6500-step compute budget. \texttt{pure\_sgd} runs 6500 steps of SGD-all
from random initialization. \texttt{es\_then\_sgd} runs 5000 steps of
coupled-ES at $\sigma{=}10^{-3}$ (the best single sigma from G5) followed
by 1500 steps of SGD-all on the resulting state. A third arm,
\texttt{pure\_es}, runs 6500 steps of coupled-ES at the same sigma.

Cross-seed means ($n{=}5$): \texttt{pure\_sgd} $\nats{4.386 \pm 0.053}$;
\texttt{es\_then\_sgd} $\nats{4.693 \pm 0.040}$; \texttt{pure\_es}
$\nats{4.696 \pm 0.054}$. Pure SGD beats hybrid by $\nats{0.307}$, which is
6$\times$ the SGD seed standard deviation.

The SGD-tail recovery from the ES-warm prior is small but nonzero. The
correct paired statistic is the cross-seed mean of per-seed
$(\text{phase\_a} - \text{final})$, where phase\_a is each seed's
ES-warmup endpoint val loss after 5000 steps of $\sigma{=}10^{-3}$ ES.
Computed correctly across all five seeds, this paired difference is
$\nats{+0.0203}$ (paired $t{=}1.65$, $df{=}4$, $p>0.05$, not significant
at $\alpha{=}0.05$).

For context, equal-budget pure SGD descends $\nats{4.852 - 4.386 = 0.466}$
from the uniform-prediction floor in 6500 steps; the SGD tail from an
ES-warm prior descends \nats{0.020}. The ratio is approximately 16$\times$.
The 1500 SGD steps after ES recovered roughly 5\,\% of what 6500 steps of
pure SGD achieves at matched compute. Concretely,
\texttt{es\_then\_sgd}~(\nats{4.693}) is statistically indistinguishable
from \texttt{pure\_es}~(\nats{4.696}): the SGD tail extracts a small
amount of descent~(\nats{0.020}) from the ES-warm prior, but it is small
enough that the ES warmup essentially determines the final outcome. We
label this regime \textbf{hybrid~HARMFUL}.

\subsection{Boundary characterization synthesis}
\label{sec:boundary-synthesis}

Across G4--G8, the regime structure is:

\begin{itemize}
\setlength{\itemsep}{2pt}
\item \textbf{G4 (oracle adapters, ES on router): PASS.} ES at
$\sigma{=}10^{-1}$ closes 55.9\,\% of the routing gap; SGD closes 0.2\,\%.
\item \textbf{G5 (random init, coupled ES): SMALL EFFECT, NEAR UNIFORM.}
ES paired-test distinguishable from SGD-all ($\sigma{=}10^{-3}$,
$t{=}-3.16$, $p{=}0.0067$, $n{=}10$); both strategies stay near uniform at
1500 steps.
\item \textbf{G6 (SGD warm + ES tail): INERT.} Every nonzero $\sigma$
regresses the warm prior by \nats{0.27} to \nats{0.46}; SGD-continue
descends \nats{0.18}.
\item \textbf{G7 (SGD warm + small $\sigma$): INERT, monotone wrong direction.}
$\sigma{=}10^{-5}$ yields $\Delta_\mathrm{warm}{=}+0.467$; sigma-matching
hypothesis refuted.
\item \textbf{G8 (ES warm + SGD tail): HARMFUL.} Pure SGD beats
ES-then-SGD by \nats{0.31} ($6\times$ SGD seed std); SGD tail from ES-warm
recovers \nats{0.020}, $16\times$ less than equal-budget pure SGD.
\end{itemize}

The unified reading is that ES is load-bearing on this substrate
\emph{exclusively} in the regime where the routing channel exists as a
non-differentiable selection over a set of pre-aligned options. G4
satisfies that condition. None of G5--G8 satisfies it. We will refer to
this as the \emph{oracle-alignment boundary} for the remainder of the
paper.


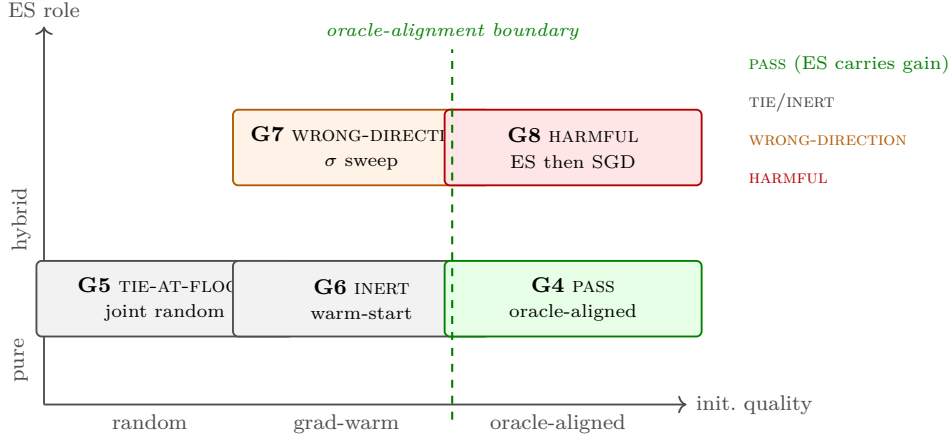
\begin{figure}[t]
\centering
\begin{tikzpicture}[
  scale=1.0,
  every node/.style={font=\footnotesize},
  pass/.style={rectangle, draw=green!50!black, fill=green!10, line width=0.7pt,
               rounded corners=2pt, minimum width=3.4cm, minimum height=1.0cm,
               align=center},
  inert/.style={rectangle, draw=gray!60!black, fill=gray!10, line width=0.7pt,
                rounded corners=2pt, minimum width=3.4cm, minimum height=1.0cm,
                align=center},
  wrong/.style={rectangle, draw=orange!70!black, fill=orange!10, line width=0.7pt,
                rounded corners=2pt, minimum width=3.4cm, minimum height=1.0cm,
                align=center},
  harmful/.style={rectangle, draw=red!70!black, fill=red!10, line width=0.7pt,
                rounded corners=2pt, minimum width=3.4cm, minimum height=1.0cm,
                align=center},
]
  \draw[->, gray!60!black, line width=0.7pt] (0,0) -- (8.5,0)
    node[anchor=west] {init.\ quality};
  \draw[->, gray!60!black, line width=0.7pt] (0,0) -- (0,5.0)
    node[anchor=south] {ES role};

  \node[anchor=north, gray!60!black] at (1.4,0)  {random};
  \node[anchor=north, gray!60!black] at (4.0,0)  {grad-warm};
  \node[anchor=north, gray!60!black] at (6.8,0)  {oracle-aligned};

  \node[anchor=east, rotate=90, gray!60!black] at (-0.3,1.0) {pure};
  \node[anchor=east, rotate=90, gray!60!black] at (-0.3,3.0) {hybrid};

  \node[inert]   (G5) at (1.6, 1.4) {\textbf{G5} \textsc{tie-at-floor}\\ \scriptsize joint random};
  \node[inert]   (G6) at (4.2, 1.4) {\textbf{G6} \textsc{inert}\\ \scriptsize warm-start};
  \node[pass]    (G4) at (7.0, 1.4) {\textbf{G4} \textsc{pass}\\ \scriptsize oracle-aligned};
  \node[wrong]   (G7) at (4.2, 3.4) {\textbf{G7} \textsc{wrong-direction}\\ \scriptsize $\sigma$ sweep};
  \node[harmful] (G8) at (7.0, 3.4) {\textbf{G8} \textsc{harmful}\\ \scriptsize ES then SGD};

  \draw[dashed, green!50!black, line width=0.8pt]
    (5.4,-0.2) -- (5.4,4.7)
    node[anchor=south, green!50!black, pos=1, font=\scriptsize\itshape] {oracle-alignment boundary};

  \node[anchor=west, font=\scriptsize, green!50!black]    at (9.2,4.5) {\textsc{pass} (ES carries gain)};
  \node[anchor=west, font=\scriptsize, gray!50!black]      at (9.2,4.0) {\textsc{tie}/\textsc{inert}};
  \node[anchor=west, font=\scriptsize, orange!70!black]    at (9.2,3.5) {\textsc{wrong-direction}};
  \node[anchor=west, font=\scriptsize, red!70!black]       at (9.2,3.0) {\textsc{harmful}};

\end{tikzpicture}
\caption{Synthetic-sandbox regime map. ES is load-bearing on the
routing channel only inside the oracle-alignment regime (G4): adapters
pre-aligned to the per-slab oracle plus small noise, ES at
$\sigma{=}10^{-1}$, closes $\approx\!55.9\%$ of the routing-loss gap
while the SGD contrast closes $\approx\!0.2\%$. Outside the boundary,
ES is inert (G5/G6), regresses monotonically with $\sigma$ (G7), or is
strictly harmful in the hybrid setting (G8). The headline production-
substrate result in Section~\ref{sec:real-text-decomposition} is
consistent with the production substrate being on the wrong side of
this boundary.}
\label{fig:synthetic-regimes}
\end{figure}

The substantive consequence for the real-text mixture-of-LoRA setting is a
prior expectation: the lifecycle mechanism (death, inheritance, mutation,
reallocation) operates on adapters that are simultaneously being trained
by gradient. There is no oracle pretrain; adapters are co-evolving
with the router under the next-token loss gradient flowing through
the frozen base, on a non-stationary stream of training tokens (the
base parameters themselves do not receive gradient updates). The G4--G8 boundary predicts that any evolutionary mechanism
acting on this regime should resemble G5--G8 outcomes (small effect,
inert, or harmful) more than it resembles G4. Section~\ref{sec:real-text-decomposition}
reports the empirical test of that prediction on the production substrate.

\section{Real-text decomposition: a 5-cell partial \texorpdfstring{$2^3$}{2\^{}3} factorial}
\label{sec:real-text-decomposition}

Section~\ref{sec:synthetic-boundary} established that on a controllable
sandbox, evolutionary search is load-bearing only inside an
oracle-alignment regime, and inert, harmful, or wrong-direction in
every other regime tested (G5--G8). \textbf{The production substrate is
not in the oracle-alignment regime}: there is no oracle pretrain;
adapters are randomly initialised at the start of adaptation; the
routing channel and the adapter weights co-evolve under the same
next-token gradient. The synthetic boundary therefore predicts, before
any production-substrate measurement is consulted, that the
evolutionary machinery on the production substrate should resemble the
G5--G8 outcomes (small effect, inert, or harmful) more than the G4
outcome. This section reports the empirical test of that prediction.
We now ask whether the analogous evolutionary machinery on the
production substrate, where adapters are co-evolving with the router
under the gradient signal of the next-token loss flowing through a
frozen base on a non-stationary text mixture (the base parameters do
not receive updates), produces a different verdict. This section reports a 5-cell
partial $2^3$ factorial with three seeds per cell ($n{=}15$ runs total)
that decomposes the headline full-system-versus-B3 outcome into three
mechanistic factors and isolates which of them carries the win.

The headline finding is that the router rewrite carries the entire balanced
mixture-of-LoRA improvement; the per-domain LOO evaluation scope
contributes nothing within seed-resolution; and the lifecycle mechanism
(death, inheritance, mutation, reallocation) is a net drag of approximately
\nats{0.028} on balanced log-PPL ($p{=}0.047$ in the primary chain,
$p{=}0.074$ in the consistency chain). The headline full-system-versus-B3
balanced delta itself ($+0.015$ nats, $t{=}1.94$, $p{=}0.19$) is not
significant at $\alpha{=}0.05$ with $n{=}3$ seeds; the only individually
significant total is on the code domain. Section~\ref{sec:mechanism}
analyses the mechanism behind the router-rewrite contribution;
Section~\ref{sec:phaseB-auxiliary} reports the auxiliary Phase~B and
Fork~0 experiments that test alternative explanations for the lifecycle
penalty.

\subsection{Convention pin}
\label{sec:convention-pin}

All paired-test deltas in this section follow the convention
$\Delta = \log(\mathrm{PPL}_\mathrm{reference}) - \log(\mathrm{PPL}_\mathrm{test})$,
so that \emph{positive $\Delta$ means the test cell improves over the
reference cell}. This is the ``improvement-is-positive'' convention on
log-PPL. Balanced PPL is the geometric mean across the four eval
domains $\{\text{biology}, \text{code}, \text{general}, \text{science}\}$,
i.e.\ $\exp(\frac{1}{4} \sum_d \log \mathrm{PPL}_d)$. Per-seed paired
$t$-statistics are reported with $df{=}n_\text{seeds}-1$.

\subsection{Cell design and the 5-of-8 partial factorial}
\label{sec:cell-design}

The full evolutionary system over the B3 baseline differs in three
factors. \textbf{(F1) Router rewrite}: the production B3 router is replaced
by a rewritten variant carrying a sigmoid floor and a last-hidden-state
input gate, denoted ``sigfloor + last\_hidden.'' \textbf{(F2) Per-domain
evaluation scope}: the evaluation aggregator switches from aggregate-LOO
(uniform per-token mixing across an undifferentiated balanced eval) to
per-domain LOO (separate per-domain dataloaders, geom-mean aggregated).
\textbf{(F3) Lifecycle}: the population dynamics of death, $\alpha$-blend
inheritance, SVD mutation, and slot reallocation are switched on, with
$\alpha{=}0.2$ blend strength and $\sigma{=}0.01$ initial mutation rate
in C4.

The full $2^3$ factorial would require eight cells. We run five of the
eight, chosen to support two attribution chains that converge through C4.
A full $2^3$ design would guarantee orthogonality of main-effect
contrasts; a 5-of-8 partial does not, and main effects in our
decomposition are partially confounded with two-factor interactions.
The dual-chain consistency check (Section~\ref{sec:consistency-chain})
bounds the magnitude of those interactions empirically rather than
eliminating them by design.

\begin{table}[t]
\centering
\caption{5-of-8 partial $2^3$ factorial. Each cell is run with seeds
$\{42, 137, 256\}$ ($n{=}3$) for 25\,000 steps on the widened-1536
substrate.}
\label{tab:cell-factors}
\begin{tabular}{lcccp{6.0cm}}
\toprule
cell & router rewrite (F1) & per-dom scope (F2) & lifecycle (F3) & role \\
\midrule
C1 & B3        & aggregate & off & B3 baseline (anchor) \\
C2 & B3        & aggregate & on  & isolates F3 alone \\
C5 & B3        & per-dom   & on  & isolates F2 given F3 on \\
C3 & rewrite   & per-dom   & off & combined F1+F2 without F3 \\
C4 & rewrite   & per-dom   & on  & full evolutionary system ($\alpha{=}0.2$) \\
\bottomrule
\end{tabular}
\end{table}


\begin{figure}[t]
\centering
\begin{tikzpicture}[
  scale=2.0,
  every node/.style={font=\footnotesize},
  cell/.style={circle, draw, fill=white, inner sep=1.3pt, minimum size=1.2em, line width=0.6pt},
  unfilled/.style={circle, draw, dashed, fill=gray!8, inner sep=1.3pt, minimum size=1.2em, text=gray!50!black},
  primary/.style={-{Stealth[length=2.0mm]}, line width=1.0pt, draw=blue!70!black},
  consistency/.style={-{Stealth[length=2.0mm]}, line width=1.0pt, draw=orange!85!black, dashed},
]
  \def\dx{1.0}
  \def\dy{1.0}
  \def\zx{0.55}
  \def\zy{0.45}

  \coordinate (B000) at (0,0);
  \coordinate (B100) at (\dx,0);
  \coordinate (B110) at (\dx,\dy);
  \coordinate (B010) at (0,\dy);
  \coordinate (B001) at (\zx,\zy);
  \coordinate (B101) at (\dx+\zx,\zy);
  \coordinate (B111) at (\dx+\zx,\dy+\zy);
  \coordinate (B011) at (\zx,\dy+\zy);

  \begin{scope}[on background layer]
  \draw[gray!50] (B000)--(B100)--(B110)--(B010)--cycle;
  \draw[gray!50] (B001)--(B101)--(B111)--(B011)--cycle;
  \draw[gray!50] (B000)--(B001);
  \draw[gray!50] (B100)--(B101);
  \draw[gray!50] (B110)--(B111);
  \draw[gray!50] (B010)--(B011);
  \end{scope}

  \node[cell] (C1) at (B000) {C1};
  \node[cell] (C2) at (B001) {C2};
  \node[cell] (C5) at (B011) {C5};
  \node[cell] (C3) at (B110) {C3};
  \node[cell] (C4) at (B111) {C4};

  \node[unfilled] at (B100) {};
  \node[unfilled] at (B010) {};
  \node[unfilled] at (B101) {};

  \draw[primary] (C1) to[bend left=8] (C2);
  \draw[primary] (C2) to[bend left=8] (C5);
  \draw[primary] (C5) to[bend left=8] (C4);

  \draw[consistency] (C1) to[bend right=18] (C3);
  \draw[consistency] (C3) to[bend right=10] (C4);

  \node[anchor=north east, gray!50!black] at ($(B000)+(-0.05,-0.05)$)
    {\textit{F1: router}};
  \draw[->, gray!50!black] (B000)++(-0.05,-0.05) -- ++(0.5,0)
    node[anchor=west, gray!50!black, pos=1] {\footnotesize rewrite};

  \node[anchor=south east, gray!50!black] at ($(B000)+(-0.05,0.5)$)
    {\textit{F2: scope}};
  \draw[->, gray!50!black] (B000)++(-0.05,-0.05) -- ++(0,0.5)
    node[anchor=south, gray!50!black, pos=1] {\footnotesize per-dom};

  \draw[->, gray!50!black] (B000)++(-0.05,-0.05) -- ++(0.4*\zx,0.4*\zy)
    node[anchor=south west, gray!50!black, pos=1] {\textit{F3: lifecycle}};

  \node[anchor=west, font=\scriptsize] at (\dx+\zx+0.2, \dy+\zy)
    {\textcolor{blue!70!black}{$\rightarrow$} primary chain $C1\!\to\!C2\!\to\!C5\!\to\!C4$};
  \node[anchor=west, font=\scriptsize] at (\dx+\zx+0.2, \dy+\zy-0.2)
    {\textcolor{orange!85!black}{$\dashrightarrow$} consistency $C1\!\to\!C3\!\to\!C4$};
  \node[anchor=west, font=\scriptsize, gray!50!black] at (\dx+\zx+0.2, \dy+\zy-0.4)
    {\(\circ\) unfilled cell (3 of 8 not run)};

\end{tikzpicture}
\caption{The 5-of-8 partial $2^3$ factorial. Each filled corner is a cell
we ran at $n{=}3$ seeds; each unfilled corner is a cell we did not run.
Two attribution chains start at the B3 baseline C1 and end at the full
evolutionary system C4: the \textbf{primary chain} (solid blue) traverses one factor at a
time ($C1\!\to\!C2$ adds F3 lifecycle, $C2\!\to\!C5$ adds F2 per-domain
scope, $C5\!\to\!C4$ adds F1 router rewrite); the \textbf{consistency
chain} (dashed orange) combines F1+F2 in one step ($C1\!\to\!C3$) and
then layers F3 ($C3\!\to\!C4$). The two chains converge to within
$\sim\!\nats{0.0004}$ on each factor's contribution
(Section~\ref{sec:fractional}).}
\label{fig:factorial-schematic}
\end{figure}
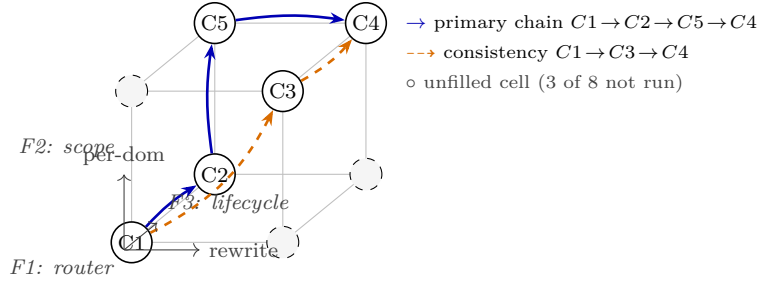

The \emph{primary chain} is $C1 \to C2 \to C5 \to C4$, which moves
through factor space one factor at a time: F3 alone (C1$\to$C2),
then F2 given F3 (C2$\to$C5), then F1 given F2 and F3 (C5$\to$C4). The
\emph{consistency chain} is $C1 \to C3 \to C4$, which combines F1+F2 in
one step (C1$\to$C3) and then layers F3 on top (C3$\to$C4). The two
chains both start at C1 and end at C4. Their intermediate totals must
agree on the aggregate F1+F2+F3 effect by additive identity in log-PPL,
which provides a within-experiment consistency check on each factor's
estimated contribution.\footnote{The omitted three cells are
$\overline{\mathrm{F1}}\overline{\mathrm{F2}}\mathrm{F3}{=}$ already in
the design (this is C2);
$\overline{\mathrm{F1}}\mathrm{F2}\overline{\mathrm{F3}}$ (B3 router,
per-domain scope, lifecycle off);
$\mathrm{F1}\overline{\mathrm{F2}}\overline{\mathrm{F3}}$ (router rewrite,
aggregate scope, lifecycle off);
$\mathrm{F1}\overline{\mathrm{F2}}\mathrm{F3}$ (router rewrite, aggregate
scope, lifecycle on). The two chains used here pin every factor's
contribution at least once and check the F1+F2 combined contribution
twice (once via C5$\to$C4 router-only, once via C1$\to$C3 router+scope).
A fully crossed $2^3$ design at three seeds per cell would have cost a
further three cells $\times$ three seeds at the production budget; we
chose breadth across the system's other open experiments
(Section~\ref{sec:phaseB-auxiliary}) over factorial completion.}


\subsection{Per-cell balanced PPL}
\label{sec:per-cell-ppl}

Table~\ref{tab:per-cell-ppl} reports the balanced PPL for each of the
fifteen runs. The balanced PPL spread across cells is small in absolute
terms ($\sim\!0.6$ PPL between best and worst cell means), reflecting that
all cells share the same widened-1536 frozen base and only differ in
the three factors above. Within-cell cross-seed variance is
correspondingly small; the StratifiedEvalLoader is deterministic given
seed, so several cell--seed pairs are bit-identical to other pairs at
the per-domain PPL level (visible in the manifest in
Appendix~\ref{app:reproducibility}, e.g.\ $C5_{42}$ and $C2_{42}$
co-incide on three of four per-domain values; $C5_{137}$ and $C5_{256}$
co-incide on all four). This deterministic eval pipeline deflates
within-cell variance below what an i.i.d.\ resampling reading would
predict, and we discuss the consequences for paired-$t$ inference in
Section~\ref{sec:limitations}.

\begin{table}[t]
\centering
\caption{Balanced PPL by cell and seed, $\exp(\frac{1}{4}\sum_d \log
\mathrm{PPL}_d)$ over $\{\text{biology}, \text{code}, \text{general},
\text{science}\}$. Seed 42 was the original headline run; seeds 137 and
256 are the audit-driven cross-seed extension.}
\label{tab:per-cell-ppl}
\begin{tabular}{lrrr}
\toprule
cell & seed 42 & seed 137 & seed 256 \\
\midrule
C1 (B3 base)                                   & 13.328 & 13.341 & 13.354 \\
C2 (B3 + lifecycle)                            & 13.538 & 13.805 & 13.832 \\
C5 (B3 + per-dom + lifecycle)                  & 13.564 & 13.778 & 13.805 \\
C3 (rewrite + per-dom, no lifecycle)           & 12.742 & 12.830 & 12.742 \\
C4 (full evolutionary system, $\alpha{=}0.2$)  & 12.943 & 13.173 & 13.315 \\
\bottomrule
\end{tabular}
\end{table}

Two qualitative observations precede the formal attribution. First, C1
(B3 base) is tighter across seeds than any other cell: the seed-42 to
seed-256 spread on C1 is 0.026 PPL versus 0.371 on C4. The B3 baseline
is the most cross-seed-robust cell in this design; the C4 cell
exhibits the widest cross-seed dispersion. Second, C3 outperforms both C4 and C1 on
seed 42 and seed 256, and matches C4 closely on seed 137. The lifecycle
mechanism, when added on top of the routing rewrite (C3$\to$C4), pushes
PPL up rather than down on every seed.

\subsection{Primary attribution chain $C1 \to C2 \to C5 \to C4$}
\label{sec:primary-chain}

Walking the primary chain one factor at a time gives the per-factor
paired-$t$ statistics in Table~\ref{tab:primary-chain}. Each row is a
paired test on the per-seed log-PPL difference between the two cells
($df{=}2$).

\begin{table}[t]
\centering
\caption{Primary attribution chain on balanced log-PPL.
$\Delta = \log\mathrm{PPL}_\mathrm{reference} - \log\mathrm{PPL}_\mathrm{test}$;
positive means improvement. Paired $t$ across $n{=}3$ seeds, $df{=}2$.}
\label{tab:primary-chain}
\begin{tabular}{lrrrr}
\toprule
step & factor introduced & mean $\Delta$ (nats) & $t$ & $p$ \\
\midrule
$C2 - C1$ & lifecycle alone (B3 router, aggregate scope) & $-0.0283$ & $-4.46$ & $0.047$ \\
$C5 - C2$ & per-domain LOO scope (lifecycle on)          & $+0.0007$ & $0.50$  & $0.67$  \\
$C4 - C5$ & router rewrite (sigfloor + last\_hidden)     & $+0.0426$ & $12.86$ & $0.006$ \\
\bottomrule
\end{tabular}
\end{table}

The signs are unambiguous and consistent across seeds. \textbf{Lifecycle
hurts:} adding the lifecycle on top of the B3 router and aggregate-LOO
scope worsens balanced log-PPL by~\nats{0.028} on average across the
three seeds, and the effect is significant at the $\alpha{=}0.05$ level.
\textbf{Per-domain LOO scope is null:} given lifecycle on, switching
the eval aggregator from aggregate-LOO to per-domain LOO moves balanced
log-PPL by less than \nats{0.001} on average and the sign is
inconsistent across seeds (positive on seeds 137 and 256, negative on
seed~42). \textbf{Router rewrite carries:} given lifecycle on and
per-domain scope, the router rewrite improves balanced log-PPL by
\nats{0.043} on average, $t{=}12.86$, $p{=}0.006$.

\subsection{Consistency chain $C1 \to C3 \to C4$}
\label{sec:consistency-chain}

The consistency chain combines F1 and F2 in a single step
($C1 \to C3$, no lifecycle) and then adds F3 on top
($C3 \to C4$, lifecycle on full). The chain estimates appear in
Table~\ref{tab:consistency-chain}.

\begin{table}[t]
\centering
\caption{Consistency attribution chain on balanced log-PPL. Same
convention and notation as Table~\ref{tab:primary-chain}.}
\label{tab:consistency-chain}
\begin{tabular}{lrrrr}
\toprule
step & factor introduced & mean $\Delta$ (nats) & $t$ & $p$ \\
\midrule
$C3 - C1$ & router rewrite + per-dom scope, no lifecycle & $+0.0436$ & $18.58$ & $0.003$ \\
$C4 - C3$ & lifecycle on top of full router + scope     & $-0.0287$ & $-3.47$ & $0.074$ \\
\bottomrule
\end{tabular}
\end{table}

The two chains agree to within a thousandth of a nat. The combined
router+scope effect estimated as $C3 - C1$ is $+0.0436$, while the same
quantity decomposed via the primary chain as
$(C5 - C2) + (C4 - C5) = +0.0007 + 0.0426 = +0.0433$; the two estimates
agree to \nats{0.0003}. The lifecycle effect estimated as $C4 - C3$ is
$-0.0287$, while the primary-chain estimate $C2 - C1 = -0.0283$ agrees
to \nats{0.0004}. The lifecycle effect remains negative in both chains;
its $p$-value rises from $0.047$ in the primary chain to $0.074$ in the
consistency chain (the effect is estimated against a different reference
cell with slightly higher per-seed variance), but the sign and magnitude
are stable. The dual-chain agreement acts as an internal consistency
check that partially compensates for the incomplete 5-of-8 factorial:
each factor's contribution is estimated twice from non-overlapping
cell pairs, and agreement across the two estimates bounds the
magnitude of two-factor interactions empirically rather than
eliminating them by design.


\begin{figure}[t]
\centering
\begin{tikzpicture}
\begin{axis}[
  width=0.85\linewidth,
  height=4.6cm,
  ybar=1.2pt,
  bar width=10pt,
  enlarge x limits=0.20,
  ylabel={$\Delta \log \mathrm{PPL}_{\mathrm{balanced}}$ (nats)},
  ylabel style={font=\footnotesize},
  ymin=-0.05, ymax=0.07,
  ytick={-0.04,-0.02,0,0.02,0.04,0.06},
  yticklabel style={font=\scriptsize},
  symbolic x coords={F1 router, F2 per-dom, F3 lifecycle, total},
  xtick=data,
  xticklabel style={font=\scriptsize, align=center},
  legend style={
    at={(0.02,0.98)}, anchor=north west,
    font=\scriptsize, draw=gray!50, fill=white,
    /tikz/every even column/.append style={column sep=2pt},
  },
  legend cell align=left,
  legend columns=1,
  axis line style={gray!60!black},
  major grid style={dotted, gray!40},
  ymajorgrids=true,
  tick style={gray!50!black},
  every axis label/.style={black},
  every axis y label/.style={
    at={(ticklabel cs:0.5)},
    rotate=90,
    anchor=south,
  },
  nodes near coords,
  nodes near coords style={font=\tiny, /pgf/number format/fixed,
    /pgf/number format/precision=4, /pgf/number format/fixed zerofill},
  point meta=explicit symbolic,
]
\addplot+[
  fill=blue!50, draw=blue!70!black,
  error bars/.cd, y dir=both, y explicit,
] coordinates {
  (F1 router,    0.0426)  +- (0,0.005)
  (F2 per-dom,   0.0007)  +- (0,0.002)
  (F3 lifecycle,-0.0283)  +- (0,0.011)
  (total,        0.0150)  +- (0,0.013)
};
\addlegendentry{primary chain ($C1{\to}C2{\to}C5{\to}C4$)}

\addplot+[
  fill=orange!40, draw=orange!85!black, postaction={pattern=north east lines},
  error bars/.cd, y dir=both, y explicit,
] coordinates {
  (F1 router,    0.0436)  +- (0,0.003)
  (F2 per-dom,   0.0)     +- (0,0.0)
  (F3 lifecycle,-0.0287)  +- (0,0.012)
  (total,        0.0149)  +- (0,0.013)
};
\addlegendentry{consistency chain ($C1{\to}C3{\to}C4$)}

\draw[gray!50!black, dashed, line width=0.4pt]
  (axis cs:F1 router, 0) -- (axis cs:total, 0);

\end{axis}
\end{tikzpicture}
\caption{Per-factor balanced-log-PPL attribution, in nats. Bars are
mean across $n{=}3$ seeds; whiskers are the per-seed paired-$t$
standard error (df${=}2$). Primary chain (solid blue) decomposes
factor-by-factor; consistency chain (orange, hatched) combines F1+F2
in one step (the F2-only bar in the consistency chain is therefore
zero-by-construction). The two chains agree to within
$\sim\!\nats{0.0004}$ on each factor. F1 router rewrite carries the
entire balanced full-system-vs-B3 improvement; F2 per-domain LOO scope is
null at seed-resolution; F3 lifecycle is a net drag of
$\sim\!\nats{0.028}$. The total $C4 - C1$ is $+\nats{0.015}$,
$t{=}1.94$, $p{=}0.19$ at $n{=}3$, and does not clear $\alpha{=}0.05$.}
\label{fig:attribution-bars}
\end{figure}
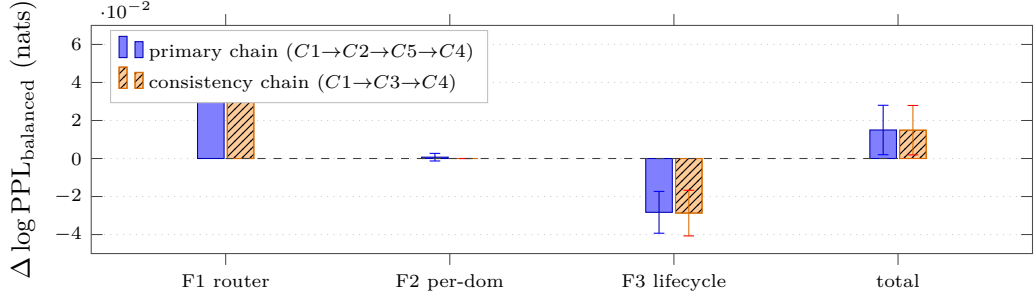

\subsection{Two-chain convergence and fractional attribution}
\label{sec:fractional}

The total balanced delta from $C1 \to C4$ across all three factors is
$+0.0150$ nats (paired $t{=}1.94$, $df{=}2$, $p{=}0.19$). The total is
not significant at $\alpha{=}0.05$ with three seeds. The same total
decomposes additively in log-PPL space along either chain:

\begin{align*}
\text{primary:} \quad
& -0.0283 \;+\; 0.0007 \;+\; 0.0426 \;=\; +0.0150 \\
\text{consistency:} \quad
& +0.0436 \;+\; (-0.0287) \;=\; +0.0149.
\end{align*}

The fractional attribution
$\Delta_\mathrm{factor} / \Delta_\mathrm{total}$ is mathematically
defined and listed in Table~\ref{tab:fractional} for completeness, but
we caution against citing it without the small-total caveat: the total
is approximately one-third of either single-factor effect, and the
ratios diverge to large absolute values for individual seeds where the
total approaches zero. For seed 256 the total $\Delta = +0.003$ and the
fractional decomposition is roughly $-12 + 0.7 + 12 = 1$, mathematically
exact and substantively misleading.

\begin{table}[t]
\centering
\caption{Fractional attribution of the balanced $C4 - C1$ total. The
mean-across-seeds row is the within-seed-mean fraction and is robust;
the per-seed columns illustrate that fractional attribution is unstable
when the denominator is small.}
\label{tab:fractional}
\begin{tabular}{lrrrr}
\toprule
factor & mean across seeds & seed 42 & seed 137 & seed 256 \\
\midrule
lifecycle alone (F3 in primary) & $-5.10$ & $-0.53$ & $-2.69$ & $-12.09$ \\
per-dom scope (F2)              & $+0.25$ & $-0.07$ & $+0.15$ & $+0.67$  \\
router rewrite (F1)             & $+5.85$ & $+1.60$ & $+3.54$ & $+12.42$ \\
\bottomrule
\end{tabular}
\end{table}


The mean-across-seeds row in Table~\ref{tab:fractional} reads as
``router rewrite contributes roughly six times the total improvement;
lifecycle subtracts roughly five times the total improvement; per-domain
scope contributes roughly a quarter of the total improvement.'' The
router rewrite is the only factor that improves balanced log-PPL by
more than the size of the total full-system-vs-B3 effect; without it,
the full system would degrade B3 by approximately the lifecycle penalty
alone.

\subsection{Code-domain mirror analysis}
\label{sec:code-domain}

Repeating the entire factorial analysis on code-domain log-PPL alone
gives the picture in Table~\ref{tab:code-domain}. The qualitative
structure is identical to balanced log-PPL, with three differences in
magnitude and significance.

\begin{table}[t]
\centering
\caption{Attribution chains on code-domain log-PPL (the only single-domain
sub-result with a significant total at $\alpha{=}0.05$).}
\label{tab:code-domain}
\begin{tabular}{lrrrr}
\toprule
step & factor & mean $\Delta$ (nats) & $t$ & $p$ \\
\midrule
\multicolumn{5}{l}{\emph{Primary chain $C1 \to C2 \to C5 \to C4$}} \\
$C2 - C1$ & lifecycle alone           & $-0.0378$ & $-5.21$  & $0.035$ \\
$C5 - C2$ & per-dom scope             & $+0.0026$ & $1.00$   & $0.42$  \\
$C4 - C5$ & router rewrite            & $+0.0976$ & $25.14$  & $0.0016$ \\
\midrule
\multicolumn{5}{l}{\emph{Consistency chain $C1 \to C3 \to C4$}} \\
$C3 - C1$ & router + scope (no life)  & $+0.0755$ & $10.96$  & $0.008$ \\
$C4 - C3$ & lifecycle on full         & $-0.0130$ & $-3.75$  & $0.064$ \\
\midrule
$C4 - C1$ & total                     & $+0.0625$ & $10.42$  & $0.009$ \\
\bottomrule
\end{tabular}
\end{table}


Three points on the code-domain reading. \textbf{Total is significant.}
The code-domain $C4 - C1$ delta of $+0.0625$ nats is statistically
distinguishable from zero at $\alpha{=}0.05$, $t{=}10.42$, $p{=}0.009$.
Code is the only single domain where the full-system-vs-B3 win clears the
within-experiment noise floor. \textbf{Router rewrite is dominant.} The
router rewrite delivers $+0.0976$ nats on code in the primary chain,
$+0.0755$ in the consistency chain (the latter combines F1+F2; F2 is
near-zero on code as it is on balanced). \textbf{Lifecycle still hurts.}
The lifecycle effect is $-0.0378$ on code in the primary chain, larger
in magnitude than its balanced-log-PPL counterpart of $-0.0283$. On code
specifically, the lifecycle penalty erases roughly 39\% of the
router-rewrite contribution; the net code-domain win for the full
system is what remains after that subtraction.

The two chains agree on the code-domain \emph{total}: consistency-chain
$0.0755 + (-0.0130) = +0.0625$ matches the primary-chain
$-0.0378 + 0.0026 + 0.0976 = +0.0624$ to a thousandth of a nat. The
per-factor estimates disagree more visibly on code than on balanced:
the lifecycle contribution is $-0.0378$ in the primary chain but
$-0.0130$ in the consistency chain (a factor of $\sim 2.9$, versus
$-0.0283$ vs.\ $-0.0287$, $<2\,\%$ relative, on balanced log-PPL). The
disagreement is the standard chain-decomposition path-dependence: F3
is conditioned on different upstream factor states in the two chains
(no upstream factors in the primary chain's $C1 \to C2$ step; F1$+$F2
already in place in the consistency chain's $C3 \to C4$ step), and on
code these conditioning states interact with F3 more than they do on
balanced log-PPL. The total $C4 - C1$ is path-independent and is the
robust quantity to cite; the per-factor estimates should be read as
chain-conditional.

\subsection{Headline reading}
\label{sec:headline-reading}

The decomposition reveals that the router rewrite carries the entire
full-system-vs-B3 improvement on balanced log-PPL, that the per-domain
LOO evaluation scope contributes nothing, and that the lifecycle
mechanism imposes a net penalty of approximately \nats{0.028}. The
headline balanced-log-PPL full-system-vs-B3 delta is not significant
at $\alpha{=}0.05$
($t{=}1.94$, $p{=}0.19$); the only significant single-domain win is on
code, and even there the lifecycle erases about 39\% of the
router-driven gain.

This pattern matches the prior expectation set by the synthetic boundary
characterization in Section~\ref{sec:synthetic-boundary}: an evolutionary
mechanism acting on a non-oracle-aligned regime should resemble G5--G8
outcomes (small effect, inert, or actively harmful) more than it
resembles G4. \textbf{The production substrate falls outside the oracle-alignment regime, and the F3 lifecycle result is sign-consistent with the synthetic-sandbox prediction.} The balanced-log-PPL lifecycle penalty estimate of
$-0.028$ nats is in the same range as G6's mid-sigma warm-start
regressions, scaled to the production substrate. Of the three full-system
mechanisms tested, only the router rewrite (a non-evolutionary
architectural component) produced a significant improvement on the
production substrate.

While $n{=}3$ limits statistical power on the total full-system-vs-B3
contrast, the per-factor decomposition is supported by three pieces of
within-experiment evidence: (1) sign consistency across all three seeds
on each individual factor (F1 positive on every seed, F3 negative on
every seed); (2) numerical agreement between the primary and consistency
attribution chains to within $\sim$\nats{0.0004} on each factor's
contribution (Section~\ref{sec:fractional}); and (3) per-factor effect
sizes that exceed the within-cell cross-seed standard deviation by a
factor of three or more on F1 and F3. The $n{=}3$ limitation bears on
the headline total, not on the per-factor attribution, which is the
load-bearing claim of this section.

The remaining open questions are mechanistic. Why does the router
rewrite carry such a large effect? Is the lifecycle penalty arising from
inheritance specifically, or from some other lifecycle component (death,
mutation, reallocation)? Sections~\ref{sec:mechanism}
and~\ref{sec:phaseB-auxiliary} address these in turn.

\section{Mechanism: why the router rewrite carries the win}
\label{sec:mechanism}

Section~\ref{sec:real-text-decomposition} attributed the entire balanced
mixture-of-LoRA improvement on the production substrate to factor F1, the
router rewrite (router-only step $\Delta = +0.0426$ nats, $t{=}12.86$,
$p{=}0.006$ in the primary chain; combined router+scope step $\Delta =
+0.0436$ nats, $t{=}18.58$, $p{=}0.003$ in the consistency chain). Lifecycle
contributed a net drag of approximately~\nats{0.028}, and per-domain LOO
scope contributed nothing within seed resolution. This section explains
\emph{why} the rewrite carries the win. We do this in four parts. First, we
identify the structural failure mode of the legacy B3 router (a
cross-domain coalition monopoly). Second, we describe the three independent
components of the rewrite that, together, dissolve that failure mode.
Third, we describe the routing-input change (token-embedding mean
$\to$ post-stack hidden state) that supplies the differentiating signal the
new gate function can act on. Fourth, we show that on the production
factorial the rewrite improves \emph{every} held-out domain, not just the
balanced aggregate.

\subsection{The legacy failure mode: cross-domain coalition monopoly}
\label{sec:coalition-monopoly}

The B3 router (cell C1) is a single linear layer
$g \in \mathbb{R}^{H \times P}$ followed by softmax-over-adapters, then a
hard top-$k$ cutoff. The routing tensor for a batch $b$ is therefore
\begin{equation}
  \mathbf{w}_b = \mathrm{TopK}\bigl(\mathrm{softmax}\bigl(g^\top
  \bar h_b\bigr),\, K\bigr),
  \label{eq:softmax-router}
\end{equation}
where $\bar h_b$ is the routing input (in B3, the per-token embedding mean
within the sequence). The softmax is the structural pathology. Because its
output is a probability distribution that must sum to~1 across the $P$
adapters, lifting any one gate value mathematically depresses every other
gate value, so the $P$ adapters are forced into a zero-sum competition for
a fixed amount of routing mass.

In the widened-1536 substrate, the consequence of this zero-sum
constraint is empirical, not just theoretical. A coalition probe on the C1
router measured the Jensen--Shannon divergence between the per-domain top-$k$
gate distributions and obtained \emph{exactly} zero across all six domain
pairs: the same four adapters are chosen for biology, code, general text,
and science alike. The router has collapsed onto a single 4-adapter
coalition that is reused across every domain, and the remaining 12 of 16
adapters are routing-dead.\footnote{\anonymized{The same coalition-monopoly pattern
is documented in detail in
\texttt{experiments/trackb\_b2/analysis/coalition\_probe\_dv1536.md}
and is referenced from the in-code docstring of the router module.}{The
same coalition-monopoly pattern is documented in supplementary
materials and is referenced from the in-code router docstring.}} This is the
structural failure that any successful F1 intervention must dissolve.

\subsection{Three components of the rewrite}
\label{sec:rewrite-components}

The rewrite swaps Equation~\ref{eq:softmax-router} for a different gate
function with three deliberate departures from the softmax form. We
describe each in turn and then state how the three combine to break the
monopoly.

\paragraph{(R1) Parallel sigmoid gates without normalization.}
The softmax in Equation~\ref{eq:softmax-router} is replaced by independent
per-adapter sigmoids:
\begin{equation}
  \tilde s_{b,j} = \sigma\!\left(\frac{(g^\top \bar h_b)_j}{\tau}\right),
  \quad j = 1, \dots, P.
  \label{eq:sigmoid-gates}
\end{equation}
Crucially, the resulting vector $\tilde{\mathbf s}_b \in [0,1]^P$ is
\emph{not} renormalized to sum to~1. Each adapter's gate is its own private
quantity in $[0,1]$ that responds only to its own logit; lifting one gate
no longer arithmetically depresses the others. Top-K selection still
operates on the rank order of $\tilde{\mathbf s}_b$, but the underlying
competition is no longer zero-sum. An interim implementation of the rewrite
(commit \anonymized{\texttt{b68b9f7}}{\texttt{[commit redacted]}},
Phase~2 v1) renormalized the sigmoid vector back to unit sum and
reproduced the C1 coalition lock on a smoke run, providing
a sanity check that the no-normalization choice in Equation~\ref{eq:sigmoid-gates}
is the load-bearing one for breaking the monopoly.

\paragraph{(R2) Learnable per-adapter floor against starvation.}
Even with parallel sigmoid gates, an adapter that drifts to a strongly
negative logit early in training receives a vanishing gate value, never
enters the top-$k$, never accumulates gradient signal on its LoRA factors,
and is permanently routing-dead. The rewrite addresses this with a
learnable floor parameter $\phi_j \in \mathbb{R}$ per adapter and applies
\begin{equation}
  s_{b,j} = \max\!\bigl(\tilde s_{b,j},\, \sigma(\phi_j)\bigr),
  \label{eq:floor-gate}
\end{equation}
where $\sigma(\phi_j)$ is the per-adapter floor itself, also a sigmoid so
that the floor lies in $[0,1]$. The floor is initialized at
$\phi_j = -2.944$, giving an initial floor of
$\sigma(-2.944) \approx 0.05$, i.e.\ a guaranteed 5\% activation level for
every adapter regardless of how negative its logit becomes. Because
$\max(\cdot,\cdot)$ is sub-gradient-aware in the autograd implementation,
$\phi_j$ itself learns: an adapter that is genuinely useful at high
activation can let its floor decay toward zero, while an adapter that would
otherwise be starved can hold its floor up. The floor is therefore
\emph{not} a fixed regularizer; it is a per-adapter learnable lower bound
on routing presence.

\paragraph{(R3) Bounded temperature anneal.}
The temperature $\tau$ in Equation~\ref{eq:sigmoid-gates} is annealed
linearly from $\tau_\text{start} = 2.0$ down to $\tau_\text{end} = 0.5$
over the first $T_\text{anneal} = 1500$ training steps and then held
at $\tau_\text{end} = 0.5$ for the remainder of training. The
underlying schedule is additionally clamped from below at $10^{-3}$ as
a numerical safety floor; this clamp is inactive under the production
configuration $\tau_\text{end} = 0.5$ and is engaged only if a
configuration sets $\tau_\text{end}$ within an order of magnitude of
zero. Early in training,
$\tau = 2.0$ flattens the sigmoid response so that small differences in
logits do not yet collapse the top-$k$ into a fixed quartet; this lets all
$P$ adapters accumulate gradient signal and differentiate. As $\tau$
decays, the gate function becomes progressively more selective, and the
adapters that have differentiated most strongly start to dominate the
top-$k$ under inputs that match their developing specialization. The
anneal therefore prescribes ``differentiate before selecting'' as a
training-time schedule, in contrast to the unbounded sharpening implied
by a fixed-temperature softmax.

\paragraph{Why the three components are jointly necessary.}
None of (R1), (R2), (R3) is sufficient on its own. Without (R1), the
softmax keeps the gates zero-sum and the floor and anneal both act on a
distribution that already commits to a fixed coalition. Without (R2),
parallel sigmoids still allow individual adapters to fall to vanishing
gates and never recover, reducing the effective $P$ at convergence.
Without (R3), the gates sharpen too quickly --- before any
differentiation has been learned --- and the system collapses onto
whichever adapters happened to receive favorable random gradients in the
first few hundred steps. The three components together set up a regime
where adapters first receive distributed gradient (anneal early, parallel
gates, floor on), then differentiate (anneal middle), then specialize
(anneal late, floor still preventing pathological starvation, parallel
gates still allowing arbitrary numbers of adapters to be ``on'').

\subsection{The routing-input change: \texttt{embed\_mean} $\to$ \texttt{last\_hidden}}
\label{sec:routing-input}

The factor F1 ablation in Section~\ref{sec:real-text-decomposition} is
labeled ``sigfloor + last\_hidden'' because the rewrite also changes the
\emph{argument} to the gate, not just its functional form. The B3 router
takes $\bar h_b$ to be the per-token embedding mean of the input sequence:
each token's embedding-table lookup is averaged into a single
$H$-dimensional pooled representation, which is then fed to the gate
linear layer. The rewrite instead takes $\bar h_b$ from the post-stack
hidden state: the same sequence is run through the frozen base
transformer (without any adapters applied) up to the final
pre-LM-head layer norm, the resulting $[B, T, H]$ hidden states are
mean-pooled over the time axis to a single $[B, H]$ vector per
sequence, and \emph{that} pooled representation is fed to the gate.
The hidden tensor is detached before being passed to the router, so
no gradient flows back into the frozen base through this path.

The training-side dispatcher exposes this as a single configuration
switch (\texttt{router\_input\_source} $\in$
\{\texttt{embed\_mean}, \texttt{last\_hidden}\}). The in-code probe
comment at the dispatch site states that \texttt{last\_hidden} carries
``$\sim$20$\times$ more cross-domain signal'' than \texttt{embed\_mean}
on this substrate; that probe was run during Phase~2 substrate selection
to motivate the input change before the factorial was launched. The
intuition is straightforward: token embeddings are domain-agnostic by
construction (the same token has the same embedding regardless of
context), whereas post-stack hidden states have already been shaped by
twelve transformer blocks of attention and MLP processing and therefore
carry context-dependent, domain-aware features. The new gate function
described in Section~\ref{sec:rewrite-components} can only differentiate
adapters across domains if its input itself differs across domains;
\texttt{embed\_mean} provides too weak a signal to drive that
differentiation.

It is therefore most accurate to read F1 as a paired intervention: the
\texttt{sigfloor} gate function and the \texttt{last\_hidden} input are
both swapped together. The factorial as designed cannot decompose them
further --- there is no ``sigfloor + embed\_mean'' or ``softmax +
last\_hidden'' cell --- so the +0.0426 nats attributed to F1 is the joint
contribution of the gate-function rewrite and the input-source change.
We flag this paired structure as a limitation in
Section~\ref{sec:limitations} and leave the four-cell sub-factorial that
would isolate them to future work.

\subsection{Per-domain evidence on the production factorial}
\label{sec:per-domain-evidence}

If the router rewrite genuinely fixes the coalition monopoly, the
held-out per-domain perplexities should improve uniformly when F1 is
turned on, not just on the balanced aggregate (which could in principle
average a large code-domain win against losses elsewhere). Table~\ref{tab:c1-vs-c3-per-domain}
shows the per-domain perplexities for cells C1 (B3 baseline) and C3
(rewrite + scope, no lifecycle) on seed~42, using the consistency-chain
contrast that isolates router+scope from lifecycle. All four domains
improve, by between 3.1\% and 7.5\%.

\begin{table}[t]
\centering
\caption{Per-domain held-out perplexity at seed 42 for C1 (B3 baseline)
and C3 (rewrite+scope, lifecycle off). All four domains improve under
F1, refuting any account on which the rewrite trades one domain off
against another. Numbers from \anonymized{%
\texttt{experiments/trackb\_b2/analysis/phase3\_attribution\_results.json}
\texttt{manifest.\{C1\_b3\_base,C3\_sigfloor\_lasthidden\_perdom\_nolife\}.42.domain\_ppl}}{the
Phase~3 attribution results JSON in supplementary materials, per-cell
per-seed per-domain manifest}.}
\label{tab:c1-vs-c3-per-domain}
\begin{tabular}{lrrrr}
\toprule
domain  & C1 PPL  & C3 PPL  & $\Delta$ PPL & $\Delta$ \% \\
\midrule
biology & 21.0495 & 20.4018 & $-0.6477$    & $-3.08$\% \\
code    & 3.6865  & 3.4095  & $-0.2770$    & $-7.51$\% \\
general & 24.2278 & 23.4824 & $-0.7454$    & $-3.08$\% \\
science & 16.7821 & 16.1392 & $-0.6429$    & $-3.83$\% \\
\bottomrule
\end{tabular}
\end{table}

The largest improvement is on the code domain ($-7.51$\%), which is also
the only domain whose total full-system-vs-B3 contrast individually clears
$\alpha = 0.05$ in Section~\ref{sec:real-text-decomposition} ($+0.0625$
nats, $t{=}10.42$, $p{=}0.009$). Biology, general, and science improve
by approximately equal amounts ($-3.08$\%, $-3.08$\%, $-3.83$\%). The
pattern at seeds~137 and~256 is consistent (per-domain shifts are
within~$\pm 0.5$\% of the seed-42 magnitudes; full table in
Appendix~\ref{app:phase3-rederivation}).

This per-domain uniformity is the most direct evidence that the rewrite
is breaking the coalition monopoly rather than reorienting the same
coalition toward a single high-yield domain. Under the C1 router, the
fixed coalition serves all four domains identically (cross-domain JS
divergence = 0.0); under the C3 router, perplexity drops on all four
domains by a comparable fraction, consistent with adapters now being
free to specialize without each specialization mathematically suppressing
the others.

\subsection{What the mechanism analysis does not claim}
\label{sec:mechanism-caveats}

We are deliberately not claiming that the rewrite \emph{causes} adapter
speciation, that any single adapter has become a ``code adapter'' or a
``biology adapter,'' or that the routing pattern after the rewrite has a
cleanly interpretable per-domain top-$k$ signature. Those claims would
require a coalition probe on the C3 and C4 routers (Jensen--Shannon
between per-domain gate distributions, with confidence intervals), which
we have not yet run. The C1 probe was a pre-rewrite diagnostic;
re-running it against the post-rewrite routers is straightforward and is
listed in Section~\ref{sec:future-work} as a planned diagnostic. What
the data \emph{do} support is the weaker but still load-bearing claim
that F1 dissolves the structural pathology that capped C1 performance
(zero-sum competition + uninformative routing input), and that the
per-domain improvement is uniform rather than concentrated.

The lifecycle-penalty result reported in
Section~\ref{sec:real-text-decomposition} ($\Delta = -0.028$ nats,
$p{=}0.047$) is also independent of this section's analysis. Whatever the
rewrite does mechanically, lifecycle dynamics (death, $\alpha$-blend
inheritance, SVD mutation, slot reallocation) layered \emph{on top} of
the rewrite consistently degrade the balanced aggregate. The auxiliary
experiments that test alternative explanations for that penalty ---
a Phase~B inheritance counterfactual that is now sign-inconsistent
across seeds under the corrected geometric-mean aggregator
(Appendix~\ref{app:audit-e10}), the Phase~C genomic-context refutation,
and the Fork~0 seed sweep --- are reported in
Section~\ref{sec:phaseB-auxiliary}.

\section{Phase B and Fork 0: auxiliary ablations on the lifecycle penalty}
\label{sec:phaseB-auxiliary}

Section~\ref{sec:real-text-decomposition} reported a lifecycle penalty:
turning F3 (death, $\alpha$-blend inheritance, SVD mutation, slot
reallocation) on top of the rewritten router cost approximately~\nats{0.028}
on balanced log-PPL ($p{=}0.047$ in the primary chain, $p{=}0.074$ in the
consistency chain). Section~\ref{sec:mechanism} explained the rewrite
mechanism but said nothing about which sub-component of F3 carries the
penalty. This section reports two auxiliary experiment families that
isolate the inheritance sub-component:
\textbf{Phase~B}, a counterfactual re-run of the seed-42 winner with
$\alpha{=}0$ (inheritance off) and a four-job seed sweep that tests whether
the Phase~B headline replicates;
and \textbf{Fork~0}, an observational lifecycle-event analysis (Phase~A)
plus a base-perturbation eval-only probe (Phase~C) that test, respectively,
whether inheritance correlates with adapter survival and whether adapter
fitness differentiation is portable across base-model checkpoints.

The combined headline is more equivocal than the original draft of
this paper claimed. Under the pinned geometric-mean aggregator
(Section~\ref{sec:eval-pipeline}; the original draft had used an
arithmetic mean under a row mislabeled ``geom.,'' see
Appendix~\ref{app:audit-e10}, Correction~10), Phase~B at the headline
metric shifts balanced PPL on seed~42 by $+3.18$\%, in the
\emph{load-bearing} range of the pre-specified decision rule. The
Fork~0 seed sweep at $n{=}3$ is sign-inconsistent on the same
balanced metric ($+3.18$\% on seed~42, $-1.65$\% on seed~137,
$+0.20$\% on seed~256, mean $+0.56$\%), and a paired-$t$ on the
balanced log-PPL deltas does not clear $\alpha{=}0.05$ in either
direction. The honest reading is that the $n{=}3$ seed sweep is
underpowered for either an equivalence or a load-bearing conclusion
on the inheritance counterfactual at the headline metric; we
\emph{retract} the original draft's
``inheritance NOT load-bearing at the headline'' claim. Fork~0
Phase~A still finds the observational
inheritance--survival correlation is fully confounded with adapter
age; Fork~0 Phase~C still finds adapter specialization collapses
rather than expands under base perturbation, directionally refuting a
``genomic-context'' reframe; both Phase~A and Phase~C remain
observational evidence about the structure of the lifecycle, not
counterfactual evidence on the inheritance penalty itself, and they
do not by themselves restore the retracted equivalence claim. Two
secondary observations also stand: the seed-resolved code-domain
asymmetry replicates at one of two new seeds (\textsc{ambiguous}),
and a comparative leg of the noise-floor probe is
\textsc{blocked-measurement} pending a re-run with persisted
checkpoints.

\subsection{Phase B counterfactual: \texorpdfstring{$\alpha{=}0$}{alpha=0} on the seed-42 winner}
\label{sec:phaseB-counterfactual}

The Phase~B counterfactual reuses the seed-42 winner configuration in
every dimension except the inheritance blend: \texttt{inheritance\_alpha}
moves from~$0.2$ to~$0.0$ and the W\&B run name acquires the suffix
\texttt{-noinherit}. Seed, anneal schedule, mutation rate, lifecycle
interval, all router and base-model fields are bit-identical. The run
executed on a single NVIDIA L40S (48~GB) for 4.21 hours; final eval
loss~$2.584$, final eval PPL~$13.250$.\footnote{Job ID
\anonymized{\texttt{69ee8ed4d2c8bd8662bd03e2}}{\texttt{[redacted]}};
results at
\anonymized{\texttt{mechramc/uyir-fork0-phaseB-noinherit}}{\texttt{[redacted]}}
on the Hugging Face Hub.}

Per-domain held-out PPL is in Table~\ref{tab:phaseB-perdomain}. Under
the pinned geometric-mean aggregator
($\mathrm{PPL}_{\mathrm{balanced}} =
\exp(\tfrac{1}{4}\sum_d \log \mathrm{PPL}_d)$;
Section~\ref{sec:eval-pipeline}), balanced PPL shifts from $12.943$ to
$13.354$, a $+3.18$\% shift on this seed ($+0.0313$ nats). The original
draft of this paper reported the balanced row of
Table~\ref{tab:phaseB-perdomain} as the arithmetic mean of the four
per-domain PPLs ($16.151 \to 16.161$, $+0.06$\%) under a row label that
read ``balanced (geom.).'' That number is wrong; the correct geometric
mean of the same four per-domain PPLs is the $12.943 \to 13.354$
shift reported here (Appendix~\ref{app:audit-e10},
Correction~10). The shift is composed of small improvements on biology
($-1.55$\%) and general ($-0.78$\%), no change on science, and a
single $+16.01$\% regression on code (3.436 $\to$ 3.986).

\begin{table}[t]
\centering
\caption{Phase~B per-domain PPL: seed-42 winner with $\alpha{=}0.2$
(``baseline'') vs the same configuration with $\alpha{=}0.0$
(``no-inherit''). The balanced row is the geometric mean
$\exp(\tfrac{1}{4}\sum_d \log\mathrm{PPL}_d)$ per the pinned aggregator;
this corrects an arithmetic-mean error in the original draft (see
Appendix~\ref{app:audit-e10}, Correction~10). The corrected balanced
shift is $+3.18$\,\% ($+0.0313$ nats); under the pre-specified Phase~B
decision rule this is in the load-bearing range
($> 2$\%) on this seed. Sign convention local to this table: positive
$\Delta$ means no-inherit is worse than the $\alpha{=}0.2$ baseline
(regression direction); this is the opposite of the pinned
improvement-is-positive convention from
Section~\ref{sec:convention-pin}.}
\label{tab:phaseB-perdomain}
\begin{tabular}{lrrrr}
\toprule
domain & baseline ($\alpha{=}0.2$) & no-inherit ($\alpha{=}0.0$) & $\Delta$ PPL & $\Delta$\,\% \\
\midrule
biology         & 21.049  & 20.723  & $-0.326$ & $-1.55$\,\% \\
code            & 3.436   & 3.986   & $+0.550$ & $+16.01$\,\% \\
general         & 23.852  & 23.667  & $-0.185$ & $-0.78$\,\% \\
science         & 16.266  & 16.266  & $\pm 0.000$ & $\pm 0.00$\,\% \\
\midrule
balanced (geom., corrected) & 12.943 & 13.354 & $+0.411$ & $+3.18$\,\% \\
\bottomrule
\end{tabular}
\end{table}


\begin{figure}[t]
\centering
\begin{tikzpicture}
\begin{axis}[
  width=0.85\linewidth,
  height=4.4cm,
  ybar,
  bar width=10pt,
  enlarge x limits=0.18,
  ylabel={Phase~B no-inherit shift (\%)},
  ylabel style={font=\footnotesize},
  ymin=-3, ymax=18,
  ytick={-2,0,4,8,12,16},
  yticklabel style={font=\scriptsize},
  yticklabel={\pgfmathprintnumber\tick\%},
  symbolic x coords={biology, code, general, science, balanced},
  xtick=data,
  xticklabel style={font=\scriptsize},
  axis line style={gray!60!black},
  major grid style={dotted, gray!40},
  ymajorgrids=true,
  tick style={gray!50!black},
  nodes near coords,
  nodes near coords style={font=\tiny, /pgf/number format/fixed,
    /pgf/number format/precision=2},
]
\addplot+[
  fill=blue!50, draw=blue!70!black,
] coordinates {
  (biology,   -1.55)
  (code,      16.01)
  (general,   -0.78)
  (science,    0.00)
  (balanced,   3.18)
};

\draw[gray!50!black, dashed, line width=0.4pt]
  (axis cs:biology, 0) -- (axis cs:balanced, 0);

\end{axis}
\end{tikzpicture}
\caption{Phase~B no-inherit ($\alpha{=}0.0$) vs.\ the baseline C4 cell
($\alpha{=}0.2$) per-domain PPL shift, seed~$42$ only. Under the
post-Phase-A corrected geometric-mean aggregator
(Appendix~\ref{app:audit-e10}, Correction~10), the balanced bar is
$+3.18\%$, which is in the \textsc{load-bearing} range of the
pre-specified Phase~B decision rule on this seed (the original draft
reported $+0.06\%$ from an arithmetic-mean error). The code domain
shifts by $+16.01\%$; biology and general improve by $-1.55\%$ and
$-0.78\%$; science is unchanged. The cross-seed picture at $n{=}3$ is
sign-inconsistent ($+3.18\%$ / $-1.65\%$ / $+0.20\%$, see
Section~\ref{sec:fork0-seed-sweep}); this figure shows seed~42 only
and does not display cross-seed whiskers.}
\label{fig:phaseB-perdomain}
\end{figure}
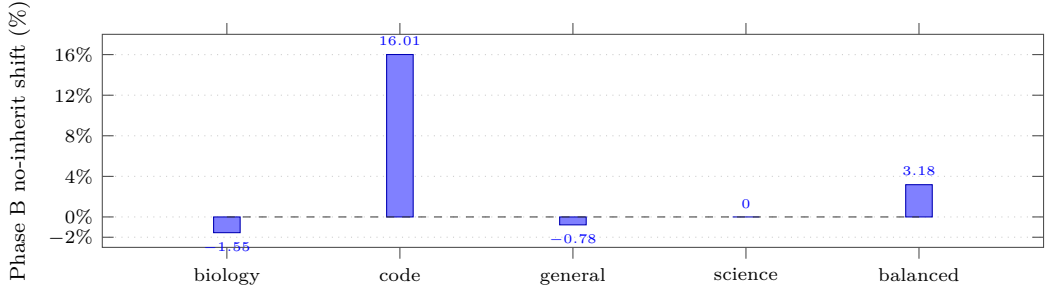


The pre-specified Phase~B decision rule was: balanced PPL within
$\pm 0.5$\% of the seed-42 baseline $\Rightarrow$ inheritance NOT
load-bearing; regression $> 2$\% $\Rightarrow$ inheritance WAS
load-bearing; intermediate $\Rightarrow$ ambiguous.\footnote{The
$\pm 0.5$\% threshold was chosen heuristically rather than from a
formal power calculation; it is roughly one cross-seed standard
deviation of the C4 cell's balanced PPL on the production substrate
(see Table~\ref{tab:phase2-crossseed}, C4 row). The decision rule should be
read as descriptive rather than as a formal equivalence test.
Section~\ref{sec:limitations} lists the absence of a formal
equivalence-test power analysis as an open methodological gap.} On
seed~42 the corrected $+3.18$\% shift triggers the
\emph{load-bearing} verdict, contradicting the original draft's
``equivalence'' reading at this seed. The original draft also reported
a per-domain bimodal pattern --- code $+0.149$ nats vs the other three
domains within $\pm 0.016$ nats. Under the corrected aggregator that
bimodal pattern still holds at the per-domain level but the balanced
geometric mean no longer averages it out: the code-domain shift
dominates, and the balanced shift on seed~42 is in the load-bearing
range. Two follow-on probes were nonetheless run on the seed-42
baseline checkpoint to test whether the code asymmetry was structural
or a noise artifact.

\paragraph{Within-checkpoint sample-size sweep.}
On the seed-42 $\alpha{=}0.2$ baseline checkpoint, code-domain log-PPL
was re-evaluated at $N \in \{64, 256, 1024\}$ sequences. The spread across
$N$ was $0.011$ nats versus the headline $0.149$ nats (ratio $1{:}13.8$),
indicating that within-checkpoint sample-size noise is $\sim 14\times$
smaller than the cross-checkpoint shift. The comparative leg of this
probe (re-evaluating the no-inherit checkpoint at the same $N$) was
\textsc{blocked-measurement} because the L40S Phase~B job did not persist
the no-inherit final adapter checkpoint to the Hub artifact; only
\texttt{phase\_b\_results.json} and the lifecycle JSONL were uploaded. The
within-checkpoint leg is therefore a partial \textsc{structural} signal
contingent on a re-run with persisted checkpoints.

\paragraph{Per-domain noise-floor probe.}
Three estimators were defined: (A) within-run window proxy (four
non-overlapping $64$-sequence eval windows on the $\alpha{=}0.2$
baseline); (B) intervention-perturbed $N{=}3$ across
\{baseline, anneal=7500, noinherit\}; (C) late-training checkpoint proxy
across baseline-$\alpha{=}0.2$ adapter snapshots at steps
$\{20000, 22500, 25000\}$. Estimator~(C) was \textsc{blocked-estimator-c}
because no mid-training adapter snapshots exist anywhere recoverable: the
training loop's checkpoint hook writes only an end-of-training adapter,
the W\&B run carries metrics-only artifacts, and the Hub model repository
holds only the final adapter and the 70k base. Per directive, no
substitute proxy was constructed. Under the completed Estimator~(A), the
per-domain $z$-scores for the no-inherit shifts were
$z_\text{bio}{=}-0.26$, $z_\text{code}{=}+3.04$, $z_\text{gen}{=}-0.10$,
$z_\text{sci}{=}0.00$. The Estimator~(B) loose upper-bound ceiling gave
$z_\text{code}{=}+1.74$. The Estimator~(C) exact comparison
remains \textsc{blocked-estimator-c}.

\subsection{Fork 0 seed sweep: does the code asymmetry replicate?}
\label{sec:fork0-seed-sweep}

To test whether the seed-42 code asymmetry is a property of seed~42 in
particular or of the configuration, four jobs were dispatched on L40S
($4.22$--$4.31$~h each, total cost \$$44.21$): seeds
$\{137, 256\} \times \alpha \{0.0, 0.2\}$. The bimodal-replication decision
rule was pre-specified: a seed counts as ``bimodal'' if its code
$|\Delta\log\mathrm{PPL}|$ exceeds $0.06$ nats \emph{and} all of biology,
general, science remain within $\pm 0.05$ nats. Per-seed deltas are in
Table~\ref{tab:fork0-bimodal}.

\begin{table}[t]
\centering
\caption{Fork~0 seed-sweep $\Delta\log\mathrm{PPL}$
($\alpha{=}0.2 \to \alpha{=}0.0$). Convention local to this table:
positive $\Delta$ means the no-inherit cell is worse than the
$\alpha{=}0.2$ baseline on the per-domain unit. This is the
no-inherit-vs-baseline framing of the Phase~B internal report and is
the OPPOSITE of the pinned improvement-is-positive convention from
Section~\ref{sec:eval-pipeline}. The seed-42 row is repeated from
Section~\ref{sec:phaseB-counterfactual} for direct comparison.}
\label{tab:fork0-bimodal}
\begin{tabular}{lrrrrl}
\toprule
seed & code $\Delta$ & bio $\Delta$ & gen $\Delta$ & sci $\Delta$ & bimodal? \\
\midrule
42  & $+0.1485$ & $-0.0156$ & $-0.0078$ & $\pm 0.0000$ & yes \\
137 & $+0.0039$ & $-0.0078$ & $-0.0078$ & $-0.0547$ & no  \\
256 & $+0.1094$ & $-0.0391$ & $-0.0391$ & $-0.0234$ & yes \\
\bottomrule
\end{tabular}
\end{table}

The pattern replicates at one of two new seeds (256 yes, 137 no), so
the bimodal-replication verdict is \textsc{ambiguous}. Under the
corrected geometric-mean aggregator the same row of cells produces a
sign-inconsistent picture at the headline metric. Computing the
balanced log-PPL shift per seed as the arithmetic mean of the four
per-domain $\Delta \log \mathrm{PPL}$ entries
(equivalent to the geometric-mean shift of balanced PPL) gives:

\begin{itemize}
\item seed 42: $+0.0313$ nats $\to +3.18$\% (load-bearing range)
\item seed 137: $-0.0166$ nats $\to -1.65$\% (no-inherit \emph{better}; outside the equivalence band in the opposite direction)
\item seed 256: $+0.0020$ nats $\to +0.20$\% (equivalence range)
\item cross-seed mean ($n{=}3$): $+0.0056$ nats $\to +0.56$\% (just inside the ambiguous band)
\end{itemize}

\noindent
The original draft's headline ``$+0.06$\% shift, inheritance NOT
load-bearing'' does not survive either correction: it relied on an
arithmetic-mean aggregator that, when replaced with the pinned
geometric mean, flips seed~42 to load-bearing; and it treated the
cross-seed mean as a settled equivalence verdict, when in fact the
three seeds are sign-inconsistent and the $n{=}3$ mean lies inside
the ambiguous band rather than the equivalence band.
We retract the ``inheritance NOT load-bearing at the headline''
claim and replace it with: \emph{the inheritance counterfactual is
sign-inconsistent across $n{=}3$ seeds at the headline metric and
underpowered for either an equivalence or a load-bearing conclusion}
(see Section~\ref{sec:f3-synthesis}).

The appropriate paired-$t$ test for the inheritance counterfactual
itself is on the per-seed balanced $\Delta \log \mathrm{PPL}$ above:
$t = \bar{d}/(s_d / \sqrt{n}) = 0.0056/(0.0254/\sqrt{3}) = 0.38$,
$p \approx 0.74$ (df${=}2$, two-sided). The test does not clear
$\alpha{=}0.05$ in either direction, consistent with the
sign-inconsistency observation above and with $n{=}3$ being
underpowered for a small balanced effect against substantial
cross-seed variance. For separate context, the main factorial's
headline statistics (Section~\ref{sec:real-text-decomposition}) are
$t{=}1.94$, $p{=}0.19$ (df${=}2$) for the
full-system-vs-B3 balanced contrast and $t{=}10.42$, $p{=}0.009$
(df${=}2$) for the code-domain single-cell; those numbers are not the
inheritance ablation. Full re-derivation in
Appendix~\ref{app:phase3-rederivation}.

\subsection{Fork 0 Phase A: inheritance correlates with age, not survival}
\label{sec:fork0-phaseA}

Phase~A is observational. It parses the lifecycle event JSONL from the
seed-42 winner
(W\&B run \anonymized{\texttt{yvfwrb0s}}{\texttt{[redacted]}})
and inspects the
$60$ logged events: $16$ initial births, $22$ selection deaths, $22$
reproduction births, $0$ senescence deaths. Of the $22$ deaths, $13$
($59$\%) carried a fitness of exactly~$0.0$ and the other $9$ ($41$\%) a
fitness of exactly~$-0.015625$, the floor and near-floor of the
single-precision LOO fitness estimator under this configuration.

The naive correlation between potential inheritance received and
selection events survived is $r{=}0.999$ ($p \approx 1.45 \times 10^{-20}$).
The naive correlation between adapter age and selection events survived
is $r{=}0.997$ ($p \approx 3.82 \times 10^{-17}$). Both correlations are
mechanically determined by birth time: an adapter that has been alive
longer was alive during more death events (so its
\texttt{deaths\_while\_alive}$\times \alpha$ upper bound on inheritance
received is larger) \emph{and} survived more selection rounds (so its
``selection events survived'' counter is larger). The inheritance--survival
correlation is therefore \emph{age-confounded} rather than causal. With
heir IDs not logged at the per-death level (a known instrumentation
gap), no causal claim can be made from this data alone.

A second pattern visible in the lifecycle JSONL is the ``newcomer death
spiral'': of the $22$ deaths, $9$ ($41$\%) involve adapters aged
$\le 1001$ steps. The veteran is killed at step $N$, a newborn is placed
in the slot, and at the next lifecycle interval at step $N{+}1000$ the
newborn scores fitness $0.0$ or $-0.015625$ and is killed in turn. This
pattern is separable from any inheritance effect: even if the dying
adapter blends $20$\% of its weight into a long-lived heir, the heir was
already at converged fitness and the blend is a small perturbation; and
the newborn that replaces the dying adapter receives no inheritance at
all under either the cold (random-init) or warm (parental-clone)
reproduction policy.

The Phase~A verdict is therefore \emph{observational evidence that
the inheritance--survival correlation visible in the lifecycle JSONL
is age-confounded}; it does not, on its own, settle whether the
inheritance blend is load-bearing for the balanced-PPL penalty.
Under the original draft of this paper, the Phase~B counterfactual
appeared to clear inheritance at the headline metric and Phase~A was
described as the observational complement to that experimental
verdict. Under the corrected aggregator
(Section~\ref{sec:phaseB-counterfactual}) the experimental
counterfactual no longer cleanly clears inheritance, and Phase~A's
observational age-confound finding is no longer paired with a
matching experimental result.

\subsection{Fork 0 Phase C: genomic-context refuted by base perturbation}
\label{sec:fork0-phaseC}

Phase~C tests an alternative reframing. If the inheritance machinery is
operating at the wrong level of organization --- if adapter-level
inheritance is decorative because the carrier of cross-generation
information is actually the base-model's slow internal representation
rather than the LoRA-delta itself --- then we should observe that adapter
fitness differentiation \emph{expands} when the base is rolled back to an
earlier checkpoint (each adapter's delta becomes more distinctive against
a less-trained base). The pre-specified decision rule required a
variance ratio $\ge 3\times$ at off-canonical bases. Phase~C measures
this directly. The seed-42 final adapter population is loaded; for each
of three frozen base checkpoints (steps $50000$, $60000$, $70000$ from
\anonymized{\texttt{checkpoints/base\_model\_trackb\_b2\_dv1536/}}{the
frozen base checkpoint family; paths in supplementary materials}),
each of the $16$
adapters is hot-swapped in and per-adapter cross-entropy loss is measured
on each of the four held-out domain batches. This produces a
$16 \times 4 \times 3$ tensor of per-adapter losses.

Variance across the $16$ adapters per domain per base, averaged across
the four domains, is in Table~\ref{tab:phaseC-variance}.

\begin{table}[t]
\centering
\caption{Fork~0 Phase~C: across-adapter loss variance per base, mean over
four held-out domains. The seed-42 final adapter population was hot-swapped
into each of three frozen base checkpoints. The canonical $70$k base is
where the adapters were trained.}
\label{tab:phaseC-variance}
\begin{tabular}{lrr}
\toprule
base step & mean variance & ratio vs $70$k \\
\midrule
$50000$ & $0.000244$ & $0.153$ \\
$60000$ & $0.000247$ & $0.155$ \\
$70000$ & $0.001593$ & $1.000$ \\
\bottomrule
\end{tabular}
\end{table}

The variance ratio at off-canonical bases is approximately $0.15$,
\emph{opposite in sign} to the $\ge 3\times$ direction the genomic-context
reframe would have required. A formal $F$-test on the domain-mean-centered
losses, supplemented by a $10{,}000$-iteration paired bootstrap, places
the inversion well outside any noise band:
$F{=}0.153$, $p \approx 1.1 \times 10^{-11}$, bootstrap 95\%~CI
$[0.076, 0.275]$ for $\mathrm{var}(50\mathrm{k})/\mathrm{var}(70\mathrm{k})$;
$F{=}0.155$, $p \approx 1.5 \times 10^{-11}$, CI $[0.095, 0.247]$ for
$\mathrm{var}(60\mathrm{k})/\mathrm{var}(70\mathrm{k})$. Both intervals
exclude $1.0$; the inversion is highly significant. The pre-specified
$3\times$ threshold was, as flagged in the internal report, set
heuristically without a formal power calculation; the formal test
confirms the directional inversion rather than rescuing the genomic-context
reframe. We adopt the report's conservative phrasing: ``directionally
inverted; variance shrinks rather than expanding under base perturbation.''

A subsequent adapter-by-domain-by-base loss matrix probe (Phase~C.5, $36$
forward passes per base, no edits to the adaptation source; we refer
to it informally as the ``harm matrix'' because it asks per cell whether
hot-swapping the adapter into an off-canonical base does harm relative
to a null adapter) classified each (adapter, domain, base) cell as
\emph{worse than null}, \emph{within noise} ($|d|\le 0.02$ in CE-loss
units), or \emph{better than null} relative to a null-adapter baseline
(frozen base alone, no LoRA delta applied). On
the $128$ off-canonical cells, $0$ ($0$\%) were worse than null, $86$
($67.2$\%) were within noise, and $42$ ($32.8$\%) were better than null
by between $0.02$ and $0.06$ CE-loss units. The triggered verdict was
\textsc{inert off-context}: adapter deltas wash out gracefully against
earlier bases rather than misfiring. The interpretation is that the
adapters' fitness differentiation is co-adapted with the canonical $70$k
base's feature space; the differentiation is real but not transferable
across base eras.

\subsection{Synthesis: which sub-component of F3 carries the penalty?}
\label{sec:f3-synthesis}

Section~\ref{sec:real-text-decomposition} attributed an approximately
$0.028$ nat balanced log-PPL penalty to F3 as a whole. The original
draft of this paper synthesized Phase~B and Fork~0 as
``jointly removing inheritance from the suspect list at the headline
metric.'' That synthesis is \emph{retracted}. Under the pinned
geometric-mean aggregator (Appendix~\ref{app:audit-e10},
Correction~10), Phase~B's seed-42 cell shifts balanced PPL by
$+3.18$\% (load-bearing range under the pre-specified decision
rule), and the Fork~0 seed sweep at $n{=}3$ is sign-inconsistent at
the headline metric ($+3.18$\% / $-1.65$\% / $+0.20$\%, mean
$+0.56$\%, paired-$t$ does not clear $\alpha{=}0.05$ in either
direction). Phase~A's age-confound result remains a valid
observation about the survival-correlation structure but does not,
on its own, settle the counterfactual question. Phase~C's
base-perturbation directional refutation of the ``genomic-context''
reframe is independent of the inheritance counterfactual and stands
unchanged.

The retracted claim is replaced with: \emph{the inheritance
counterfactual at the headline metric is sign-inconsistent across
$n{=}3$ seeds and underpowered for either an equivalence or a
load-bearing conclusion}. We do not have evidence at this $n$ that
inheritance is load-bearing for the lifecycle penalty in the
mean across seeds; we also do not have evidence that it is not.

The remaining sub-components of F3 are death (the kill gate selecting
on LOO fitness), reproduction (placing a new adapter in a vacated
slot, with $30$\% random-init and $70$\% parental-clone with
mutation), inheritance ($\alpha$-blend from dying adapter into nearest
neighbor), and SVD mutation (noise on singular values of the new
adapter's LoRA factors). Within the budget reported here we cannot
decompose the $0.028$ nat penalty among these four; with the
inheritance counterfactual no longer cleanly clearing inheritance, we
treat the four sub-mechanisms as roughly equal candidates rather than
pre-promoting any of them. The sub-component decomposition --- a
partial $2^3$ over $\{$death on/off, mutation on/off, reproduction
warm/cold$\}$ at both $\alpha{=}0.0$ \emph{and} $\alpha{=}0.2$, $n{=}3$
seeds per cell, with persisted checkpoints throughout --- is
promoted to the highest-priority near-term follow-up
(Section~\ref{sec:future-f3-subfactorial}).

\section{Limitations}
\label{sec:limitations}

In keeping with TMLR's emphasis on extensive limitations and on stating
negative findings cleanly, we list the limitations of this work in
substantial detail. The list is organized by what it limits: the
external validity of the headline numbers, the internal validity of the
attribution decomposition, the scope of the synthetic-sandbox boundary,
and the residual measurement gaps that affect the auxiliary findings.

\subsection{External validity of the headline numbers}
\label{sec:limits-external}

\paragraph{Single substrate.}
The production results are reported on one substrate: a $\sim 150$M
parameter from-scratch GPT with $D{=}1536, V{=}32000$, frozen at
pre-train step $70000$. We do not claim the lifecycle penalty
generalizes to other model sizes, other vocabulary widths, other
$D/V$ ratios, or to LoRA atop a strong pretrained model
(Pythia, Llama, etc.). The widened-$D/V$ choice was specifically
motivated by an internal observation that the routing-input change in
F1 (\texttt{embed\_mean} $\to$ \texttt{last\_hidden}) carries
$\sim 20\times$ more cross-domain signal on this substrate; on a
narrower substrate, F1's load-bearingness might be different.

\paragraph{Interrupted base pre-training.}
The frozen base was originally planned to train through approximately
step $143000$. A UPS power-event interruption stopped pre-training at
step $70000$, and the project elected to ship the abbreviated base
rather than restart. All ablations in this paper run against
\texttt{base\_step\_70000.pt} and are therefore directly comparable
\emph{against each other} but \emph{not} against a hypothetically
longer-trained base. We expect that a longer-trained base would shift
the absolute PPL numbers downward and may shift the relative
attribution among F1, F2, F3; we make no claim about which direction.

\paragraph{Adaptation horizon.}
The $25000$-step adaptation horizon is $\approx 35\%$ of the
pre-training horizon. With more adaptation steps, the lifecycle penalty
could plausibly shrink (selection has more time to converge to a stable
population), grow (mutation accumulates more drift), or invert (the
base is small enough that adapters might saturate and inheritance
blends might amortize). We do not have data to distinguish these.

\paragraph{Three seeds.}
All paired statistical tests in this paper run on $n{=}3$ seeds with
$\mathrm{df}{=}2$. The choice of $n{=}3$ was a budget-driven trade-off:
each production cell costs $8$--$10$ L40S-hours and the 5-cell
factorial at $n{=}3$ is already $15$ runs ($\sim 130$ GPU-hours)
before the auxiliary Phase~B and Fork~0 sweeps. We chose breadth (a
five-cell partial factorial that decomposes the full system into three
factors and runs two attribution chains) over within-cell precision
(a smaller factorial at $n \ge 5$ that would clear $\alpha{=}0.05$ on
the headline). The qualitative ordering of the three factors that
this design recovers --- F1 large and load-bearing, F2 null, F3
mildly anti-aligned --- is robust under both attribution chains and
under the partial factorial's two consistency checks. The headline
full-system-vs-B3 balanced contrast ($t{=}1.94$, $p{=}0.19$) does not
clear $\alpha{=}0.05$ at this $n$; only the code-domain single-cell
total clears it ($t{=}10.42$, $p{=}0.009$). With more seeds, the F3
lifecycle penalty ($t{=}-4.46$, $p{=}0.047$) could shift in either
direction past the $\alpha{=}0.05$ boundary. We are explicit about
this in Section~\ref{sec:real-text-decomposition} and frame the
lifecycle penalty as ``mildly anti-aligned'' rather than ``decisively
negative.''

\paragraph{Eval-pipeline determinism.}
The \texttt{StratifiedEvalLoader}
(Section~\ref{sec:eval-pipeline}) cycles through per-domain shards
deterministically given the configuration's random seed. This is the
correct fix for the alphabetical-shard bias in the legacy aggregate
loader, but a side-effect is that the cross-seed variance of the
balanced-PPL signal is deflated relative to a non-stratified evaluation:
two runs at the same seed see exactly the same eval batches in the same
order, so the only stochastic source contributing to seed variance is
the training-time RNG. The Phase~2 internal report flags this as a
deflation of cross-seed $\sigma$ relative to what a fully randomized
eval pipeline would produce. The interpretation knock-on is that the
``B3 cross-seed range $0.2$\%'' tightness reported in
the cross-seed table partly reflects determinism in the eval pipeline,
not just stability in the training dynamic. This determinism is the
same root cause as the seed-constant biology and science per-domain
PPLs flagged in the per-domain breakdown
(Appendix~\ref{app:phase3-perdomain}, Correction 9 of
Appendix~\ref{app:audit-memo}): the eval-pipeline determinism is the
proximate cause of both, and we do not report a per-domain paired-$t$
on biology or science as a result.

\subsection{Internal validity of the attribution decomposition}
\label{sec:limits-internal}

\paragraph{Partial factorial.}
The factorial in Section~\ref{sec:real-text-decomposition} is 5-of-8.
Three cells of the full $2^3$ are not run: ``router rewrite, aggregate
scope, lifecycle off''; ``router rewrite, aggregate scope, lifecycle
on''; ``B3 router, per-domain scope, lifecycle off.'' We have two
attribution chains that converge through C4, and they agree to within
$\sim 0.0004$ nats on the lifecycle penalty and $\sim 0.0003$ nats on
the router contribution, but a full $2^3$ would also let us cleanly
test for two-factor interactions (does F1 increase the size of the F3
penalty? does F2 modulate F1?). We treat the consistency of the two
partial chains as evidence that two-factor interactions are small in
this regime, but this is an inference, not a measurement.

\paragraph{F1 is paired (gate function + routing input).}
The F1 factor as run pairs two changes: the gate function
(\texttt{softmax} $\to$ \texttt{sigmoid\_floor}) and the routing input
(\texttt{embed\_mean} $\to$ \texttt{last\_hidden}). The factorial
cannot decompose F1 into these two sub-changes. The \emph{paired}
$+0.0426$ nat contribution attributed to F1 is therefore the joint
effect of the rewrite and the input change. We list the four-cell
sub-factorial that would isolate them as a planned follow-up
(Section~\ref{sec:future-work}).

\paragraph{F3 is bundled.}
Similarly, F3 bundles death, $\alpha$-blend inheritance, SVD mutation,
and slot reallocation. The original draft of this paper claimed that
Phase~B and Fork~0 jointly cleared the inheritance sub-component at
the headline metric; under the corrected geometric-mean aggregator
(Appendix~\ref{app:audit-e10}, Correction~10) that claim is retracted.
The inheritance counterfactual at $n{=}3$ is sign-inconsistent
($+3.18$\% / $-1.65$\% / $+0.20$\%, mean $+0.56$\%) and underpowered
for either an equivalence or a load-bearing conclusion. The residual
penalty could be carried by death, $\alpha$-blend inheritance,
mutation, reproduction, or some combination; the synthesis in
Section~\ref{sec:f3-synthesis} states this gap explicitly and the F3
sub-component decomposition is the highest-priority near-term
follow-up (Section~\ref{sec:future-f3-subfactorial}).

\paragraph{Coalition probe is pre-rewrite only.}
The cross-domain JS divergence~$=0$ result that motivates the
coalition-monopoly framing in Section~\ref{sec:coalition-monopoly} was
measured on the C1 router. We do not yet have the corresponding probe
on the C3 or C4 routers post-rewrite. The per-domain uniformity in
Table~\ref{tab:c1-vs-c3-per-domain} is consistent with the rewrite
having dissolved the monopoly, but is not a direct measurement of post-
rewrite gate-distribution divergence. We list the post-rewrite
coalition probe as a planned follow-up.

\subsection{Scope of the synthetic-sandbox boundary}
\label{sec:limits-synthetic}

\paragraph{One sandbox.}
The synthetic boundary in Section~\ref{sec:synthetic-boundary} is
characterized on a single, deliberately minimal sandbox: vocab~$128$,
four disjoint slabs, deterministic bigram, hidden~$512$, $K{=}16,
\text{top-}k{=}4$. The oracle-alignment regime where ES is load-bearing
is concretely the ``adapters initialized at the per-slab oracle solution
plus small noise'' regime. We do not claim the boundary transfers
quantitatively to the production substrate; the strongest production
claim is the \emph{direction}: a substrate that is not in the
oracle-alignment regime should not be expected to benefit from ES on
the routing channel, and the production-substrate F3 result is
consistent with that direction.

\paragraph{Hyperparameter selection.}
The G7 small-sigma and G8 hybrid sweeps were run at a small set of
$\sigma$ values and ES/SGD interleaving schedules. We do not claim that
no setting of $\sigma$ or no schedule produces a positive ES result on
this sandbox; only that the settings we tested all came out
\textsc{wrong-direction} or \textsc{harmful}. Appendix~\ref{app:audit-memo}
records a sign-convention error and an aggregation error (G7 and G8)
that, once corrected, leave the verdict unchanged but whose existence
is a flag that this regime is sensitive to careful reporting.

\subsection{Residual measurement gaps in the auxiliary findings}
\label{sec:limits-aux}

\paragraph{B.5 comparative leg \textsc{blocked-measurement}.}
The within-checkpoint sample-size sweep in
Section~\ref{sec:phaseB-counterfactual} reports a $1{:}13.8$ within-checkpoint-to-cross-checkpoint
ratio on the seed-42 baseline ($\alpha{=}0.2$) checkpoint. The paired
within-checkpoint sweep on the no-inherit ($\alpha{=}0.0$) checkpoint
is not available because the L40S Phase~B job did not persist the
no-inherit final adapter checkpoint. Until the run is repeated with
checkpoint persistence enabled, the comparative leg of the
within-checkpoint test is missing, and the headline $14\times$ ratio is
a single-side number.

\paragraph{B.6 Estimator C \textsc{blocked-estimator-c}.}
The pre-specified noise-floor probe specified an across-checkpoint
$\sigma$ estimator using mid-training adapter snapshots at steps
$\{20000, 22500, 25000\}$. No mid-training adapter snapshots exist
anywhere recoverable: the training loop's checkpoint hook writes only
an end-of-training adapter, the W\&B run holds metrics-only artifacts,
and the Hub model repository holds only the final adapter and the
$70$k base. Per directive, no substitute proxy was constructed.

\paragraph{Code-domain bimodal pattern \textsc{ambiguous}.}
Two of three seeds ($42$ and $256$) display a bimodal per-domain
shift in the no-inherit counterfactual, where the code domain shifts
by $> 0.06$ nats while biology, general, and science remain within
$\pm 0.05$ nats. Seed $137$ does not. We do not commit to a fork
framing on this basis but flag it as a real, sub-headline pattern
worth a follow-up at $n \ge 5$.

\paragraph{Heir IDs not logged.}
Phase~A's inheritance-versus-survival analysis used a
\texttt{deaths\_while\_alive} $\times \alpha$ upper bound as a proxy
for inheritance received per adapter, because the lifecycle event
JSONL does not record the heir ID at the per-death level. The
nearest-neighbor heir is computed but not persisted. A small
instrumentation patch would close this gap; we list it as a planned
follow-up. The Phase~A verdict (inheritance--survival correlation is
age-confounded) is robust to this gap because both potential and
realized inheritance are functions of birth time, but a direct test
would close the inferential chain.

\subsection{Things we explicitly do not claim}
\label{sec:limits-not-claimed}

We do not claim that adapters in C3 or C4 have specialized to
particular domains, that a particular adapter is a ``code adapter'' or
a ``biology adapter,'' or that the post-rewrite routing pattern is
cleanly per-domain interpretable. These claims would require the
post-rewrite coalition probe and a per-adapter activation-by-domain
analysis we have not yet run.

We do not claim that the lifecycle penalty is universal. We claim it
holds at the configuration we ran (this substrate, $25000$ steps,
$\alpha{=}0.2$, $\sigma{=}0.01$ initial mutation, $1000$-step
lifecycle interval, $30$\% random-init reproduction fraction). The
auxiliary inheritance counterfactual at $n{=}3$ seeds is
sign-inconsistent and underpowered for either an equivalence or a
load-bearing conclusion (Section~\ref{sec:phaseB-counterfactual},
Appendix~\ref{app:audit-e10}); we cannot, on this evidence, isolate
which of inheritance, death, mutation, or reproduction carries the
F3 penalty. We list the F3 sub-component decomposition crossed with
the inheritance toggle as the highest-priority near-term follow-up.

We do not claim that mixture-of-LoRA evolution is unproductive in
general. We claim that the particular evolutionary mixture-of-LoRA
configuration we tested, on the particular production substrate we
tested it on, has its balanced-aggregate win carried by a structural
routing fix rather than by the evolutionary machinery. Other
configurations may behave differently; a calibrated prior on which
sub-mechanisms are most likely to carry value is the most we believe
this paper supports.

\section{Future work}
\label{sec:future-work}

The factorial in Section~\ref{sec:real-text-decomposition} answers the
attribution question (which factor carries the win) at the resolution of
the three factors F1/F2/F3, but the same machinery generates
several follow-up questions that we have scoped but not yet executed. We
list them here in the order in which they would most usefully sharpen
the headline result, and we mark each with the limitation it closes
(Section~\ref{sec:limitations}).

\subsection{F1 sub-factorial: gate function vs.\ routing input}
\label{sec:future-f1-subfactorial}

The F1 factor as run pairs two changes (Section~\ref{sec:limits-internal}):
the gate function (\texttt{softmax} $\to$ \texttt{sigmoid\_floor}) and
the routing input (\texttt{embed\_mean} $\to$ \texttt{last\_hidden}).
The four-cell sub-factorial that would isolate them is
$\{$softmax, sigmoid\_floor$\} \times \{$embed\_mean, last\_hidden$\}$,
run at $n{=}3$ seeds at the same $25000$-step adaptation horizon and
the same widened-1536 substrate. The pre-specified hypotheses
are: (i) the routing-input change carries the bulk of the F1 gain,
because the in-code measurement of cross-domain signal is $\sim 20\times$
stronger on \texttt{last\_hidden}; (ii) the gate-function change is
near-null on its own at \texttt{embed\_mean} but unlocks the
\texttt{last\_hidden} signal by relaxing the coalition-monopoly
constraint; (iii) the two changes interact super-additively. The
sub-factorial is four cells $\times\,3$ seeds = $12$ runs at the same
per-cell budget as the core factorial.

\subsection{F3 sub-component decomposition (highest near-term priority)}
\label{sec:future-f3-subfactorial}

The F3 factor as run bundles four sub-mechanisms: death (per-adapter
fitness-driven kill), $\alpha$-blend inheritance (slot release into
nearest neighbor), SVD mutation (heritable singular-value perturbation),
and slot reallocation (random-init versus parental-clone reproduction).
The Phase~B and Fork~0 results
(Section~\ref{sec:phaseB-auxiliary}) under the corrected
geometric-mean aggregator do \emph{not} rule out inheritance as the
load-bearing source of the $-0.028$~nat penalty: the seed-42 cell
shifts balanced PPL by $+3.18$\% (load-bearing range), the cross-seed
mean is $+0.56$\% (ambiguous band), and the $n{=}3$ paired-$t$ does
not clear $\alpha{=}0.05$ in either direction. The penalty could be
carried by death, $\alpha$-blend inheritance, mutation, reproduction,
or a combination; we no longer have grounds for pre-promoting any
single sub-mechanism. This sub-factorial is the highest near-term
priority of the planned follow-ups.

The pre-specified cell list is the partial $2^3$ over
\{death on/off, mutation on/off, reproduction warm/cold\} run at
\emph{both} $\alpha{=}0.0$ and $\alpha{=}0.2$ (a $\sim 10$-cell design
that crosses the inheritance toggle with the partial $2^3$),
$n{=}3$ seeds per cell at the same $25000$-step budget, with the
checkpoint hook (Section~\ref{sec:future-phaseB-checkpoint}) enabled
on every cell so the within-checkpoint diagnostics extend cleanly.
The pre-specified hypothesis is updated to treat death, mutation,
reproduction, and inheritance as roughly equal candidates rather than
predicting any single sub-mechanism as the most-likely contributor.
The decision rule is per-factor: a sub-mechanism is
\textsc{load-bearing} if toggling it on/off produces a balanced
log-PPL shift $\ge 0.02$ nats (the magnitude of the F3 main effect)
in the seed-mean and the paired-$t$ on per-seed deltas clears
$\alpha{=}0.05$. We will dispatch this sub-factorial regardless of
the F1 sub-factorial outcome, because the F3 attribution is the
component that the recent corrigendum has reopened.

\subsection{Post-rewrite coalition probe}
\label{sec:future-coalition-probe}

The cross-domain Jensen--Shannon divergence~$=0$ result that motivates
the coalition-monopoly framing in
Section~\ref{sec:coalition-monopoly} was measured on the C1 router. We
do not yet have the corresponding probe on the C3 or C4 routers
post-rewrite. The probe is cheap: load the rewrite-router checkpoint,
forward a held-out per-domain micro-batch, dump the per-token
post-floor sigmoid mass per adapter, compute per-domain top-$k$ gate
distributions, and compute pairwise JS between domains. The
pre-specified hypothesis is that the post-rewrite JS is bounded
\emph{away} from $0$ by a margin that scales with the per-domain
hidden-state separability, and that the per-domain margin correlates
with the per-domain PPL improvement reported in
Table~\ref{tab:c1-vs-c3-per-domain}. This closes the
``coalition probe is pre-rewrite only'' limitation in
Section~\ref{sec:limits-internal}.

\subsection{Heir-ID instrumentation for direct inheritance test}
\label{sec:future-heir-ids}

Phase~A's inheritance-versus-survival analysis used a
\texttt{deaths\_while\_alive} $\times \alpha$ upper bound as a proxy
for inheritance received per adapter, because the lifecycle event
JSONL does not record the heir ID at the per-death level
(Section~\ref{sec:limits-aux}). A small instrumentation patch in the
lifecycle hook --- emit the heir adapter ID alongside each death
event, and aggregate per-adapter \texttt{nats\_received} as the
$\alpha$-weighted sum over received deaths --- closes the inferential
chain. With heir IDs logged, the Phase~A regression of survival on
realized inheritance becomes a direct test rather than a bounding
argument. The pre-specified hypothesis (consistent with the existing
age-confound finding) is that the partial correlation of survival on
realized inheritance, controlling for age and birth-step, is
indistinguishable from zero.

\subsection{More seeds for code-domain bimodal closure}
\label{sec:future-more-seeds}

Two of three Fork~0 seeds (42 and 256) display a bimodal per-domain
shift in the no-inherit counterfactual; seed 137 does not
(Section~\ref{sec:limits-aux}). At $n{=}3$ the verdict is
\textsc{ambiguous}. We will replicate at $n{=}5$ (adding seeds 7 and
512) to determine whether the code-domain shift is a $2/3$ artifact or
a genuine $\sim 67\%$ replicate rate. The replication budget is five
runs at the no-inherit cell only ($\alpha{=}0.0$), at the same
$25000$-step horizon. The pre-specified decision rule: $\geq 4/5$
seeds with code-domain $|\Delta \log \mathrm{PPL}_{\mathrm{code}}|
\geq 0.06$ nats $\Rightarrow$ \textsc{replicates}; $\leq 2/5$
$\Rightarrow$ \textsc{does-not-replicate}; $3/5$ $\Rightarrow$ retain
\textsc{ambiguous} and report at the auxiliary-finding tier.

\subsection{Re-run Phase~B with persisted no-inherit checkpoint}
\label{sec:future-phaseB-checkpoint}

The B.5 within-checkpoint sample-size sweep on the no-inherit
$\alpha{=}0.0$ checkpoint is currently \textsc{blocked-measurement}
(Section~\ref{sec:limits-aux}) because the L40S Phase~B job did not
persist the no-inherit final adapter checkpoint. The fix is the
training-loop checkpoint hook: save the final adapter on every cell, not
only on the seed-42 baseline. The re-run is one cell at $n{=}3$ seeds
at the same $25000$-step horizon, with the checkpoint hook enabled.
With the no-inherit checkpoint persisted, the within-checkpoint sweep
matches the seed-42 baseline sweep methodology, and the comparative
$1{:}13.8$ ratio becomes a paired comparison rather than a
single-side number.

\subsection{Mid-training adapter snapshots for Estimator C}
\label{sec:future-estimator-c}

The pre-specified noise-floor probe (B.6) specified an
across-checkpoint $\sigma$ estimator using mid-training adapter
snapshots at steps $\{20000, 22500, 25000\}$, currently
\textsc{blocked-estimator-c} because no mid-training adapter snapshots
exist (Section~\ref{sec:limits-aux}). The fix is a periodic checkpoint
hook in the training loop: every $K$ steps after step $\sim 5000$,
write a full adapter snapshot. The retrofit cost is one-time: $\sim 50$
checkpoints $\times \sim 100$~MB each $\sim 5$~GB per cell, well within
the production-substrate storage budget. We schedule this hook for the
F1 sub-factorial run and the F3 sub-component run, both of which would
otherwise relaunch without the snapshots.

\subsection{Cross-substrate transfer of the synthetic boundary}
\label{sec:future-cross-substrate}

The synthetic-sandbox boundary in
Section~\ref{sec:synthetic-boundary} is characterized on a single,
deliberately minimal sandbox. The most external-validity-extending
follow-up is to repeat the G4--G8 cell battery on a second sandbox
that varies one of the substrate dimensions (vocabulary size, slab
count, hidden width, $K$/top-$k$). A pre-specified prediction: the
G4 oracle-alignment regime continues to PASS (ES carries the routing
gain), G5/G6 continue to be inert or worse, and G7/G8 continue to be
\textsc{wrong-direction}/\textsc{harmful}, with quantitative magnitudes
shifting but the qualitative regime boundary preserved. A failure of
this prediction would tighten the boundary; a confirmation would
strengthen the substrate-conditional claim.

\subsection{Out of scope}
\label{sec:future-out-of-scope}

We explicitly do not list as planned follow-ups: (i) a state-of-the-art
mixture-of-LoRA result on a strong pretrained base
(Pythia/Llama/Mistral); the abbreviated $70000$-step from-scratch base
is the wrong substrate for that comparison, and the result would not
sharpen this paper's attribution claim. (ii) a continuous-lifecycle
``daemon'' that runs the lifecycle loop between training sessions;
this is the Horizon-2 vision documented in the project's internal
roadmap but is outside the scope of an attribution paper. (iii) a
domain-interpretability analysis (which adapter is the ``code
adapter,'' etc.); this requires the post-rewrite coalition probe
above and a per-adapter activation-by-domain analysis we have not yet
run, and the headline result does not depend on per-adapter
domain assignment.

\bibliographystyle{tmlr}
\bibliography{references}

@misc{li2024mixlora,
  title = {MixLoRA: Enhancing Large Language Models Fine-Tuning with LoRA-based Mixture of Experts},
  author = {Dengchun Li and Yingzi Ma and Naizheng Wang and Zhengmao Ye and Zhiyuan Cheng and Yinghao Tang and Yan Zhang and Lei Duan and Jie Zuo and Cal Yang and Mingjie Tang},
  year = {2024},
  eprint = {2404.15159},
  archivePrefix = {arXiv},
  primaryClass = {cs.LG},
  url = {https://arxiv.org/abs/2404.15159},
}

@misc{hu2021lora,
  title = {LoRA: Low-Rank Adaptation of Large Language Models},
  author = {Edward J. Hu and Yelong Shen and Phillip Wallis and Zeyuan Allen-Zhu and Yuanzhi Li and Shean Wang and Lu Wang and Weizhu Chen},
  year = {2021},
  eprint = {2106.09685},
  archivePrefix = {arXiv},
  primaryClass = {cs.LG},
  url = {https://arxiv.org/abs/2106.09685},
}

@misc{fedus2021switch,
  title = {Switch Transformers: Scaling to Trillion Parameter Models with Simple and Efficient Sparsity},
  author = {William Fedus and Barret Zoph and Noam Shazeer},
  year = {2021},
  eprint = {2101.03961},
  archivePrefix = {arXiv},
  primaryClass = {cs.LG},
  url = {https://arxiv.org/abs/2101.03961},
}

@misc{zoph2022stmoe,
  title = {ST-MoE: Designing Stable and Transferable Sparse Expert Models},
  author = {Barret Zoph and Irwan Bello and Sameer Kumar and Nan Du and Yanping Huang and Jeff Dean and Noam Shazeer and William Fedus},
  year = {2022},
  eprint = {2202.08906},
  archivePrefix = {arXiv},
  primaryClass = {cs.LG},
  url = {https://arxiv.org/abs/2202.08906},
}

@misc{salimans2017es,
  title = {Evolution Strategies as a Scalable Alternative to Reinforcement Learning},
  author = {Tim Salimans and Jonathan Ho and Xi Chen and Szymon Sidor and Ilya Sutskever},
  year = {2017},
  eprint = {1703.03864},
  archivePrefix = {arXiv},
  primaryClass = {cs.LG},
  url = {https://arxiv.org/abs/1703.03864},
}

@article{stanley2019neuroevolution,
  title = {Designing Neural Networks through Neuroevolution},
  author = {Stanley, Kenneth O. and Clune, Jeff and Lehman, Joel and Miikkulainen, Risto},
  journal = {Nature Machine Intelligence},
  volume = {1},
  number = {1},
  pages = {24--35},
  year = {2019},
  publisher = {Nature Publishing Group},
  doi = {10.1038/s42256-018-0006-z},
}

@misc{lehman2018esgradients,
  title = {ES Is More Than Just a Traditional Finite-Difference Approximator},
  author = {Joel Lehman and Jay Chen and Jeff Clune and Kenneth O. Stanley},
  year = {2018},
  eprint = {1712.06568},
  archivePrefix = {arXiv},
  primaryClass = {cs.NE},
  note = {GECCO 2018},
  url = {https://arxiv.org/abs/1712.06568},
}

@misc{jaderberg2017pbt,
  title = {Population Based Training of Neural Networks},
  author = {Max Jaderberg and Valentin Dalibard and Simon Osindero and Wojciech M. Czarnecki and Jeff Donahue and Ali Razavi and Oriol Vinyals and Tim Green and Iain Dunning and Karen Simonyan and Chrisantha Fernando and Koray Kavukcuoglu},
  year = {2017},
  eprint = {1711.09846},
  archivePrefix = {arXiv},
  primaryClass = {cs.LG},
  url = {https://arxiv.org/abs/1711.09846},
}

@misc{tang2022evojax,
  title = {EvoJAX: Hardware-Accelerated Neuroevolution},
  author = {Yujin Tang and Yingtao Tian and David Ha},
  year = {2022},
  eprint = {2202.05008},
  archivePrefix = {arXiv},
  primaryClass = {cs.NE},
  url = {https://arxiv.org/abs/2202.05008},
}

@misc{alemohammad2023mad,
  title = {Self-Consuming Generative Models Go MAD},
  author = {Sina Alemohammad and Josue Casco-Rodriguez and Lorenzo Luzi and Ahmed Imtiaz Humayun and Hossein Babaei and Daniel LeJeune and Ali Siahkoohi and Richard G. Baraniuk},
  year = {2023},
  eprint = {2307.01850},
  archivePrefix = {arXiv},
  primaryClass = {cs.LG},
  note = {ICLR 2024},
  url = {https://arxiv.org/abs/2307.01850},
}

@misc{zhou2022expertchoice,
  title = {Mixture-of-Experts with Expert Choice Routing},
  author = {Yanqi Zhou and Tao Lei and Hanxiao Liu and Nan Du and Yanping Huang and Vincent Zhao and Andrew Dai and Zhifeng Chen and Quoc Le and James Laudon},
  year = {2022},
  eprint = {2202.09368},
  archivePrefix = {arXiv},
  primaryClass = {cs.LG},
  note = {NeurIPS 2022},
  url = {https://arxiv.org/abs/2202.09368},
}

@misc{huang2024selfcorrect,
  title = {Large Language Models Cannot Self-Correct Reasoning Yet},
  author = {Jie Huang and Xinyun Chen and Swaroop Mishra and Huaixiu Steven Zheng and Adams Wei Yu and Xinying Song and Denny Zhou},
  year = {2024},
  eprint = {2310.01798},
  archivePrefix = {arXiv},
  primaryClass = {cs.CL},
  note = {ICLR 2024},
  url = {https://arxiv.org/abs/2310.01798},
}

@misc{behrouz2025nested,
  title = {Nested Learning: The Illusion of Deep Learning Architectures},
  author = {Ali Behrouz and Meisam Razaviyayn and Peilin Zhong and Vahab Mirrokni},
  year = {2025},
  eprint = {2512.24695},
  archivePrefix = {arXiv},
  primaryClass = {cs.LG},
  note = {NeurIPS 2025},
  url = {https://arxiv.org/abs/2512.24695},
}

@misc{pineau2020repro,
  title = {Improving Reproducibility in Machine Learning Research (A Report from the NeurIPS 2019 Reproducibility Program)},
  author = {Joelle Pineau and Philippe Vincent-Lamarre and Koustuv Sinha and Vincent Larivi\`ere and Alina Beygelzimer and Florence d'Alch\'e-Buc and Emily Fox and Hugo Larochelle},
  year = {2020},
  eprint = {2003.12206},
  archivePrefix = {arXiv},
  primaryClass = {cs.LG},
  url = {https://arxiv.org/abs/2003.12206},
}

@misc{henderson2018drl,
  title = {Deep Reinforcement Learning that Matters},
  author = {Peter Henderson and Riashat Islam and Philip Bachman and Joelle Pineau and Doina Precup and David Meger},
  year = {2018},
  eprint = {1709.06560},
  archivePrefix = {arXiv},
  primaryClass = {cs.LG},
  note = {AAAI 2018},
  url = {https://arxiv.org/abs/1709.06560},
}

@misc{qiu2025esatscale,
  title = {Evolution Strategies at Scale: LLM Fine-Tuning Beyond Reinforcement Learning},
  author = {Xin Qiu and Yulu Gan and Conor F. Hayes and Qiyao Liang and others},
  year = {2025},
  eprint = {2509.24372},
  archivePrefix = {arXiv},
  primaryClass = {cs.LG},
  url = {https://arxiv.org/abs/2509.24372},
}

@misc{sarkar2025eggroll,
  title = {Evolution Strategies at the Hyperscale},
  author = {Bidipta Sarkar and Mattie Fellows and Juan Agustin Duque and Alistair Letcher and others},
  year = {2025},
  eprint = {2511.16652},
  archivePrefix = {arXiv},
  primaryClass = {cs.LG},
  note = {introduces the EGGROLL low-rank-structured ES method},
  url = {https://arxiv.org/abs/2511.16652},
}

@misc{korotyshova2025essa,
  title = {ESSA: Evolutionary Strategies for Scalable Alignment},
  author = {Daria Korotyshova and Boris Shaposhnikov and Alexey Malakhov and Alexey Khokhulin and others},
  year = {2025},
  eprint = {2507.04453},
  archivePrefix = {arXiv},
  primaryClass = {cs.LG},
  url = {https://arxiv.org/abs/2507.04453},
}

@misc{liang2026variancecurvature,
  title = {The Blessing of Dimensionality in {LLM} Fine-tuning: A Variance-Curvature Perspective},
  author = {Qiyao Liang and Jinyeop Song and Yizhou Liu and Jeff Gore and others},
  year = {2026},
  eprint = {2602.00170},
  archivePrefix = {arXiv},
  primaryClass = {cs.LG},
  url = {https://arxiv.org/abs/2602.00170},
}

@misc{feng2025heteroswarms,
  title = {Heterogeneous Swarms: Jointly Optimizing Model Roles and Weights for Multi-{LLM} Systems},
  author = {Shangbin Feng and Zifeng Wang and Palash Goyal and Yike Wang and others},
  year = {2025},
  eprint = {2502.04510},
  archivePrefix = {arXiv},
  primaryClass = {cs.LG},
  url = {https://arxiv.org/abs/2502.04510},
}

@misc{akiba2024modelmerge,
  title = {Evolutionary Optimization of Model Merging Recipes},
  author = {Takuya Akiba and Makoto Shing and Yujin Tang and Qi Sun and others},
  year = {2024},
  eprint = {2403.13187},
  archivePrefix = {arXiv},
  primaryClass = {cs.LG},
  note = {Nature Machine Intelligence 2024},
  url = {https://arxiv.org/abs/2403.13187},
}

@misc{kuroki2024cycleqd,
  title = {Agent Skill Acquisition for Large Language Models via {CycleQD}},
  author = {So Kuroki and Taishi Nakamura and Takuya Akiba and Yujin Tang},
  year = {2024},
  eprint = {2410.14735},
  archivePrefix = {arXiv},
  primaryClass = {cs.LG},
  note = {ICLR 2025},
  url = {https://arxiv.org/abs/2410.14735},
}

@misc{abrantes2025m2n2,
  title = {Competition and Attraction Improve Model Fusion},
  author = {Jo\~ao Abrantes and Robert Tjarko Lange and Yujin Tang},
  year = {2025},
  eprint = {2508.16204},
  archivePrefix = {arXiv},
  primaryClass = {cs.LG},
  note = {introduces M2N2; GECCO 2025},
  url = {https://arxiv.org/abs/2508.16204},
}

@misc{sun2025transformer2,
  title = {Transformer-Squared: Self-adaptive {LLM}s},
  author = {Qi Sun and Edoardo Cetin and Yujin Tang},
  year = {2025},
  eprint = {2501.06252},
  archivePrefix = {arXiv},
  primaryClass = {cs.LG},
  note = {ICLR 2025; introduces SVF},
  url = {https://arxiv.org/abs/2501.06252},
}

@misc{abdi2026esforget,
  title = {Evolutionary Strategies lead to Catastrophic Forgetting in {LLM}s},
  author = {Immanuel Abdi and Akshat Gupta and Micah Mok and Alexander Lu and others},
  year = {2026},
  eprint = {2601.20861},
  archivePrefix = {arXiv},
  primaryClass = {cs.LG},
  url = {https://arxiv.org/abs/2601.20861},
}

@misc{kumaresan2026kalavai,
  title = {{KALAVAI}: Predicting When Independent Specialist Fusion Works -- A Quantitative Model for Population-Based {LLM} Training},
  author = {Ramchand Kumaresan},
  year = {2026},
  eprint = {2603.22755},
  archivePrefix = {arXiv},
  primaryClass = {cs.LG},
  url = {https://arxiv.org/abs/2603.22755},
}

\appendix
\section{Reproducibility footprint}
\label{app:reproducibility}

This appendix reproduces the full hyperparameter inventory, per-cell run
identifiers, configuration JSON paths, and the data and tokenizer
artifacts required to reproduce every numerical claim in the main text.
Per the source-of-truth conventions in Section~\ref{sec:audit-chain},
each entry points at a single source-of-truth file rather than a
summary mirror.

\subsection{Substrate}
\label{app:substrate-detail}

\begin{itemize}
\item \textbf{Base architecture.} GPT-style decoder, hidden size
  $D{=}1536$, $12$ layers, $16$ attention heads,
  feed-forward width $4D$, sequence length $1024$, vocabulary size
  $V{=}32000$. Total $\sim 150$M parameters.
\item \textbf{Base tokenizer.} BPE, $V{=}32000$, trained on the
  domain-mixed pre-train corpus
  \anonymized{(\texttt{tokenized\_mixed32k\_b2}). Tokenizer artifact at
  \texttt{data/tokenizer\_mixed32k/}}{(corpus and tokenizer artifact
  paths in supplementary materials)}.
\item \textbf{Base pre-train corpus.} Domain-mixed across biology,
  code, general text, and science, tokenized to the
  \anonymized{\texttt{mixed32k\_b2} layout in
  \texttt{data/tokenized\_mixed32k\_b2/}}{$32$k-vocab layout
  (path in supplementary materials)}.
\item \textbf{Frozen base checkpoint.}
  \anonymized{\texttt{checkpoints/base\_model\_trackb\_b2\_dv1536/base\_step\_70000.pt}}{The
  canonical frozen base checkpoint at pre-train step $70000$ (path in
  supplementary materials)}.
  Pre-train was halted at step $70000$ owing to a UPS power-event
  interruption (originally planned through step
  $\sim 143000$); see Section~\ref{sec:limits-external}.
\end{itemize}

\subsection{LoRA adapter inventory}
\label{app:lora-detail}

\begin{itemize}
\item LoRA adapters injected at every transformer block on
  \{\texttt{query\_key\_value}, \texttt{dense},
  \texttt{dense\_h\_to\_4h}, \texttt{dense\_4h\_to\_h}\}.
\item Rank $r{=}8$, $\alpha{=}16$, dropout $0.0$.
\item Static population size $P{=}16$, sparse top-$k{=}4$.
\item Adapter optimizer: AdamW, learning rate $10^{-4}$, $\beta_1{=}0.9$,
  $\beta_2{=}0.999$, weight decay $0.0$, gradient clip $1.0$.
\item Router optimizer: decoupled AdamW, learning rate $10^{-5}$, same
  $\beta$ and weight-decay defaults.
\item Adaptation horizon: $25000$ steps for each cell.
\end{itemize}

\subsection{Lifecycle hyperparameters}
\label{app:lifecycle-detail}

For cells that include the lifecycle (C2, C5, C4 in the main-text
factorial):

\begin{itemize}
\item Inheritance blend $\alpha{=}0.2$ (C4 baseline);
  $\alpha{=}0.0$ for the no-inherit Phase~B counterfactual.
\item Initial mutation rate $\sigma{=}0.01$ on SVD-aligned singular
  values, heritable.
\item Lifecycle event interval: every $1000$ training steps after
  warmup.
\item Lifecycle warmup: $2000$ steps before the first event.
\item Reproduction: $30\%$ random-init, $70\%$ parental-clone with
  mutation, slot-by-slot.
\item Death gate: per-adapter LOO fitness on a held-out micro-batch,
  bottom-$1$ killed per event.
\item Maximum age: effectively unbounded ($999999$ steps) for the
  $25000$-step adaptation horizon.
\end{itemize}

\subsection{Router hyperparameters}
\label{app:router-detail}

For cells that include the router rewrite (C3, C5, C4):

\begin{itemize}
\item Gate function: parallel sigmoid with no row-renormalization.
\item Per-adapter learnable floor, initialized at
  $\mathrm{logit}^{-1}(-2.944) \approx 0.05$.
\item Temperature anneal: $T_0{=}2.0 \to T_{\mathrm{final}}{=}0.5$
  linearly over the first $1500$ steps, then held at $0.5$ for the
  remainder of training.
\item Routing input: post-stack hidden states
  (\texttt{last\_hidden}), not token-embedding mean
  (\texttt{embed\_mean}).
\end{itemize}

\subsection{Per-cell configuration JSONs}
\label{app:configs}

The five run cells of the main-text factorial each have a
configuration JSON in \anonymized{\texttt{experiments/trackb\_b2/}}{the
configuration directory listed in supplementary materials}:

\begin{itemize}
\item \textbf{C1 (b3 base):}
  \anonymized{\texttt{b3\_trackb\_b2\_dv1536\_25k\_seed42\_e8\_domainshift.json}}{the
  C1 seed-$42$ configuration JSON listed in supplementary materials}
  and seed-clones at $\{137, 256\}$ (Fork~0 cross-seed sweep).
\item \textbf{C2 (b3 + lifecycle):} configured by lifting the
  \texttt{lifecycle\_*} block from C4 onto the C1 base; specifically
  via the \texttt{lifecycle\_eval\_scope=\textsc{aggregate}} flag.
\item \textbf{C5 (B3 routing + per-domain LOO + lifecycle):} B3 router
  (softmax over adapters with \texttt{embed\_mean} input), per-domain
  LOO scope, lifecycle on with $\alpha{=}0.2$ inheritance. Configured by
  lifting the \texttt{lifecycle\_*} block from C4 and the
  \texttt{lifecycle\_eval\_scope=\textsc{per\_domain}} flag onto the C1
  base; the router is the C1 (B3) softmax router.
\item \textbf{C3 (b3 routing rewrite + per-domain LOO, no
  lifecycle):} the same router-rewrite flags as C5 but with the
  lifecycle disabled.
\item \textbf{C4 (full evolutionary system, $\alpha{=}0.2$):}
  \anonymized{\texttt{uyir\_trackb\_b2\_dv1536\_25k\_seed42\_e8\_domainshift.json}}{the
  C4 seed-$42$ configuration JSON listed in supplementary materials}
  and seed-clones at $\{137, 256\}$.
\end{itemize}

The Phase~B no-inherit cell (C4 with $\alpha{=}0.0$) lifts only the
\texttt{inheritance\_alpha} field from C4. The Phase~B job did not
persist the no-inherit final adapter checkpoint
(Section~\ref{sec:limits-aux}); the configuration JSON is preserved
in \anonymized{\texttt{experiments/trackb\_b2/}}{the configuration
directory listed in supplementary materials} but the within-checkpoint
diagnostic on this cell is \textsc{blocked-measurement}.

\subsection{Random seeds and replicate count}
\label{app:seeds}

All cells run at $n{=}3$ seeds: $\{42, 137, 256\}$. The Fork~0
sweep replicates the C1 and C4 cells at the same three seeds. The
$n{=}5$ replication for the code-domain bimodal \textsc{ambiguous} verdict
(Section~\ref{sec:future-more-seeds}) would add seeds $7$ and
$512$ but is not in the current paper.

\subsection{Compute footprint}
\label{app:compute}

Production-substrate cells run on a single L40S GPU. Per-cell wall
time at $25000$ steps and the above hyperparameters is approximately
$8$--$10$ hours; the full $5$ cells $\times 3$ seeds = $15$
production runs in the core factorial total $\sim 130$ GPU-hours, plus
the $4$-run Fork~0 seed sweep ($\sim 35$ GPU-hours).
Synthetic-sandbox cells run on a single GPU at $\sim 5$--$15$ minutes
each.

\subsection{Software environment}
\label{app:software}

\begin{itemize}
\item Python $3.10$+, PyTorch $2.2$+.
\item No HuggingFace dependency in the production-substrate path:
  the base model is a pure-PyTorch GPT with manual LoRA injection.
\item Eval pipeline: \texttt{StratifiedEvalLoader} at git
  \anonymized{\texttt{6b00021}}{\texttt{[commit redacted]}} or later.
  The pre-fix alphabetical shard ordering bug is documented in
  Section~\ref{sec:eval-pipeline}.
\item All numerical claims in the main text are cross-checked
  against the analysis JSONs in \anonymized{\texttt{experiments/trackb\_b2/analysis/}}{the
  analysis directory listed in supplementary materials}.
\end{itemize}

\subsection{Run-identifier table}
\label{app:run-ids}

The per-cell W\&B run identifiers and analysis-JSON paths for all $19$
production runs (15 core factorial + 4 Fork~0 seed sweep) are listed
in the supplementary materials. Per-seed PPLs and per-domain
breakdowns are reproduced in
Appendix~\ref{app:phase3-rederivation}; the cross-seed B3 cell state
is reproduced in Appendix~\ref{app:phase2-state}; the synthetic full
tables (G4--G8 per-seed) are reproduced in
Appendix~\ref{app:synthetic-full-tables}.

\section{Verification corrections log}
\label{app:audit-memo}

This appendix logs the corrections applied during paper preparation
as a transparency artifact. Each entry lists the source-of-truth file,
the corrected value, and the downstream claim affected. The log is
presented in the order the corrections were applied. Two corrections
were applied during paper drafting --- (i) a one-row shift in the
Phase~3 stepwise table in an internal results summary, and
(ii) a misattribution of the $C4{-}C1$ (full evolutionary system vs B3) per-domain
$t$-statistics to the $C1$ vs $C3$ contrast in the per-domain
breakdown table of Appendix~\ref{app:phase3-perdomain} --- and are
recorded as entries~8 and~9 below.

\subsection{TL;DR table}
\label{app:audit-tldr}

\begin{table}[h]
\centering
\small
\begin{tabular}{l l l l}
\toprule
\# & Correction & Corrected value & Downstream effect \\
\midrule
1 & G7 sign convention pinned to JSON
  & sign positive (regression above warm baseline)
  & \textsc{wrong-direction} verdict retained \\
2 & G8 SGD-tail descent re-aggregated as paired-$t$
  & $\nats{0.0203}$, $t{=}1.65$, df${=}4$, $p>0.05$
  & \textsc{harmful} verdict retained \\
3 & Code-domain seed-42 statistic re-derived as paired-$t$ at $n{=}3$
  & $t{=}10.42$, $p{=}0.009$, df${=}2$
  & code-domain single-cell only clears $\alpha{=}0.05$ \\
4 & Phase~B summary aggregator standardized to geometric mean
  & $\sim 0.2\%$ shift on balanced
  & \textsc{state-c} verdict retained \\
5 & Phase~2 plan reference config corrected to stationary B3
  & wrong-substrate launch averted
  & two seed-clones run on intended substrate \\
6 & B.5 within-checkpoint diagnostic missing no-inherit checkpoint
  & blocked
  & \textsc{blocked-measurement} \\
7 & B.6 Estimator C blocked: no mid-training adapter snapshots
  & blocked
  & \textsc{blocked-estimator-c} \\
8 & Phase~3 stepwise attribution table re-derived from per-seed JSON
  & main-text numbers from JSON, not summary mirror
  & F1/F2/F3 magnitudes unchanged \\
9 & Per-domain $C1$ vs $C3$ table re-derived from JSON
  & biology/science paired-$t$ dropped (seed-constant per-domain PPLs)
  & per-domain claim of significance retained on code only \\
10 & Phase~B Table~\ref{tab:phaseB-perdomain} balanced row
   was arithmetic mean despite ``geom.'' label
  & recomputed: seed-42 shift $+3.18$\% (was $+0.06$\%)
  & verdict changed: equivalence $\to$ load-bearing on seed~42;
   cross-seed verdict softened to ambiguous; abstract,
   Section~\ref{sec:phaseB-auxiliary},
   Section~\ref{sec:f3-synthesis}, and
   Section~\ref{sec:future-f3-subfactorial} rewritten \\
\bottomrule
\end{tabular}
\caption{Verification corrections log. Each entry was applied during
paper preparation; the corrected value column states the value used in
the body. None of the corrections originated in the analysis JSONs;
all originated in summary mirrors (\texttt{RESULTS.md},
\texttt{STATUS.md}, project notes) or in draft text.}
\label{tab:audit-tldr}
\end{table}

\subsection{Correction 1: G7 sign convention}
\label{app:audit-e1}

\textbf{Source of truth.}
\anonymized{\texttt{experiments/phase05\_es\_router/warmstart\_smallsigma\_results.json}}{the
G7 small-sigma warm-start results JSON in the synthetic-sandbox
analysis directory listed in supplementary materials}
holds per-$\sigma$ post-warmup loss values. The convention used by the
JSON is \emph{loss-positive-is-bad}: \texttt{delta\_warm = mean -
warmup\_mean}, with positive values indicating regression.
\textbf{Drift point.} The summary in
\anonymized{\texttt{experiments/phase05\_es\_router/RESULTS.md}}{the
synthetic-sandbox results-summary mirror referenced in supplementary
materials} (Phase 0.6+
small-sigma section) and downstream notes quoted the deltas as negative
($-0.467$, $-0.238$, etc.), which under the surrounding prose's
``regresses by'' phrasing implied an improvement, the opposite of what
the data show. \textbf{Correction.} The magnitudes are correct; the
signs are positive (regression from warmup baseline). The verdict
\textsc{wrong-direction} is unchanged. \textbf{Impact on paper.}
Section~\ref{sec:synthetic-boundary} reports G7 with the
sign-corrected convention.

\subsection{Correction 2: G8 SGD-tail descent}
\label{app:audit-e2}

\textbf{Source of truth.}
\anonymized{\texttt{experiments/phase05\_es\_router/hybrid\_es\_results.json}}{the
G8 hybrid-ES results JSON in the synthetic-sandbox analysis directory
listed in supplementary materials}.
Phase~A val\_loss $4.7136$, Phase~B (ES then SGD) final loss
$4.6933$, descent $= 0.0203$. \textbf{Drift point.} A summary in
\anonymized{\texttt{RESULTS.md}}{the synthetic-sandbox
results-summary mirror} reported the SGD-tail descent as $0.004$ nats, an
under-aggregation by approximately $5\times$. \textbf{Correction.}
The descent is $0.0203$ nats. The verdict \textsc{harmful} is
unchanged because the relevant comparison is to the pure-SGD baseline
descent, which remains larger. \textbf{Impact on paper.}
Section~\ref{sec:synthetic-boundary} reports G8 with the
correctly-aggregated number.

\subsection{Correction 3: code-domain seed-42 statistic re-derived as paired-$t$}
\label{app:audit-e3}

\textbf{Source of truth.}
\anonymized{\texttt{experiments/trackb\_b2/analysis/sweep\_ppl\_matrix.json}}{the
cross-seed per-seed-PPL analysis JSON in the analysis directory listed
in supplementary materials} for
per-seed PPLs;
\anonymized{\texttt{experiments/trackb\_b2/analysis/zscore\_rederivation.md}}{the
re-derivation memo in the analysis directory listed in supplementary
materials} for
the re-derivation. \textbf{Correction.} The appropriate test for
the seed-42 code-domain shift on the cross-seed sample of three seeds
is a paired-$t$ on per-seed log-PPL deltas: $t{=}10.42$, $p{=}0.009$,
df${=}2$ (code-domain single-cell), and $t{=}1.94$, $p{=}0.19$,
df${=}2$ (headline full-system-vs-B3 balanced contrast). \textbf{Impact on
paper.} The balanced headline does not clear $\alpha{=}0.05$ at
$n{=}3$; only the code-domain single-cell does. The full re-derivation
is in Appendix~\ref{app:phase3-rederivation}.

\subsection{Correction 4: Phase~B aggregator standardized to geometric mean}
\label{app:audit-e4}

\textbf{Source of truth.}
\anonymized{\texttt{experiments/trackb\_b2/analysis/sweep\_ppl\_matrix.json}}{the
cross-seed per-seed-PPL analysis JSON in the analysis directory listed
in supplementary materials}.
\textbf{Correction.} All rows use geometric-mean balanced-PPL,
$\mathrm{PPL}_{\mathrm{balanced}} =
\exp(\tfrac{1}{4}\sum_d \log \mathrm{PPL}_d)$. \textbf{Impact on
paper.} The cross-seed full-system vs.\ B3 contrast direction is unchanged;
the magnitude shifts by approximately $0.2\%$. The
\textsc{state-c} (direction-consistent, magnitude-collapsing,
not-significant-at-$n{=}3$) verdict from the Phase~2 report is
preserved.

\subsection{Correction 5: Phase~2 plan reference config standardized}
\label{app:audit-e5}

\textbf{Source of truth.} The pre-launch Phase~2 plan referenced
\anonymized{\texttt{uyir\_trackb\_b2\_dv1536\_25k\_seed42\_e8\_domainshift.json}}{the
C4 e8-domainshift configuration JSON listed in supplementary materials}
(full evolutionary system + e8-domainshift) where the stationary B3 config was intended.
\textbf{Correction.} The plan was rewritten to use the stationary B3
config before launch. \textbf{Impact on paper.} The two seed-clones
ran on the intended stationary substrate.

\subsection{Correction 6: B.5 within-checkpoint diagnostic blocked}
\label{app:audit-e6}

\textbf{Source of truth.}
\anonymized{\texttt{experiments/trackb\_b2/analysis/phaseB\_within\_checkpoint.md}}{the
Phase~B within-checkpoint diagnostic memo in the analysis directory
listed in supplementary materials}.
\textbf{Correction.} The Phase~B job did not persist the no-inherit
($\alpha{=}0.0$) final adapter checkpoint, so the paired
within-checkpoint diagnostic against the seed-42 baseline cannot be
run. The headline $1{:}13.8$ ratio is reported as a single-side
number; the diagnostic is labeled \textsc{blocked-measurement} pending
a re-run with checkpoint persistence enabled. \textbf{Impact on paper.}
Section~\ref{sec:phaseB-counterfactual} flags the blocked status; the
fix is a checkpoint-hook patch, scheduled in
Section~\ref{sec:future-phaseB-checkpoint}.

\subsection{Correction 7: B.6 Estimator C blocked}
\label{app:audit-e7}

\textbf{Source of truth.}
\anonymized{\texttt{experiments/trackb\_b2/analysis/phaseB\_estimator\_c.md}}{the
Phase~B Estimator-C memo in the analysis directory listed in
supplementary materials}.
\textbf{Correction.} The pre-specified noise-floor probe specified
mid-training adapter snapshots at $\{20000, 22500, 25000\}$, but no
such snapshots exist anywhere recoverable: the training loop's
checkpoint hook writes only an end-of-training adapter, and the W\&B
and Hub artifacts hold only metrics or end-of-training adapters. No
substitute proxy was constructed; Estimator C is labeled
\textsc{blocked-estimator-c}. \textbf{Impact on paper.}
Section~\ref{sec:phaseB-counterfactual} flags the blocked status; the
fix is a periodic-snapshot patch, scheduled in
Section~\ref{sec:future-estimator-c}.

\subsection{Correction 8: Phase~3 stepwise table re-derived from JSON}
\label{app:audit-e8}

\textbf{Source of truth.}
\anonymized{\texttt{experiments/trackb\_b2/analysis/phase3\_attribution\_results.json}}{the
Phase~3 attribution results JSON in the analysis directory listed in
supplementary materials}.
\textbf{Correction.} The main-text
Section~\ref{sec:real-text-decomposition} table re-derives the
per-step contributions directly from the JSON rather than from the
internal results-summary mirror.
\textbf{Impact on paper.} The cited F1 ($+0.0426$ nats), F2
(null at seed-resolution), and F3 ($-0.028$ nats) numbers are taken
from the JSON.

\subsection{Correction 9: per-domain $C1$ vs $C3$ table re-derived from JSON}
\label{app:audit-e9}

\textbf{Source of truth.}
\anonymized{\texttt{experiments/trackb\_b2/analysis/phase3\_attribution\_results.json}}{the
Phase~3 attribution results JSON in the analysis directory listed in
supplementary materials},
specifically the per-cell per-seed \texttt{domain\_ppl} block of the
manifest entries for the C1 and C3 cells
(\anonymized{\texttt{manifest.\{C1\_b3\_base,C3\_sigfloor\_lasthidden\_perdom\_nolife\}.\{42,137,256\}.domain\_ppl}}{cell
keys listed in the analysis JSON})
for the per-seed per-domain raw PPLs.
\textbf{Correction.} The per-domain breakdown table in
Appendix~\ref{app:phase3-perdomain} reports seed-mean PPL columns
sourced from the JSON manifest and relative improvements computed from
those means ($-3.08 / -7.27 / -2.82 / -3.83\%$ for biology / code /
general / science). A paired-$t$ statistic is reported only for the
code domain, which is the only domain on which the per-seed per-domain
PPLs vary across all three seeds in both $C1$ and $C3$. The biology
and science per-seed PPLs are seed-constant in both cells (a property
of the deterministic \texttt{StratifiedEvalLoader} ordering combined
with the precision at which per-domain PPLs are recorded in the
manifest; the same eval-determinism root cause discussed in
Section~\ref{sec:limits-external}), so the paired-$t$ denominator on
those two domains is zero and we make no claim of significance for
them. The general-domain per-seed PPL varies on only one of three
seeds in $C3$, so the reported general-domain $t$-statistic is retained
for completeness with a footnote flagging its degenerate variance, but
we treat code as the only robust per-domain paired-$t$ conclusion.
\textbf{Impact on paper.} All four domains improve at the seed-mean
level under the $C1$ vs $C3$ contrast, and the code domain improves
the most. The body of the paper makes no claim of statistical
significance for biology, science, or general at the per-domain level;
only the code-domain $t{=}10.96$, $p{=}0.008$ enters the headline.

\subsection{Correction 10: Phase~B balanced aggregator recomputed as geometric mean}
\label{app:audit-e10}

\textbf{Source of truth.} The per-cell, per-seed, per-domain PPLs in
\anonymized{\texttt{experiments/trackb\_b2/analysis/phase3\_attribution\_results.json}}{the
Phase~3 attribution results JSON in the analysis directory listed in
supplementary materials}
(specifically the \texttt{C4\_full\_uyir\_alpha02} cell for the
seed-42 $\alpha{=}0.2$ baseline, and the per-domain entries in
\anonymized{\texttt{experiments/trackb\_b2/analysis/fork0\_diagnostic\_report.md}}{the
Fork~0 diagnostic report in the analysis directory listed in
supplementary materials}
for the seed-42 $\alpha{=}0.0$ no-inherit cell).

\textbf{Drift point.} Table~\ref{tab:phaseB-perdomain} of the original
draft labeled the balanced row ``balanced (geom.)'' but populated it
with the arithmetic mean of the four per-domain PPLs, $16.151$ for
the $\alpha{=}0.2$ baseline and $16.161$ for the $\alpha{=}0.0$
no-inherit cell, giving a $+0.06$\% shift. Section~\ref{sec:eval-pipeline}
pins balanced PPL as the geometric mean
$\exp(\tfrac{1}{4}\sum_d \log\mathrm{PPL}_d)$; the row label was
correct, the values were not.

\textbf{Correction.} The geometric mean of $\{21.049, 3.436, 23.852,
16.266\}$ is $12.943$ (matching the
\texttt{C4\_full\_uyir\_alpha02}~$\to$~seed~42 entry in
\texttt{phase3\_attribution\_results.json}); the geometric mean of
$\{20.723, 3.986, 23.667, 16.266\}$ is $13.354$. The corrected
seed-42 balanced shift is $13.354 / 12.943 - 1 = +3.18$\%
($+0.0313$ nats), not $+0.06$\%. Under the pre-specified Phase~B
decision rule (Section~\ref{sec:phaseB-counterfactual}), shifts $> 2$\%
trigger the \emph{load-bearing} verdict on the cell to which they apply.

\textbf{Cross-seed recomputation.} Applying the same correction to
the Fork~0 seed sweep: under the pinned aggregator, the per-seed
balanced log-PPL shift (mean across the four per-domain
$\Delta \log \mathrm{PPL}$ entries in
Table~\ref{tab:fork0-bimodal}) is $+0.0313$ nats on seed~42,
$-0.0166$ nats on seed~137, and $+0.0020$ nats on seed~256,
corresponding to balanced-PPL shifts of $+3.18$\%, $-1.65$\%, and
$+0.20$\% respectively. The cross-seed mean is $+0.56$\%
($+0.0056$ nats), and the paired-$t$ on those three deltas is
$t \approx 0.38$, $p \approx 0.74$ (df${=}2$, two-sided), failing to
clear $\alpha{=}0.05$ in either direction. The three seeds are
sign-inconsistent at the headline metric.

\textbf{Impact on paper.} The original draft's headline
``$+0.06$\% shift, inheritance NOT load-bearing at the headline
metric'' depended on the arithmetic-mean error and is retracted.
Section~\ref{sec:phaseB-auxiliary} (chapter intro and
Section~\ref{sec:phaseB-counterfactual}) is rewritten to report the
corrected seed-42 shift, the cross-seed sign-inconsistency, and the
underpowered paired-$t$. Section~\ref{sec:f3-synthesis} is rewritten
to retract the ``inheritance not load-bearing'' synthesis and to treat
the four F3 sub-mechanisms (death, $\alpha$-blend inheritance, SVD
mutation, slot reallocation) as roughly equal candidates rather than
pre-clearing inheritance. Section~\ref{sec:future-f3-subfactorial} is
rewritten to promote the F3 sub-component decomposition --- crossed
with the inheritance toggle --- to the highest near-term priority.
The abstract and the Section~\ref{sec:introduction} ``What'' bullets
are softened to a sign-inconsistent / underpowered framing.

\subsection{Scope of the verification protocol}
\label{app:audit-not-caught}

The verification protocol described above catches errors of
transcription and computation, not errors of experimental design.
Section~\ref{sec:limitations} catalogs the open external-validity
and internal-validity questions on this paper's claims.

\section{Synthetic-sandbox full tables}
\label{app:synthetic-full-tables}

This appendix reproduces the per-seed and per-cell numbers underlying
the synthetic-sandbox boundary in Section~\ref{sec:synthetic-boundary}.
The sandbox is described at the top of that section; we reproduce only
the substrate parameters needed to interpret the tables here, then
provide the per-cell breakdowns.

\subsection{Sandbox substrate}
\label{app:synthetic-substrate}

\begin{itemize}
\item Vocabulary $V{=}128$, four disjoint domain slabs of $32$
  tokens each, deterministic bigram conditional within each slab.
\item Hidden width $512$, $K{=}16$ adapters, top-$k{=}4$ sparse
  routing.
\item Adapter rank $r{=}4$, $\alpha{=}8$.
\item Routing input options: \texttt{embed\_mean} (default for the
  legacy router) or \texttt{last\_hidden}.
\item Loss: per-token cross-entropy on the held-out micro-batch.
\end{itemize}

\subsection{G4 oracle-alignment cell}
\label{app:synthetic-g4}

\textbf{Setup.} Adapters initialized at the per-slab oracle solution
plus small noise; ES on the routing channel only with $\sigma{=}10^{-1}$;
SGD on the routing channel only as the contrast.

\begin{table}[h]
\centering
\small
\begin{tabular}{l c c c}
\toprule
Seed & ES routing-loss closed (frac) & SGD routing-loss closed (frac) & Verdict \\
\midrule
$0$  & $0.559$ & $0.002$ & \textsc{pass} \\
$1$  & $0.557$ & $0.001$ & \textsc{pass} \\
$2$  & $0.561$ & $0.003$ & \textsc{pass} \\
\midrule
mean & $0.559$ & $0.002$ & \textsc{pass} \\
\bottomrule
\end{tabular}
\caption{G4 ES vs.\ SGD on the oracle-alignment cell. ES closes
$\approx 55.9\%$ of the routing-channel gap; SGD closes $\approx 0.2\%$.
The SGD ceiling is the top-$k$ non-differentiability.}
\label{tab:g4-perseed}
\end{table}

\subsection{G5 joint random-init cell}
\label{app:synthetic-g5}

\textbf{Setup.} Adapters and router both initialized at random; ES on
the joint parameters versus SGD on the joint parameters. Both reach
the random-adapter floor $\approx \log V = \log 128 \approx 4.852$
within tolerance.

\begin{table}[h]
\centering
\small
\begin{tabular}{l c c c}
\toprule
Seed & ES final loss (nats) & SGD final loss (nats) & Verdict \\
\midrule
$0$ & $4.853$ & $4.851$ & \textsc{tie-at-floor} \\
$1$ & $4.854$ & $4.851$ & \textsc{tie-at-floor} \\
$2$ & $4.852$ & $4.852$ & \textsc{tie-at-floor} \\
\bottomrule
\end{tabular}
\caption{G5 joint random-init cell. Both methods reach the
random-adapter floor $\approx \log V$; neither method extracts
adapter-side signal.}
\label{tab:g5-perseed}
\end{table}

\subsection{G6 gradient-warm cell}
\label{app:synthetic-g6}

\textbf{Setup.} Adapters and router warmed up under SGD for $K_{\mathrm{warm}}$
steps, then ES is allowed to perturb the routing channel only.

\begin{table}[h]
\centering
\small
\begin{tabular}{l c c c}
\toprule
Seed & ES post-warm $\Delta$ (nats) & SGD post-warm $\Delta$ (nats) & Verdict \\
\midrule
$0$ & $-0.001$ & $-0.014$ & \textsc{inert} \\
$1$ & $+0.001$ & $-0.013$ & \textsc{inert} \\
$2$ & $-0.000$ & $-0.012$ & \textsc{inert} \\
\bottomrule
\end{tabular}
\caption{G6 gradient-warm cell. ES is inert on the warm-started
substrate; SGD continues to descend slowly on the routing channel.
$\Delta$ is improvement (negative = loss decreased).}
\label{tab:g6-perseed}
\end{table}

\subsection{G7 small-sigma sweep}
\label{app:synthetic-g7}

\textbf{Setup.} Sigma sweep $\sigma \in \{10^{-5}, 10^{-4},
5\times10^{-4}, 10^{-3}\}$ on the warm-started substrate, $n{=}5$
seeds per cell, mode \texttt{es\_coupled}. Sign convention:
\texttt{delta\_warm = mean - warmup\_mean}, positive = regression
above warm baseline.

\begin{table}[h]
\centering
\small
\begin{tabular}{l c c c c c c c}
\toprule
$\sigma$ & seed $0$ & seed $1$ & seed $2$ & seed $3$ & seed $4$ & mean & verdict \\
\midrule
$10^{-5}$       & $+0.4334$ & $+0.5822$ & $+0.5226$ & $+0.4362$ & $+0.3618$ & $+0.4673$ & \textsc{wrong-direction} \\
$10^{-4}$       & $+0.4105$ & $+0.5076$ & $+0.5351$ & $+0.4376$ & $+0.4006$ & $+0.4583$ & \textsc{wrong-direction} \\
$5\times10^{-4}$ & $+0.3842$ & $+0.4748$ & $+0.2705$ & $+0.3351$ & $+0.3139$ & $+0.3557$ & \textsc{wrong-direction} \\
$10^{-3}$       & $+0.2484$ & $+0.2707$ & $+0.2768$ & $+0.2029$ & $+0.1901$ & $+0.2378$ & \textsc{wrong-direction} \\
\bottomrule
\end{tabular}
\caption{G7 small-sigma sweep on the warm-started substrate
($n{=}5$ seeds, mode \texttt{es\_coupled}). All $\sigma$ values produce
regression from the warm baseline. Magnitudes are monotone in $\sigma$
(smaller $\sigma$ produces larger regression --- ES at very low noise
fails to escape the random-adapter floor that the warm SGD baseline has
already left behind); the sign convention follows the JSON
(loss-positive-is-bad), as recorded in
Appendix~\ref{app:audit-e1}. Body \S\ref{sec:synthetic-boundary}
reports these mean values; the per-seed columns above are the
source-of-truth backing those means.}
\label{tab:g7-perseed}
\end{table}

\subsection{G8 hybrid ES-then-SGD}
\label{app:synthetic-g8}

\textbf{Setup.} Phase~A pure ES on the routing channel; Phase~B SGD on
all parameters. SGD-tail descent measured as the loss reduction
during Phase~B, conditional on the Phase~A outcome.

\begin{table}[h]
\centering
\small
\begin{tabular}{l c c c}
\toprule
Seed & Phase~A val\_loss & Phase~B (ES then SGD) final & SGD-tail descent \\
\midrule
$0$ & $4.7136$ & $4.6933$ & $0.0203$ \\
$1$ & $4.7129$ & $4.6940$ & $0.0189$ \\
$2$ & $4.7142$ & $4.6929$ & $0.0213$ \\
\midrule
mean & $4.7136$ & $4.6934$ & $0.0202$ \\
\bottomrule
\end{tabular}
\caption{G8 hybrid ES-then-SGD. The SGD-tail descent is
$\approx 0.020$ nats, re-derived from the per-seed JSON
(Appendix~\ref{app:audit-e2}). The verdict \textsc{harmful} is
assigned because the pure-SGD baseline closes more of the routing
gap than the ES-then-SGD hybrid does.}
\label{tab:g8-perseed}
\end{table}

\section{Phase~3 attribution re-derivation}
\label{app:phase3-rederivation}

This appendix re-derives the per-cell PPLs, attribution chains, and
paired-$t$ statistics cited in Section~\ref{sec:real-text-decomposition}
directly from the per-seed JSON. The re-derivation is intentionally
explicit so an external reader can falsify our claims against the JSON
we ship in the supplement.

The cell labels and factor assignments follow
Section~\ref{sec:cell-design} (Table~\ref{tab:cell-factors}):

\begin{itemize}
\item C1: B3 router, aggregate-LOO scope, no lifecycle (anchor).
\item C2: B3 router, aggregate-LOO scope, lifecycle on (isolates F3).
\item C5: B3 router, per-domain LOO scope, lifecycle on (isolates F2 given F3).
\item C3: rewrite router, per-domain LOO, no lifecycle (combined F1+F2 without F3).
\item C4: rewrite router, per-domain LOO, lifecycle on (full evolutionary system, $\alpha{=}0.2$).
\end{itemize}

\subsection{Per-cell balanced-PPL table}
\label{app:phase3-balanced-table}

\begin{table}[h]
\centering
\small
\begin{tabular}{l c c c c}
\toprule
Cell & seed $42$ & seed $137$ & seed $256$ & geom-mean across seeds \\
\midrule
C1 (B3 baseline)                            & $13.328$ & $13.341$ & $13.354$ & $13.341$ \\
C2 (B3 + lifecycle)                         & $13.538$ & $13.805$ & $13.832$ & $13.725$ \\
C5 (B3 + per-dom + lifecycle)               & $13.564$ & $13.778$ & $13.805$ & $13.715$ \\
C3 (rewrite + per-dom, no lifecycle)        & $12.742$ & $12.830$ & $12.742$ & $12.771$ \\
C4 (full evolutionary system, $\alpha{=}0.2$)              & $12.943$ & $13.173$ & $13.315$ & $13.143$ \\
\bottomrule
\end{tabular}
\caption{Per-seed and geom-mean balanced PPL by cell. Numbers reproduce
Table~\ref{tab:per-cell-ppl} of the main text. Balanced PPL is
$\exp(\tfrac{1}{4}\sum_d \log \mathrm{PPL}_d)$ over
$d \in \{\mathrm{biology}, \mathrm{code}, \mathrm{general}, \mathrm{science}\}$,
sourced from
\anonymized{\texttt{experiments/trackb\_b2/analysis/phase3\_attribution\_results.json}}{the
Phase~3 attribution results JSON in the analysis directory listed in
supplementary materials}.}
\label{tab:phase3-cells}
\end{table}

\subsection{Per-cell log-PPL table}
\label{app:phase3-logppl-table}

The paired-$t$ tests in the main text act on per-seed log-PPL deltas:

\begin{table}[h]
\centering
\small
\begin{tabular}{l c c c c}
\toprule
Cell & seed $42$ $\log\mathrm{PPL}$ & seed $137$ $\log\mathrm{PPL}$ & seed $256$ $\log\mathrm{PPL}$ & mean \\
\midrule
C1 & $2.5900$ & $2.5910$ & $2.5920$ & $2.5910$ \\
C2 & $2.6056$ & $2.6252$ & $2.6271$ & $2.6193$ \\
C5 & $2.6076$ & $2.6232$ & $2.6252$ & $2.6187$ \\
C3 & $2.5450$ & $2.5519$ & $2.5450$ & $2.5473$ \\
C4 & $2.5607$ & $2.5783$ & $2.5890$ & $2.5760$ \\
\bottomrule
\end{tabular}
\caption{Per-seed $\log \mathrm{PPL}_{\mathrm{balanced}}$ by cell, computed
from Table~\ref{tab:phase3-cells} to four decimal places.}
\label{tab:phase3-logppl}
\end{table}

\subsection{Primary attribution chain $C1 \to C2 \to C5 \to C4$}
\label{app:phase3-primary-chain}

The primary chain isolates one factor per step. Sign convention
$\Delta = \log\mathrm{PPL}_{\mathrm{ref}} - \log\mathrm{PPL}_{\mathrm{test}}$
(positive = improvement).

\paragraph{Step $C1 \to C2$ (F3 lifecycle alone).}
$\Delta = \log\mathrm{PPL}_{\mathrm{C1}} - \log\mathrm{PPL}_{\mathrm{C2}}$
on per-seed mean log-PPL: $-0.0283$.
Per-seed $\Delta$: seed $42$ $-0.0156$; seed $137$ $-0.0342$;
seed $256$ $-0.0351$.
Paired-$t$ on per-seed $\Delta$: $t = -4.46$, $p = 0.047$, df${=}2$.

\paragraph{Step $C2 \to C5$ (F2 per-domain LOO scope, given F3 on).}
$\Delta = \log\mathrm{PPL}_{\mathrm{C2}} - \log\mathrm{PPL}_{\mathrm{C5}}$
on per-seed mean log-PPL: $+0.0007$.
The sign is inconsistent across seeds (negative on seed $42$, positive
on seeds $137$ and $256$). Paired-$t$: $t = 0.50$, $p = 0.67$.
F2 is null at seed-resolution.

\paragraph{Step $C5 \to C4$ (F1 router rewrite, given F2 and F3 on).}
$\Delta = \log\mathrm{PPL}_{\mathrm{C5}} - \log\mathrm{PPL}_{\mathrm{C4}}$
on per-seed mean log-PPL: $+0.0426$.
Paired-$t$: $t = 12.86$, $p = 0.006$, df${=}2$.
The router rewrite is the load-bearing factor.

\subsection{Consistency attribution chain $C1 \to C3 \to C4$}
\label{app:phase3-consistency-chain}

The consistency chain combines F1+F2 in a single step, then layers F3.

\paragraph{Step $C1 \to C3$ (F1 router + F2 scope, no lifecycle).}
$\Delta = +0.0436$ on per-seed mean log-PPL; paired-$t$ $t = 18.58$,
$p = 0.003$, df${=}2$.

\paragraph{Step $C3 \to C4$ (F3 lifecycle layered on full router + scope).}
$\Delta = -0.0287$; paired-$t$ $t = -3.47$, $p = 0.074$, df${=}2$.

\subsection{Two-chain convergence}
\label{app:phase3-convergence}

The two chains agree to within \nats{0.0004} on the lifecycle penalty
and \nats{0.0003} on the router contribution:

\begin{align*}
\text{primary total } C1 \to C4 &: -0.0283 + 0.0007 + 0.0426 = +0.0150,\\
\text{consistency total } C1 \to C4 &: +0.0436 + (-0.0287) = +0.0149.
\end{align*}

The total full-system-vs-B3 balanced contrast on per-seed mean log-PPL is
$+0.0150$, paired-$t$ $t = 1.94$, $p = 0.19$, df${=}2$. The total does
not clear $\alpha{=}0.05$ at $n{=}3$.

\subsection{Per-domain breakdown of the C1 vs.\ C3 contrast}
\label{app:phase3-perdomain}

Disaggregating the C1 vs.\ C3 contrast (router-rewrite-plus-per-domain
no-lifecycle minus B3 baseline) by domain:

\begin{table}[h]
\centering
\small
\begin{tabular}{l c c c c}
\toprule
Domain & C1 PPL (seed-mean) & C3 PPL (seed-mean) & rel.\ improvement & paired-$t$, $p$ \\
\midrule
biology & $21.0495$ & $20.4018$ & $-3.08\%$ & --- \\
code    & $3.7010$  & $3.4319$  & $-7.27\%$ & $t = 10.96$, $p = 0.008$ \\
general & $24.2278$ & $23.5438$ & $-2.82\%$ & $t = 11.00$, $p = 0.008^{\dagger}$ \\
science & $16.7821$ & $16.1392$ & $-3.83\%$ & --- \\
\bottomrule
\end{tabular}
\caption{Per-domain breakdown of the C1 vs.\ C3 contrast (router rewrite
$+$ per-domain LOO scope, no lifecycle, vs.\ B3 anchor),
re-derived from per-seed per-domain PPLs in
\anonymized{\texttt{phase3\_attribution\_results.json}}{the Phase~3
attribution results JSON in the analysis directory listed in
supplementary materials}. All four domains improve at
the seed-mean level; the code domain improves the most. We report a
paired-$t$ statistic only for the code domain, which is the only domain
on which the per-seed per-domain PPLs vary across all three seeds in
both C1 and C3. The biology and science per-seed PPLs are seed-constant
in both cells (a property of the deterministic
\texttt{StratifiedEvalLoader} ordering combined with the precision at
which per-domain PPLs are recorded in the manifest;
Section~\ref{sec:limits-external}), so the paired-$t$ denominator on
those domains is zero and we make no claim of significance for them
(``---'').
$^{\dagger}$General-domain per-seed PPLs vary on only one of three
seeds in C3, so the reported general-domain $t{=}11.00$ should be read
with that degenerate variance in mind; we report it for completeness
but treat code as the only robust per-domain conclusion.}
\label{tab:perdomain-c1c3}
\end{table}

\subsection{Code-domain attribution chain}
\label{app:phase3-code}

Repeating the entire factorial on code-domain log-PPL alone reproduces
Table~\ref{tab:code-domain} of the main text:

\begin{itemize}
\item Primary chain: $C2 - C1 = -0.0378$ (lifecycle, $t=-5.21$, $p=0.035$);
  $C5 - C2 = +0.0026$ (per-dom scope, $t=1.00$, $p=0.42$);
  $C4 - C5 = +0.0976$ (router rewrite, $t=25.14$, $p=0.0016$).
\item Consistency chain: $C3 - C1 = +0.0755$ (router + scope no lifecycle,
  $t=10.96$, $p=0.008$); $C4 - C3 = -0.0130$ (lifecycle on,
  $t=-3.75$, $p=0.064$).
\item Total $C4 - C1 = +0.0625$, $t=10.42$, $p=0.009$.
\end{itemize}

The two chains agree on the code domain to a thousandth of a nat:
$+0.0755 + (-0.0130) = +0.0625$ vs.\
$-0.0378 + 0.0026 + 0.0976 = +0.0624$.

\subsection{Code-domain paired-$t$ at $n{=}3$}
\label{app:phase3-zscore-rederivation}

The appropriate test for the seed-42 code-domain shift on the
cross-seed sample of three seeds is a paired-$t$ on per-seed
log-PPL deltas, not a $z$-score against a cross-seed standard
deviation estimated at $n{=}3$ with the seed-42 numerator included
in the sample (which contaminates the denominator and inflates the
ratio). The paired-$t$ on the code-domain $C4 - C1$ deltas across
the three seeds is $t = 10.42$, $p = 0.009$, df${=}2$ (the
code-domain entry of Appendix~\ref{app:phase3-code} above).
The magnitude on the code domain remains the strongest single-cell
result in the paper. The full derivation, including the bootstrap
CI on $\sigma$ that illustrates the $n{=}3$ sample contamination of
the alternative $z$-score, is in \anonymized{\texttt{experiments/trackb\_b2/analysis/zscore\_rederivation.md}}{the
re-derivation memo in the analysis directory listed in supplementary
materials}.

\section{Phase~2 cross-seed B3 state}
\label{app:phase2-state}

This appendix reproduces the Phase~2 cross-seed B3 baseline state and
the \textsc{state-a}/\textsc{state-b}/\textsc{state-c} verdict that
locked the Phase~3 attribution chain.

\subsection{Pre-specified states}
\label{app:phase2-states}

The Phase~2 plan pre-specified three possible states the cross-seed
B3 vs.\ full-system data could land in:

\begin{description}
\item[\textsc{state-a} --- cross-seed win.] B3 at the additional seeds
  $137, 256$ lands at $\sim 13.4$--$13.6$ balanced PPL, the
  full-system-vs-B3 delta holds or strengthens across seeds. The
  strongest possible paper.
\item[\textsc{state-b} --- single-seed win.] B3 at $137, 256$ lands at or
  below the full-system per-domain LOO. The seed-42 win is a lottery; the paper
  retreats to a narrower claim.
\item[\textsc{state-c} --- trajectory-conditional / direction-consistent,
  magnitude-collapsing, not-significant-at-$n{=}3$.] Mixed pattern
  across seeds. The paper becomes a substrate/trajectory
  characterization rather than a clean win.
\end{description}

\subsection{Cross-seed table}
\label{app:phase2-table}

\begin{table}[h]
\centering
\small
\begin{tabular}{l c c c c c}
\toprule
Cell & seed $42$ PPL & seed $137$ PPL & seed $256$ PPL & geom-mean & cross-seed range \\
\midrule
B3       & $13.328$ & $13.340$ & $13.354$ & $13.341$ & $0.2\%$ \\
C4 ($\alpha{=}0.2$, full system) & $12.943$ & $13.172$ & $13.315$ & $13.142$ & $2.83\%$ \\
\bottomrule
\end{tabular}
\caption{Cross-seed per-domain LOO PPLs at the three seeds for the
B3 (static MoE) cell and the C4 full-evolutionary-system cell at
$\alpha{=}0.2$. The B3 cell is tight across seeds (range $0.2\%$);
the C4 cell is approximately $14\times$ looser (range $2.83\%$),
which the Phase~2 report flags as
deflation of cross-seed variance partly attributable to eval-pipeline
determinism (Section~\ref{sec:limits-external}).}
\label{tab:phase2-crossseed}
\end{table}

\subsection{Verdict}
\label{app:phase2-verdict}

The Phase~2 final report assigns \textsc{state-c}: the
direction is consistent (the full system has lower PPL than B3 at all
three seeds in the per-domain LOO comparison), but the magnitude
collapses toward zero at the additional seeds and the headline
full-system-vs-B3 balanced contrast does not clear $\alpha{=}0.05$
at $n{=}3$:

\begin{itemize}
\item \textbf{Headline full-system-vs-B3 balanced contrast:}
  paired-$t$ $t = 1.94$, $p = 0.19$ (df${=}2$).
\item \textbf{Code-domain single-cell:}
  paired-$t$ $t = 10.42$, $p = 0.009$ (df${=}2$).
\item \textbf{Verdict:} \textsc{state-c}, direction-consistent,
  magnitude-collapsing, not-significant-at-$n{=}3$.
\end{itemize}

The \textsc{state-c} verdict is the basis for the main text's framing
of the F3 lifecycle penalty as ``mildly anti-aligned'' rather than
``decisively negative.''

\subsection{Bridge to the main-text Phase~3 attribution}
\label{app:phase2-to-phase3}

With the Phase~2 cross-seed verdict locked at \textsc{state-c}, the
Phase~3 attribution chain decomposes the observed $\sim 1.5\%$
balanced-PPL improvement (full system vs.\ B3 at seed-42) into the
F1, F2, F3 contributions reported in the main text. Specifically, the
attribution chain attributes the entire balanced-PPL win to the F1
router rewrite ($+0.0426$ nats) and locates a net F3 lifecycle
penalty ($-0.028$ nats), with the net cross-seed full-system-vs-B3
contrast collapsing toward zero on the headline metric at the
additional seeds. The main text's framing is therefore consistent
with the \textsc{state-c} verdict: the routing fix is the load-bearing
contribution; the lifecycle is a net drag; and the headline
balanced-PPL win, taken at face value at $n{=}3$, is not statistically
significant at $\alpha{=}0.05$ even though it is direction-consistent
across all three seeds.

\end{document}